\documentclass[Afour,times,sageh]{style/sagej}

\usepackage{amsthm}
\usepackage[T1]{fontenc}
\usepackage[latin9]{inputenc}
\usepackage{calc}
\usepackage{amssymb, amsmath}
\usepackage{graphicx}
\usepackage{xcolor}
\usepackage{mathtools}
\usepackage{hyperref}
\usepackage[capitalize]{cleveref}
\usepackage{subfigure}
\usepackage[noend]{algpseudocode}
\usepackage{algorithm}
\usepackage{multicol}
\usepackage{multirow}
\usepackage{titlesec}
\usepackage[percent]{overpic}
\usepackage[font=footnotesize, skip=1ex, belowskip=-10pt]{caption}

\usepackage{style/perl_math}

\newtheoremstyle{nospace}{2pt}{1.5pt}{\itshape}{}{\bfseries}{:}{.5em}{}
\theoremstyle{nospace}

\newtheorem{assumption}{Assumption}

\newtheorem{definition}{Definition}

\newcommand{\BF}[1]{{\color{teal}{BF: #1}}}

\begin{document}

\runninghead{Forsgren et al.}
\title{Group-$k$ consistent measurement set maximization via maximum clique over k-Uniform hypergraphs for robust multi-robot map merging}

\author{Brendon Forsgren\affilnum{1}, Ram Vasudevan\affilnum{2}, Michael Kaess\affilnum{3}, Timothy W. McLain\affilnum{1}, Joshua G. Mangelson\affilnum{4}}

\affiliation{\affilnum{1} Department of Mechanical Engineering, Brigham Young University\\ \affilnum{2}Department of Mechanical Engineering, University of Michigan\\ \affilnum{3} Robotics Institute at Carnegie Mellon University\\ \affilnum{4} Department of Electrical and Computer Engineering, Brigham Young University}

\corrauth{Brendon Forsgren}
\email{brendon5@byu.edu}


\begin{abstract}
This paper unifies the theory of consistent-set maximization for robust outlier detection in a simultaneous localization and mapping framework. We first describe the notion of pairwise consistency before discussing how a consistency graph can be formed by evaluating pairs of measurements for consistency. Finding the largest set of consistent measurements is transformed into an instance of the maximum clique problem and can be solved relatively quickly using existing maximum-clique solvers. We then generalize our algorithm to check consistency on a group-$k$ basis by using a generalized notion of consistency and using generalized graphs. We also present modified maximum clique algorithms that function on generalized graphs to find the set of measurements that is internally group-$k$ consistent. We address the exponential nature of group-$k$ consistency and present methods that can substantially decrease the number of necessary checks performed when evaluating consistency. We extend our prior work to multi-agent systems in both simulation and hardware and provide a comparison with other state-of-the-art methods.
\end{abstract}

\maketitle

\section{Introduction}

Multi-agent simultaneous localization and mapping (SLAM) refers to the problem of estimating a map of the environment by fusing the measurements collected by multiple robots as they navigate through that environment. For the estimated map to be accurate, both the local trajectories of the vehicles and the relative offsets (translation and orientation) between the trajectories need to be estimated.

In SLAM, the estimation problem is often modeled using a factor graph containing pose and landmark node variables, and factor nodes that encode the relationship between poses and landmarks. A special case of the SLAM problem, called pose graph SLAM, eliminates the landmark nodes and only estimates the vehicle trajectory. We often formulate the problem as the maximum likelihood estimation (MLE) of the time-discretized robot trajectory given odometric and loop-closure measurements as described by \cite{cadena2016past}. Assuming independence and additive Gaussian noise in the measurement and process models, the problem becomes a nonlinear, weighted-least-squares problem that can be solved quickly using available solvers like those presented by \cite{kaess2008isam,kummerle2011g2o,agarwal2012ceres}.

In multi-agent SLAM, multiple vehicles are used to map the environment, resulting in increased scalability and efficiency in the mapping process. However, in addition to estimating the local map, the vehicles must also estimate their relative pose to accurately combine their maps. Generating inter-vehicle measurements is a process that is often susceptible to perceptual aliasing and can be inaccurate. Identifying poor inter-vehicle measurements is a challenging problem given the lack of a single odometry backbone and potentially no prior information on the initial configuration of the vehicles as shown by \cite{pfingsthorn2015generalized}. Prior work by \cite{mangelson2018pairwise} has examined this problem for full-degree-of-freedom constraints between vehicles. In this work, we present a method that will work using low-degree-of-freedom measurements.


Rather than attempt to classify measurements as inliers and outliers, we find the largest consistent set of inter-robot  measurements. In our prior conference paper (\cite{mangelson2018pairwise}\footnote{\copyright 2018 IEEE. Reprinted, with permission, from Pairwise Consistent Measurement Set Maximization for Robust Multi-Robot Map Merging. 2018 IEEE International Conference on Robotics and Automation (ICRA)}), the problem is formulated as a combinatorial optimization problem that seeks to find the largest set of pairwise-consistent measurements. We then show that this problem can be transformed into an instance of the maximum-clique problem, that existing algorithms can be used to find the optimal solution for moderately sized problems, and heuristic-based methods exist that often find the optimal solution for larger numbers of measurements. Lastly, the proposed method is evaluated on both simulated and real-world data showing that the proposed algorithm outperforms existing robust SLAM algorithms in selecting consistent measurements and estimating the merged maps. These contributions are included in \cref{sec:pcm}, \cref{sec:pcm_mc}, and \cref{sec:pcm_eval}.

Our second conference paper (\cite{forsgren2022group}\footnote{\copyright 2022 IEEE. Reprinted, with permission, from Group-$k$ consistent measurement set maximization for robust outlier detection. 2022 IEEE/RSJ International Conference on Intelligent Robots and Systems (IROS)})  generalizes the concept of pairwise consistency to group-$k$ consistency for scenarios, such as range-based SLAM, where pairwise consistency is insufficient to characterize the consistency of a set of measurements. We show that by using a generalized graph, and modifying known maximum-clique algorithms to function over generalized graphs, we can robustly reject outliers in scenarios where pairwise consistency fails. The generalized method was evaluated on simulated data and showed that enforcing group-$k$ consistency outperforms enforcing pairwise consistency. These results are discussed in \cref{sec:gkcm} through \cref{sec:range_slam_results}.

This work builds on prior work and makes the following contributions:

\begin{enumerate}
  \item We develop a framework that takes advantage of the hierarchical structure of consistency to decrease the number of consistency checks needed when building the generalized graph online. (\cref{sec:build_graph})
  \item We evaluate G$k$CM on hardware data recorded by an unmanned underwater vehicle in a range-only SLAM scenario and compare with other outlier-rejection algorithms (\cref{sec:range_slam_results}).
  \item We propose a consistency function that can be used in vision-based multi-agent pose graph optimization problems. We verify this consistency function on both simulated and hardware data (\cref{sec:ma_visual_pgo,sec:ma_visual_eval}).
  \item We compare our maximum-clique algorithms over hypergraphs with other recently developed algorithms (\cref{sec:max_cliqu_of_generalized_graph}, \cref{sec:range_slam_results}, \cref{sec:ma_visual_eval}).
  \item We release a parallelized implementation of our proposed algorithm (\url{https://bitbucket.org/jmangelson/gkcm/src/master/})
\end{enumerate}

The remainder of this paper is organized as follows. In \cref{sec:related_work}, related work is discussed. In \cref{sec:problem_formulation}, the general formulation of the multi-robot pose graph SLAM problem is presented. Pairwise consistency maximization (PCM) is presented in \cref{sec:pcm} and evaluated in \cref{sec:pcm_eval}. Group-$k$ consistency maximization (G$k$CM) is presented in \cref{sec:gkcm}, and is applied to range-based SLAM in \cref{sec:range_slam} and \cref{sec:range_slam_results}, and multi-agent visual pose graph optimization (PGO) in \cref{sec:ma_visual_pgo} and \cref{sec:ma_visual_eval}. Finally in \cref{sec:conclusion}, we conclude.

\section{Related Work}
\label{sec:related_work}

The ability to remove outliers is important to many robotics and computer-vision applications. Given the sensitivity of nonlinear least-squares optimization to poor information, there has been a significant amount of effort dedicated toward developing methods to detect and remove outlier measurements from the optimization problem.

The random sample consensus (RANSAC) algorithm in \cite{hartley2003multiple} is popular in the computer vision community and detects outliers by fitting models to random subsets of the data and counting the number of inliers that belong to each model. The RANSAC algorithm struggles in scenarios where there is no unique model of the underlying data, such as in multi-agent SLAM, or when the outlier ratio is so large that no accurate model of the data can be found. Recent work by \cite{sun2021ransic} has improved the RANSAC algorithm by adding a compatibility score between the random samples. The new technique, called RANSIC, shows improved performance in high-outlier regimes but will still struggle when no unique model of the data exists. A technique called VODRAC introduced by \cite{hu2023vodrac} also improves on the RANSAC and RANSIC algorithms by using a two-point sampling strategy combined with a weight-based voting strategy that speeds up the consensus maximization and is robust in 99\% outlier regimes.

Work by \cite{burguera2022combining} introduces a three step process that combines deep learning with RANSAC and a geometric verification step. They utilize a neural network to detect possible matches between images. RANSAC is then used to estimate the rotation and translation between two images. If the number of correct correspondences found in RANSAC is not sufficiently high, then the loop closure is rejected. Accepted matches are then subjected to a geometry test by tracing a loop using the vehicle odometry and the measurement. If the error in this loop is sufficiently high, then the measurement is also rejected. The technique is strict enough to reject outliers but is not suitable for inter-vehicle measurements since a single odometry backbone may not exist.

Other approaches use the concept of M-estimation. These techniques attempt to detect the presence of outliers during the optimization process and use a robust cost function to decrease their influence in the weighted nonlinear least-squares problem. \cite{sunderhauf2012towards} use switchable constraints, which introduces a switchable error factor that can be turned off if the residual error becomes too high. Dynamic covariance scaling (DCS), introduced by \cite{agarwal2013robust}, generalizes the switchable constraints method by increasing the covariance matrix associated with measurements that have high residual error, essentially smoothing the transition to turning a constraint off. \cite{yang2020graduated} introduce graduated non-convexity (GNC), a technique that first solves a convex approximation of the original problem and iteratively solves less convex approximations until the original problem is solved. The max-mixtures technique presented by \cite{olson2013inference} uses mixtures of Gaussians to model various data modes and can detect outliers in real-time. Each of these methods was designed for a single-agent system, and assumes a trusted odometry backbone is present. To apply these systems successfully in multi-agent scenarios would require a good initialization of the relative pose between agents which is not always available. Expectation maximization techniques are used by \cite{dong2015distributed} and \cite{carlone2014selecting} to detect outliers among inter-robot measurements for multi-agent systems but the technique still requires an initial guess of the relative pose between agents. Most recently \cite{yang2022certifiably} introduce a method called STRIDE that reformulates the estimation problem using standard robust cost functions as a polynomial optimization problem. Their method is certifiably optimal and works with up to 90\% of the measurements being outliers but does not run in real-time.

\cite{carlone2014selecting} noted that classifying measurements as inliers or outliers is an unobservable task. In light of this, the focus of research has changed from classifying measurements as inliers and outliers to identifying the largest consistent or compatible set of measurements. Joint compatibility branch and bound (JCBB), first introduced by \cite{neira2001data}, is a method that searches for the largest jointly compatible set. However, utilizing JCBB in multi-robot mapping problems can be difficult because it requires solving the graph for a combinatorial number of measurement combinations to evaluate the likelihood of each measurement given each combination of the other measurements.

Single-cluster spectral graph partitioning (SCGP), used by \cite{olson2005single}, identifies an inlier set by thresholding the eigenvector associated with the largest eigenvalue of the adjacency matrix of the underlying consistency graph. SCGP has successfully been applied to pose SLAM (\cite{olson2009recognizing}) as well as range-only SLAM (\cite{olson2005single}). CLEAR generates a graph that associates noisy measurements based on a compatibility criterion (\cite{fathian2020clear}). In CLEAR, spectral methods are used to identify the number of landmarks and can generate sets of measurements associated to a unique landmark. CLIPPER thresholds the eigenvector associated with the largest eigenvalue of the affinity matrix and shows that they can identify inliers in 99\% outlier regimes and has successfully been applied in point-cloud, line-cloud, and plane-registration problems (\cite{lusk2021clipper,lusk2022global, lusk2022graffmatch}).

\cite{mangelson2018pairwise} introduce PCM which also generates a consistency graph but transforms the problem into an instance of the maximum clique problem by showing that the maximum clique represents the largest pairwise consistent set. \cite{do2020robust} build on PCM by introducing a similarity score between measurements, turning the consistency graph into a weighted graph. They select measurements by solving the maximum edge weight clique problem, a variant of the maximum clique problem used in PCM. Work done by \cite{chen2023robust} also solves a maximum edge weight clique problem, but combines multiple consistency metrics into a single consistency function. \cite{chen2022fast} use a enforce consistency on both a spatial and temporal basis. Enforcing consistency in a spatiotemporal manner allows them to significantly reduce the time to find a consistent set of measurements. Graph-based maximum consensus registration (GMCR) was presented by \cite{gentner2023gmcr} and introduces decoupled consensus functions for scale, rotation, and translation estimation in point-cloud registration. GMCR utilizes maximum-clique algorithms to find the largest consistent set of measurements from which the variables will be estimated.

All methods described previously evaluate consistency on a pairwise basis. \cite{shi2021robin} generalize the evaluation of consistency to groups of $k$ measurements and uses the maximum $k$-core of an embedded consistency graph to quickly approximate the maximum clique in an algorithm called ROBIN. \cite{forsgren2022group} introduce G$k$CM and generates a generalized consistency graph where edges in the graph connect $k$ nodes. They also modify the maximum clique algorithms presented by \cite{pattabiraman2015fast} to find the maximum clique of a generalized graph given that no other maximum-clique algorithm over a generalized graph existed. Since then, \cite{shi2022optimal} have presented a mixed-integer linear program (MILP) that will find the maximum clique of a generalized graph.

Our contributions are a generalization of the work done by \cite{mangelson2018pairwise}, and an extension of the work by \cite{forsgren2022group}, focused on unifying the theory of consistency, applications for multi-agent systems, and decreasing run-time requirements.

\section{Problem Formulation}
\label{sec:problem_formulation}

In our factor-graph formulation of SLAM, we denote time-discretized versions of the robot trajectory by $\bvec{x}_i \in \mathrm{SE}(2)$ or $\mathrm{SE}(3)$. The factors in the graph are derived from the measurements observed by the robot and penalize estimates of the trajectory that make the observed measurement unlikely. We denote measurements that relate the variables $\bvec{x}_i$ and $\bvec{x}_j$ by $\bvec{z}_{ij}$ and call them odometric measurements if $i$ and $j$ are consecutive and loop-closure measurements if $i$ and $j$ are non-consecutive in time. The goal of pose graph SLAM is, then, to estimate the most likely value of each pose variable $\bvec{x}_i$ given the measurements $\bvec{z}_{ij}$. We can formulate the single-robot pose graph SLAM problem as the MLE
\begin{align}
  \hat{\bvec{X}} &= \underset{\bvec{X}}{\operatorname{argmax}} \: P( \bvec{Z} | \bvec{X}).
  \label{eq:mle_problem}
\end{align}
where, $\bvec{X}$ is the set of all pose variables $\bvec{x}_i$, and $\bvec{Z}$ is the set of all relative pose measurements $\bvec{z}_{ij}$.

In multi-robot SLAM, we also need to estimate the relative transformation between the local coordinate frames of the respective robots. We adopt the method presented by \cite{kim2010multiple}, which proposes the use of an anchor node for each trajectory that encodes the pose of the vehicle's local coordinate frame with respect to some global reference frame. We denote the homogeneous transformation matrix representing this offset by $T^g_a$ and represent measurements relating cross-trajectory poses by $\bvec{z}^{ab}_{ij}$, where $a$ and $b$ are robot IDs and $i$ and $j$ respectively denote which poses on robots $a$ and $b$ are being related. $T^g_a$ is an element of $\mathrm{SE}(2)$ or $\mathrm{SE}(3)$. $\bvec{z}^{ab}_{ij}$ is also often an element of $\mathrm{SE}(2)$ or $\mathrm{SE}(3)$ but can be a function of this transformation in general.

In the case of two robots, the SLAM estimation problem becomes
\begin{align}
  \hat{\bvec{X}}, \hat{\bvec{T}}^g &= \underset{\bvec{X}, \bvec{T}^g}{\operatorname{argmax}} \: P( \bvec{Z}^a, \bvec{Z}^b, \bvec{Z}^{ab} |
\bvec{X}, \bvec{T}^g),
  \label{eq:ma_mle_problem}
\end{align}
where, $\bvec{X}$ now represents the trajectories of both robots, $\bvec{Z}^{ab}$ represents the set of all cross-trajectory measurements, $\bvec{Z}^r$ represents the set of measurements local to robot $r$, and $\bvec{T}^g = \{\bvec{T}^g_a, \bvec{T}^g_b\}$. This problem can be treated as weighted, nonlinear least squares and can be solved efficiently using an array of specialized optimization libraries.

Existing methods do a good job of handling outlier measurements in the local measurement sets $\bvec{Z}^a$ and $\bvec{Z}^b$, but not in the inter-robot set $\bvec{Z}^{ab}$ since no prior estimate of the initial transformation between coordinate frames of the robots exists in general. The focus of this paper is on selecting a subset of the measurements in the inter-robot set $\bvec{Z}^{ab}$ that can be trusted. The next section outlines our approach to accomplish this.


\begin{figure}[!tb]%
  \centering%
  \includegraphics[width=0.98\columnwidth]{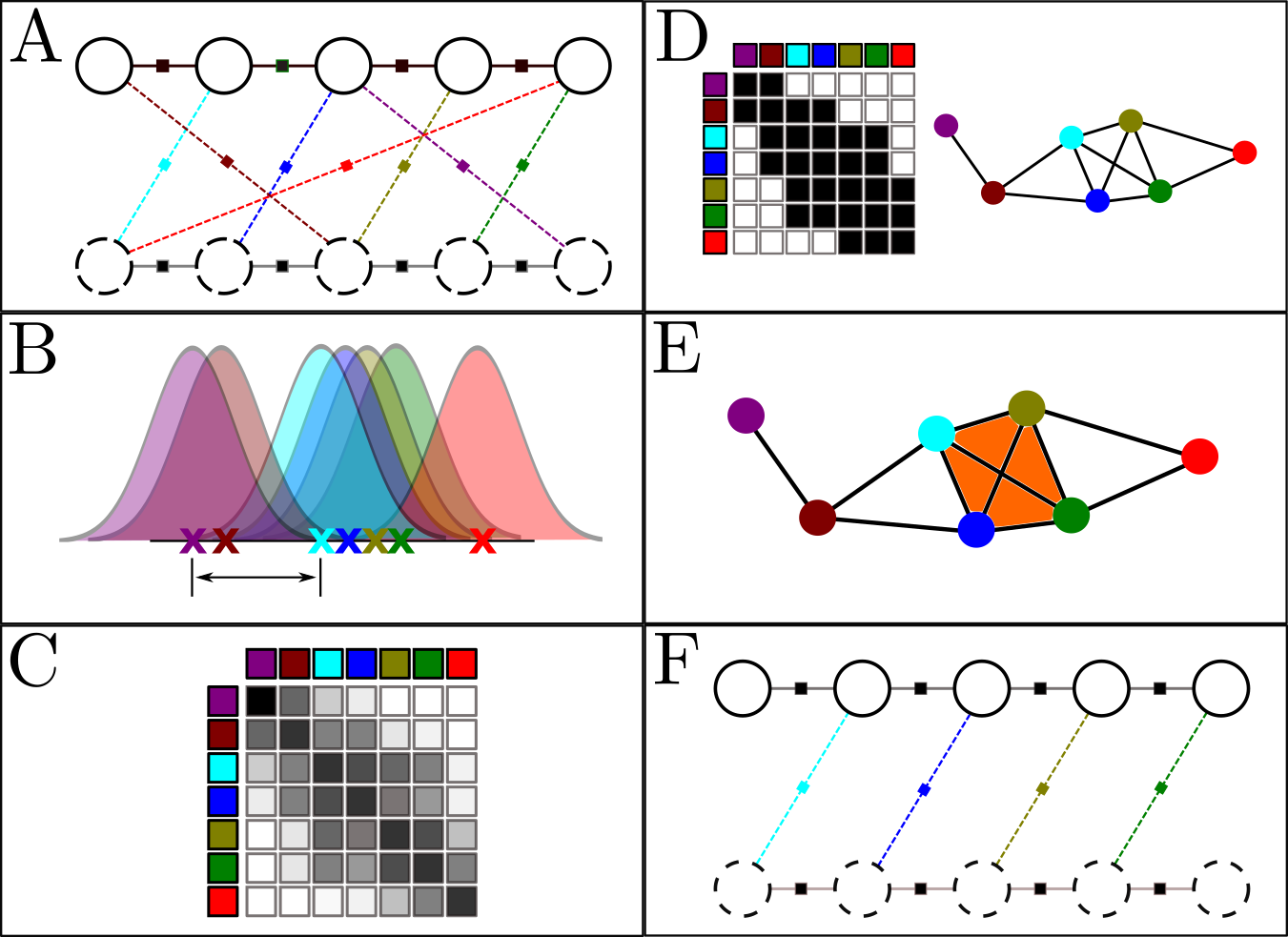}%
  \caption{An illustration of the Pairwise Consistency Maximization (PCM) algorithm for selecting consistent inter-map loop closures measurements. (A) Given two independently derived pose graphs (shown in white and black in step A) and a set of potential loop closures between them (shown by colored, dotted lines), our goal is to determine which of these inter-robot loop closures should be trusted. (B) Using a consistency metric such as Mahalanobis distance, we calculate the consistency of each pairwise combination of measurements. (C) We store these pairwise consistency values in a matrix where each element corresponds to the consistency of a pair of measurements. (D) We can transform this matrix into the adjacency matrix for a \textit{consistency graph} by thresholding the consistency and making it symmetric using the maximum consistency when associated elements across the diagonal have differing consistency values. Each node in this graph represents a measurement and edges denote consistency between measurements. Cliques in this graph are \textit{pairwise internally consistent sets}. (E) Finding the maximum clique represents finding the largest pairwise internally consistent set. (F) After determining the largest consistent set, we can robustly merge the two pose graphs using only the consistent inter-map loop closures, allowing us to reject false measurements.} 
  \label{fig:main_fig}
\end{figure}

\section{Pairwise Consistent Measurement Set Maximization}
\label{sec:pcm}

In this section, we first define a novel notion of consistency and then we use
that notion to formulate the selection of inter-robot loop-closure measurements
as a combinatorial optimization problem that finds the largest consistent set.

\subsection{Pairwise Consistency}

\label{sec:pairwise_def}
Directly determining if a measurement is an inlier or outlier from the graph
itself is unobservable as shown by \cite{carlone2014selecting}. Thus, instead of trying to
classify inlier versus outlier, we attempt to determine the maximum subset of
measurements that are internally pairwise consistent:

\begin{definition}
  A set of measurements $\bvec{\widetilde{Z}}$ is \textbf{pairwise internally consistent} with respect to a consistency metric $C$ and the threshold $\gamma$ if
\begin{align}
  C(\bvec{z}_i, \bvec{z}_j) \leq \gamma, \quad \forall \quad \bvec{z}_{i}, \bvec{z}_{j} \in \bvec{\widetilde{Z}}
  \label{eq:pairwise_consistency}
\end{align}
where, $C$ is a function measuring the consistency of measurements $\bvec{z}_i$ and $\bvec{z}_j$, and $\gamma$ is chosen a priori.
\end{definition}

This definition of consistency requires that every measurement in the set be
consistent with every other measurement in the set with respect to $C$ and
$\gamma$.

There are a variety of potential choices of a consistency metric depending on
the measurement model and the state being observed. In future sections we
present several different consistency functions for a variety of applications.
However, for the next three sections of this paper, we
assume that all inter-robot measurements are relative pose measurements with
full degrees of freedom and use the following metric based on the metric used by
\cite{olson2009recognizing}:
  \begin{align}
    \label{eq:consist_metric}
  C(\bvec{z}^{ab}_{ik}, \bvec{z}^{ab}_{jl}) &= \left|\left| (\ominus \bvec{z}^{ab}_{ik}) \oplus \hat{\bvec{x}}^a_{ij} \oplus \bvec{z}^{ab}_{jl} \oplus \hat{\bvec{x}}^b_{lk} \right|\right|_{\Sigma} \\ &\triangleq \left|\left| \epsilon_{ikjl} \right|\right|_{\Sigma_{ikjl}}
\end{align}
where, we have adopted the notation of \cite{smith1990a} to denote pose
composition using $\oplus$ and inversion using $\ominus$, $||\cdot||_{\Sigma}$
signifies the Mahalanobis distance, and the variables $\hat{\bvec{x}}^a_{ij}$
and $\hat{\bvec{x}}^b_{lk}$ are the current relative pose estimates of the
associated poses corresponding to inter-robot measurements $\bvec{z}^{ab}_{ik}$
and $\bvec{z}^{ab}_{jl}$.

This choice of metric is useful because it is both easy to compute and follows a chi-squared distribution, giving us a strategy to select the threshold $\gamma$ without knowledge of the specific dataset. The composition inside the norm of \cref{eq:consist_metric} evaluates the pose transformation around a loop and should evaluate to the identity transformation in the case of no noise (see \cite{olson2009recognizing}). With Gaussian noise, this normalized squared error follows a chi-squared distribution with degree of freedom equal to the number of degrees of freedom in our state variable. By setting $\gamma$ accordingly, we can determine if the measurements $\bvec{z}^{ab}_{ik}$ and $\bvec{z}^{ab}_{jl}$ are consistent with one another.

It should also be noted that pairwise consistency does not necessarily signify full joint consistency. It is possible that a set of measurements can be pairwise internally consistent but not jointly consistent. However, checking full joint consistency is an exponential operation and requires possibly checking every combination of measurements to evaluate their consistency. Finding the maximum-cardinality pairwise-consistent set is also exponential, but by formulating the problem in this way, we can leverage a body of literature on the maximum-clique problem in graph theory that can find or estimate the solution efficiently. In practice we observed that testing for pairwise consistency was restrictive enough to filter inconsistent measurements from typical pose graphs with full degree of freedom measurements.

\subsection{The Maximal Cardinality Pairwise Consistent Set}

\label{sec:comb_formulation}
Having this definition of pairwise internal consistency allows us to restrict our algorithm to only consider sets of measurements that are pairwise internally consistent; however, due to perceptual aliasing, we may end up with multiple subsets that are pairwise internally consistent. We need to find a way to select between these possible subsets.

The underlying assumption of our method is based on the following two initial assumptions:
\begin{assumption}
  The pose graphs are derived from multiple robots or the same robot in multiple sessions exploring the same environment.
\end{assumption}
\begin{assumption}
  The inter-robot measurements are derived from observations of that environment and the system used to derive them is not biased toward selecting incorrect measurements over correct ones.
\end{assumption}
These assumptions fit a large number of multi-robot mapping situations and are reasonable even in perceptually aliased environments whenever a place recognition system does not systematically select the perceptually aliased measurement over the correct ones.

If the above conditions are met than the following can also be safely assumed:
\begin{assumption}
  As the number of inter-robot measurements increases, the number of measurements in the correct consistent subset will grow larger than those in the perceptually aliased consistent subsets.
\end{assumption}

Our goal is, then, to efficiently find the largest consistent subset of
$\bvec{Z}^{ab}$, which we denote by $\bvec{Z}^*$.

To formalize this, we introduce a binary switch variable, $s_{u}$, for each constraint in the set $\bvec{Z}^{ab}$ and let $s_{u}$ take on the value 1 if the measurement is contained in the chosen subset and 0 otherwise. Note that there is a single $s_{u}$ for each measurement $\bvec{z}_{ij}^{ab} ~ \in ~ \bvec{Z}^{ab}$; however, for simplicity of notation, we now re-number them with the single index $u$ and denote the corresponding measurement $\bvec{z}_{ij}^{ab}$ by $\bvec{z}_u$. Letting $\bvec{S}$ be the vector containing all $s_{u}$, our goal is to find the solution, $\bvec{S}^*$, to the following optimization problem:
\begin{equation}
  \begin{gathered}
    \bvec{S}^{*} = \underset{\bvec{S}\in\{0,1\}^m}{\operatorname{argmax}} \: \norm{\bvec{S}}_0 \\
    \text{s.t.}
 \left|\left| \epsilon_{uv} \right|\right|_{\Sigma_{uv}} s_u s_v \leq \gamma ~~~ \forall~u, v,
\end{gathered}
\label{eq:comb_formulation}
\end{equation}
where, $m$ is the number of measurements in $\bvec{Z}^{ab}$, $\bvec{z}_u$ is the measurement corresponding to $s_u$, $\epsilon_{uv}$ is the associated error term corresponding to measurements $\bvec{z}_u$ and $\bvec{z}_v$, and $\Sigma_{uv}$ is the covariance matrix associated with the error $\epsilon_{uv}$. We refer to this as the PCM problem.

Once found, we can use $\bvec{S}^*$ to index into $\bvec{Z}^{ab}$ and get $\bvec{Z}^*$. This consistent subset of the measurements can then be plugged into any of the existing nonlinear least squares based solvers to merge the individual robot maps into a common reference frame. In the next section, we show how this problem can be reformulated into an equivilent problem that has been well studied.

\section{Solving PCM via Maximum Clique over Consistency Graphs}
\label{sec:pcm_mc}

In this section, we describe how to solve the PCM problem. The goal of PCM to determine the largest subset of the measurements $\bvec{Z}^{ab}$ that are pairwise internally consistent. This pairwise consistency is enforced by the $n^2$ constraints listed in \cref{eq:comb_formulation}. It is important to note that the norm on the left-hand side of the constraints does not contain any of the decision variables $s_i$. These distance measures can be calculated in pre-processing and combined into a matrix of consistency measures $\mathbf{Q}$, where each element $[\mathbf{Q}]_{uv} = q_{uv} = \left|\left|  \epsilon_{uv} \right|\right|_{\Sigma_{uv}}$, corresponds to the consistency of measurement $\bvec{z}_u$ and $\bvec{z}_v$. This process is depicted in steps B and C in \cref{fig:main_fig}.

We will now introduce the concept of a consistency graph.
\begin{definition}
  A \textbf{consistency graph} is a graph $G = \{V, \mathcal{E}\}$ where each vertex $v \in V$ represents a measurement and each edge $e \in \mathcal{E}$ denotes consistency of the vertices it connects.
\end{definition}

We can transform the matrix of consistency measures $\mathbf{Q}$ into the adjacency matrix for a consistency graph if we threshold it by $\gamma$ and make it symmetric by requiring that both $q_{uv}$ and $q_{vu}$ be less than or equal to $\gamma$ to insert an edge into the graph. An example adjacency matrix and consistency graph are shown in step D of \cref{fig:main_fig}.

A \textit{clique} in graph theory is defined as a subset of vertices in which every pair of vertices has an edge between them and the \textit{maximum clique} is the largest such subset of nodes in the graph. A clique of the consistency graph corresponds to a \textit{pairwise internally consistent set} of measurements because every measurement is pairwise consistent with every other measurement in the set. Thus, the solution to the problem defined in \cref{eq:comb_formulation} is the maximum clique of the consistency graph (see step E of \cref{fig:main_fig}).

In graph theory, the problem of finding the maximum clique for a given graph is called the maximum clique problem and is an NP-hard problem (\cite{wu2015review}). \cite{zuckerman2006linear} and \cite{feige1991approximating} show that the maximum clique problem is also hard to approximate, meaning that finding a solution arbitrarily close to the true solution is also NP-hard. Dozens of potential solutions have been proposed, each of which can be classified as either an exact or a heuristic algorithm. All of the exact algorithms are exponential in complexity and are usually based on branch and bound, while the heuristic algorithms often try to exploit some type of structure in the problem, making them faster, but not guaranteeing the optimal solution (see \cite{wu2015review}).

In 2015, \cite{pattabiraman2015fast} proposed a method that aggressively prunes the search tree and is able to find maximum-clique solutions for large sparse graphs relatively quickly. They present both an exact algorithm as well as a heuristic version that can be used when the exact algorithm becomes intractable. Though our method could theoretically use any one of the proposed maximum clique algorithms, we selected the one proposed by \cite{pattabiraman2015fast} because of its simplicity, parallelizablity, and open-source implementation.

\section{Pairwise Consistency Maximization Evaluation}
\label{sec:pcm_eval}

In this section, we evaluate the performance of PCM on a variety of synthetic and real-world datasets. For comparison, we implemented single cluster graph partitioning (SCGP) (\cite{olson2005single}), dynamic covariance scaling (DCS) (\cite{agarwal2013robust}), and random sample consensus (RANSAC) (\cite{hartley2003multiple}).

We implemented SCGP as described in \cite{olson2005single}, with the exception of using an off the shelf eigen-factorization library as opposed to the power method for simplicity. We implemented DCS as described in the original paper with $\phi = 5$.

We implemented RANSAC by iteratively selecting a single, random inter-map measurement and evaluating the likelihood of the other measurements given the model estimated from the sampled measurement. Because the processing time for this evaluation is so low (given that the Mahalanobis distance evaluations were performed in pre-processing), we exhaustively iterate through all the measurements and evaluate the likelihood of the other measurements with respect to it in turn. We then return the set of measurements that are likely given the sampled point with the largest support. As explained in \cref{sec:2d}, RANSAC is especially sensitive to the likelihood threshold and does not check pairwise consistency.

For PCM, we present results using the exact maximum clique algorithm (PCM-Exact), as well as the heuristic algorithm (PCM-HeuPatt) as explained by \cite{pattabiraman2015fast}.

\subsection{Simulated 1D World}


\begin{figure}[!tb]%
  \centering%
  \includegraphics[width=0.98\columnwidth,trim={0.2cm 0.08cm 0.1cm 0cm},clip]{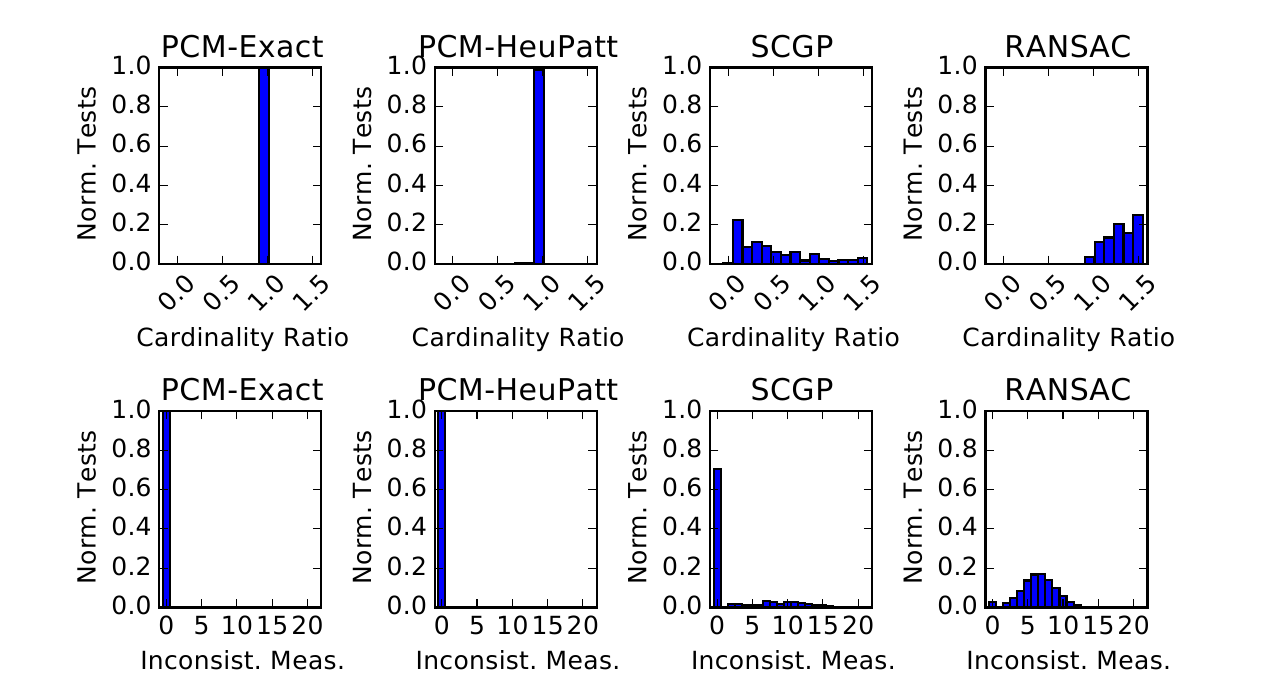}%
  \caption{Histograms that evaluate how well PCM, SCGP (\cite{olson2005single}), and RANSAC (\cite{hartley2003multiple}) approximate the combinatorial maximum pairwise consistent set in \cref{eq:comb_formulation}. The first row of histogram plots shows the size of the measurement set as compared to the maximum consistent set size. The second row of histograms shows the number of inconsistent pairs returned with respect to the set $\gamma$ threshold on Mahalanobis distance.} 
  \label{fig:1D_Comp}%
\end{figure}

First, we simulated a one dimensional world where the robot has a single state variable, $x$, and receives measurements that are direct observations of that state. We simulate inlier measurements by drawing multiple samples from a Gaussian distribution with a fixed variance and mean $x$. We simulate both random and perceptually aliased outliers by drawing multiple samples from a single Gaussian with fixed mean and variance and several others from individual Gaussians with random means and variances. We assume the variances are known and are used when computing Mahalanobis distance.

\subsubsection {Comparison with Combinatorial}
\label{sec:comp_comb}

For this first experiment, we compare how well PCM-Exact and PCM-HeuPatt approximated the combinatorial gold standard in \cref{eq:comb_formulation}. We generated 100,000 sample worlds. On each of these samples, we estimated the pairwise consistent set using the combinatorial solution as well as PCM-Exact, PCM-HeuPatt, SCGP , and RANSAC .

\cref{fig:1D_Comp} shows a comparison between these four methods with respect to the combinatorial solution. Both PCM methods enforce consistency of the returned measurements. PCM-Exact returns the same number of points as the combinatorial solution 100 percent of the time, while PCM-HeuPatt returns the same number of points 98.97 percent of the time. SCGP varies significantly in both the number of points returned and the consistency of those measurements. RANSAC also sometimes returns more or less points than the combinatorial solution and also fails to enforce measurement consistency.

Interestingly, RANSAC is especially dependent on threshold value. The threshold value for RANSAC is centered around a single point and thus is not the same as the threshold value for PCM. If the value is set too high, the number of inconsistent measurements increases. If it is set too low, the total number of returned measurements decreases below the optimal. In \cref{fig:1D_Comp}, RANSAC's threshold is set arbitrarily to show a single snapshot.

\subsubsection {Timing Comparison}

\begin{figure}[!tb]%
  \centering%
  \includegraphics[width=\columnwidth,trim={0.3cm 0.45cm 0.4cm 0cm},clip]{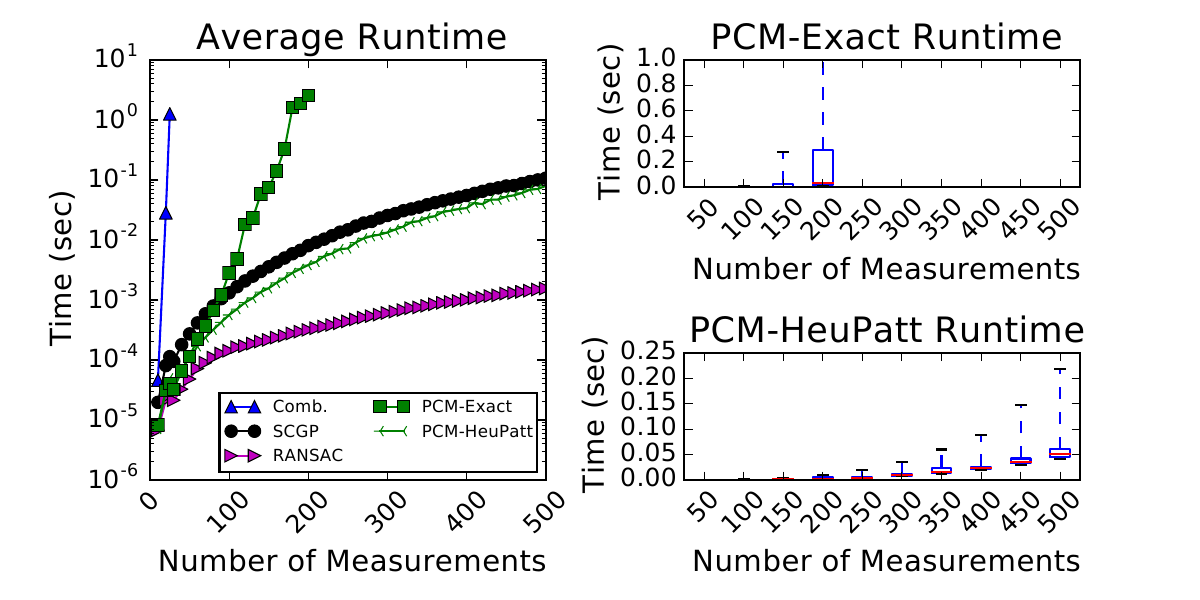}%
  \caption{A plot of the evaluation times of the different methods versus the number of measurements being tested. The combinatorial solution takes exponential time and PCM-Exact takes exponential time in the worst case, while the other methods are polynomial in the number of measurements. (This excludes the time to estimate the distance matrix $\mathbf{Q}$, which is required for all methods.)} 
  \label{fig:timing}%
\end{figure}

We also used this 1D-world to evaluate the timing characteristics of the different algorithms. To test this, we generated 500 sample worlds each for an increasing number of measurement points. The results are shown in \cref{fig:timing}. Note, these timing results can be significantly improved through parallelization.

\begin{figure}[!tb]%
  \centering%
  \includegraphics[width=0.98\columnwidth,trim={0.6cm 0.7cm 0.6cm 0cm},clip]{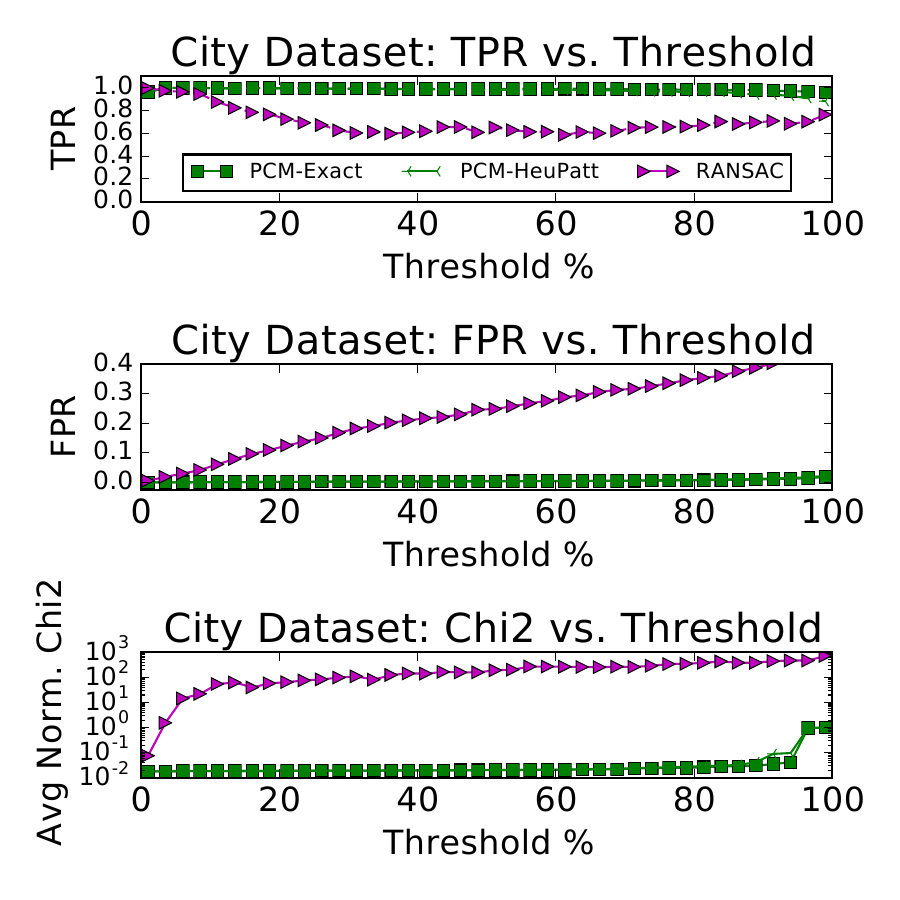}%
  \caption{The true positive rate (TPR = TP / (TP + FN)), false positive rate (FPR = FP / (FP + TN)), and average normalized chi-squared value (Chi2) of PCM-Exact, PCM-Heu, and RANSAC versus the threshold value $\gamma$. The TPR and FPR can be thought of as the probability of getting a true positive  or a false positive. The Chi2 value should be close to zero if the measurements in the graph are consistent.} 
  \label{fig:2d_tpr_fpr}%
\end{figure}

\subsection{Synthetic 2D Comparison}
\label{sec:2d}

\begin{table*}[tbh!]
  \addtolength{\tabcolsep}{-5pt}
  \centering
  \caption{Results from using DCS\iffalse (\cite{agarwal2013robust}) \fi, SCGP\iffalse (\cite{olson2005single}) \fi, RANSAC\iffalse (\cite{hartley2003multiple}) \fi(with two different thresholds), and PCM to robustly merge maps generated from a synthetic city dataset. These results are a summary of runs on 81 different generated datasets.  We evaluated the mean squared error (MSE) of the two graphs with respect to the non outlier case (NO-OUT). The worst results for each metric are shown in {\color{red} red}, the best are shown in {\textbf{\color{blue}blue}}, and the second best shown in {\bf BOLD}.}
  \scalebox{0.9}{
 \begin{tabular}{|l|cc|cc|cc|cc|cc|c|}
   \hline
   & \multicolumn{2}{c|}{Trans. MSE ($m^2$)} & \multicolumn{2}{c|}{Rot. MSE} &  \multicolumn{2}{c|}{Residual} & \multicolumn{2}{c|}{Inliers} & \multicolumn{2}{c|}{Chi2 Value} & \multicolumn{1}{c|}{Eval Time (sec)} \\ \cline{2-12}
   & \multicolumn{1}{c|}{Avg} & \multicolumn{1}{c|}{Std} & \multicolumn{1}{c|}{Avg} & \multicolumn{1}{c|}{Std} & \multicolumn{1}{c|}{Avg} & \multicolumn{1}{c|}{Std} & \multicolumn{1}{c|}{TPR} & \multicolumn{1}{c|}{FPR} & \multicolumn{1}{c|}{Avg} & \multicolumn{1}{c|}{Std} & \multicolumn{1}{c|}{Avg} \\ \hline
   \hline
   NO-OUT & 0.0 & 0.0 & 0.0 & 0.0 & 32.320 & 0.117 & 1.0 & 0.0 &  N/A & N/A & N/A  \\ \hline
 \hline
   DCS & {\color{red}183077.917} & {\color{red}1194931.105} & {\color{red}4.169} & {\color{red}3.285} & {\bf\color{blue}31.687} & {\bf\color{blue}0.507} & {\color{red}0.0} & {\bf\color{blue}0.0} & {\bf\color{blue}0.013} & {\bf\color{blue}$<$ 0.001} & N/A  \\ \hline
   SCGP & 623.278 & 1278.493 & 0.648 & 1.535 & {\color{red}237385.743} & {\color{red}894303.187} & 0.668 & {\color{red}0.051} & {\color{red}96.734} & {\color{red}364.427} & {\color{red}0.006} \\ \hline
   RANSAC-1\% & {\bf5.688} & {\bf21.976} & {\bf0.009} & {\bf0.040} & 185.190 & 587.590 &{\bf\color{blue}0.998} & 0.006 & 0.076 & 0.239 & {\bf\color{blue}$<$ 0.001} \\ \hline
   RANSAC-3.5\% & 183.150 & 636.441 & 0.236 & 0.791 & 3807.570 & 18478.340 & 0.974 & 0.019 & 1.552 & 7.530 & {\bf\color{blue}$<$ 0.001} \\ \hline
   PCM-Exact-11\% & {\bf\color{blue}0.276} & {\bf\color{blue}1.537} & {\bf\color{blue}$<$ 0.001} & {\bf\color{blue}0.003} & {\bf45.057} & {\bf105.385} & {\bf0.997} & {\bf0.001} & {\bf0.018} & {\bf0.043} & {\bf\color{blue}$<$ 0.001} \\ \hline
   PCM-HeuPatt-11\% & {\bf\color{blue}0.276} & {\bf\color{blue}1.537} & {\bf\color{blue}$<$ 0.001} & {\bf\color{blue}0.003} & {\bf45.057} & {\bf105.385} & {\bf0.997} & {\bf0.001} & {\bf0.018} & {\bf 0.043} & {\bf\color{blue}$<$  0.001}
   \\ \hline
 \end{tabular}
 }
\label{table:2D_comp}
\end{table*}
\begin{figure*}[tbh!]%
  \centering%
  \subfigure{%
    \includegraphics[width=0.40\columnwidth,trim={0.05cm 0.3cm 0.05cm 0cm},clip]{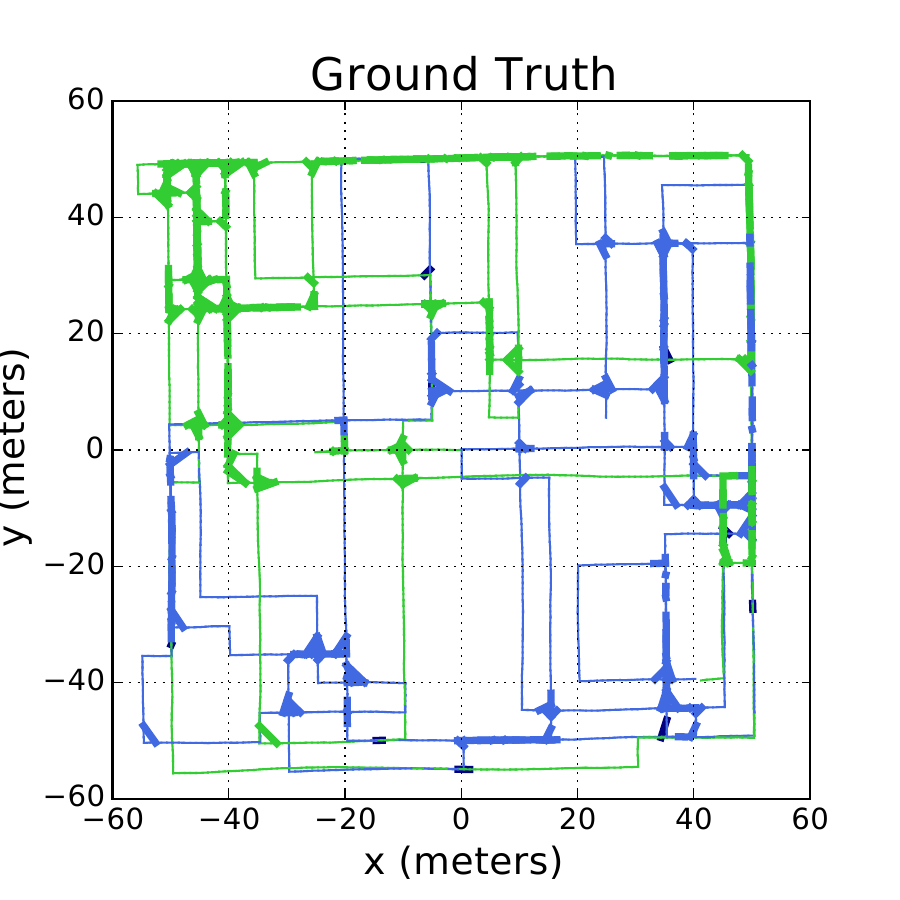}%
    \label{sfig:gt}%
  }%
  \subfigure{%
    \includegraphics[width=0.40\columnwidth,trim={0.05cm 0.3cm 0.05cm 0cm},clip]{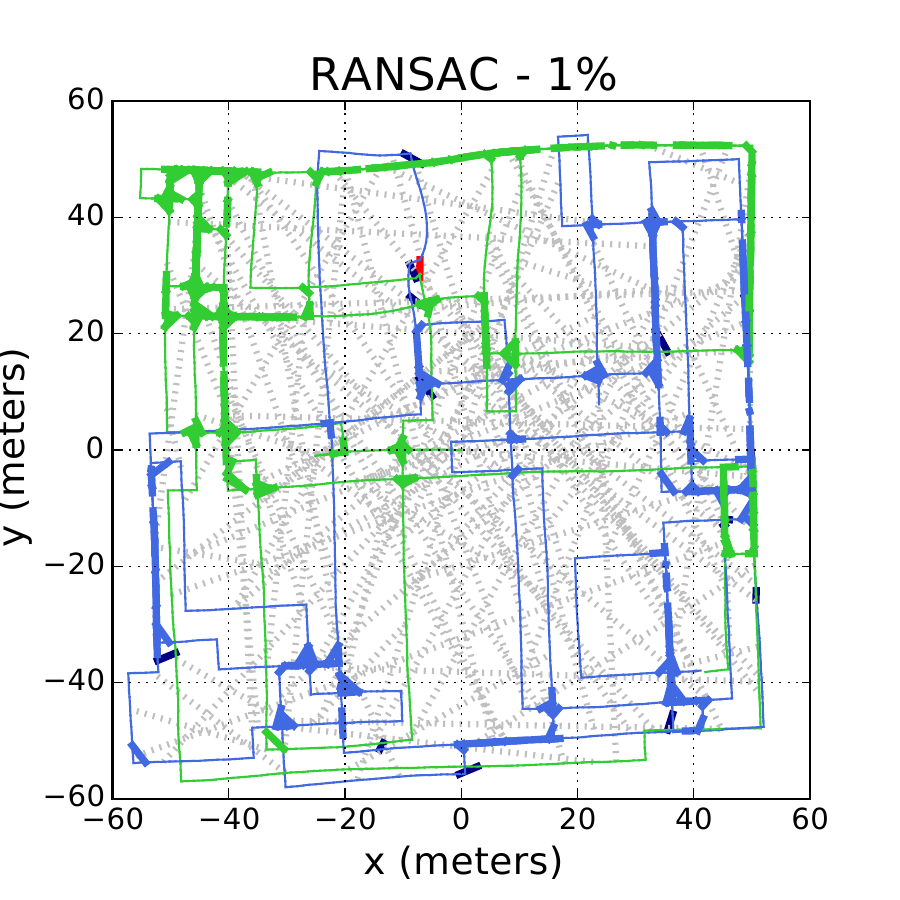}%
    \label{sfig:ransac1}%
  }%
  \subfigure{%
    \includegraphics[width=0.40\columnwidth,trim={0.05cm 0.3cm 0.05cm 0cm},clip]{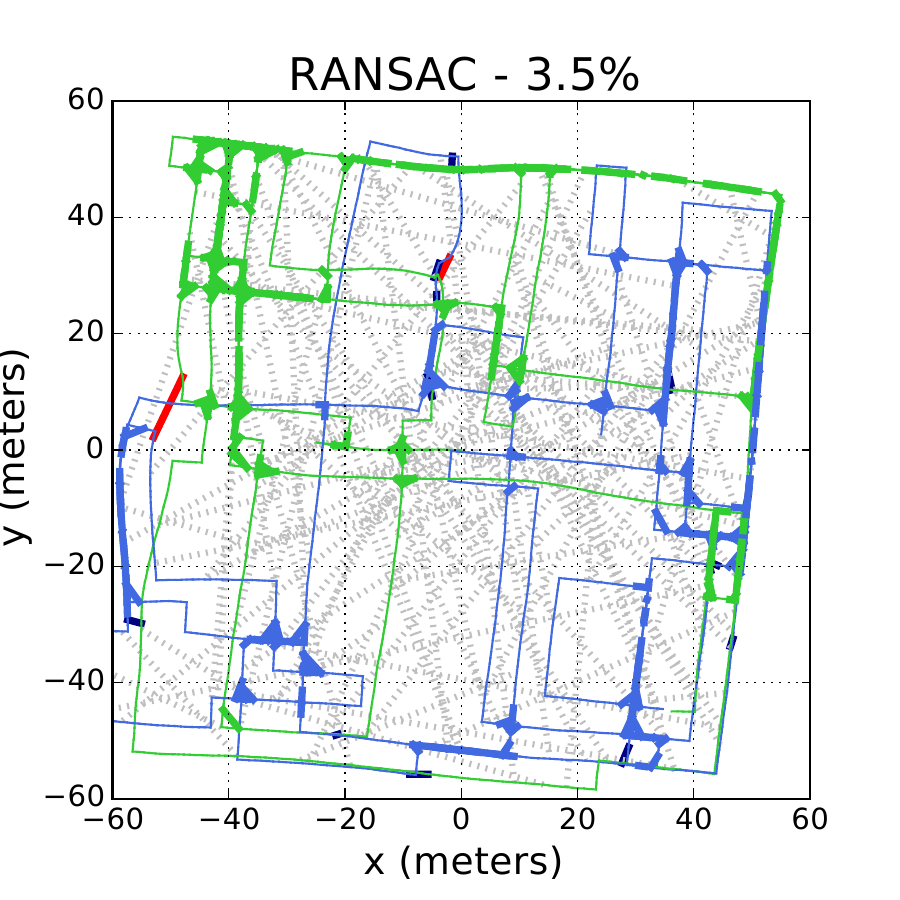}%
    \label{sfig:ransac3p5}%
  }%
  \subfigure{%
    \includegraphics[width=0.40\columnwidth,trim={0.05cm 0.3cm 0.05cm 0cm},clip]{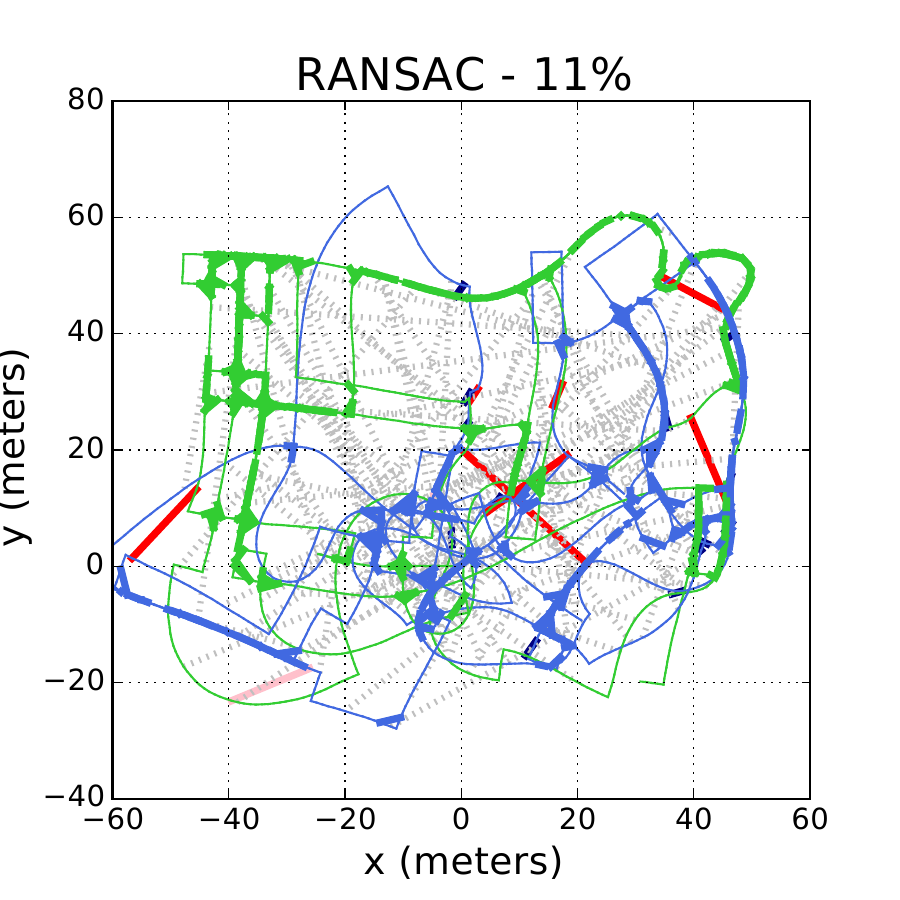}%
    \label{sfig:ransac11}%
  }%
  \subfigure{%
    \includegraphics[width=0.40\columnwidth,trim={0.05cm 0.3cm 0.05cm 0cm},clip]{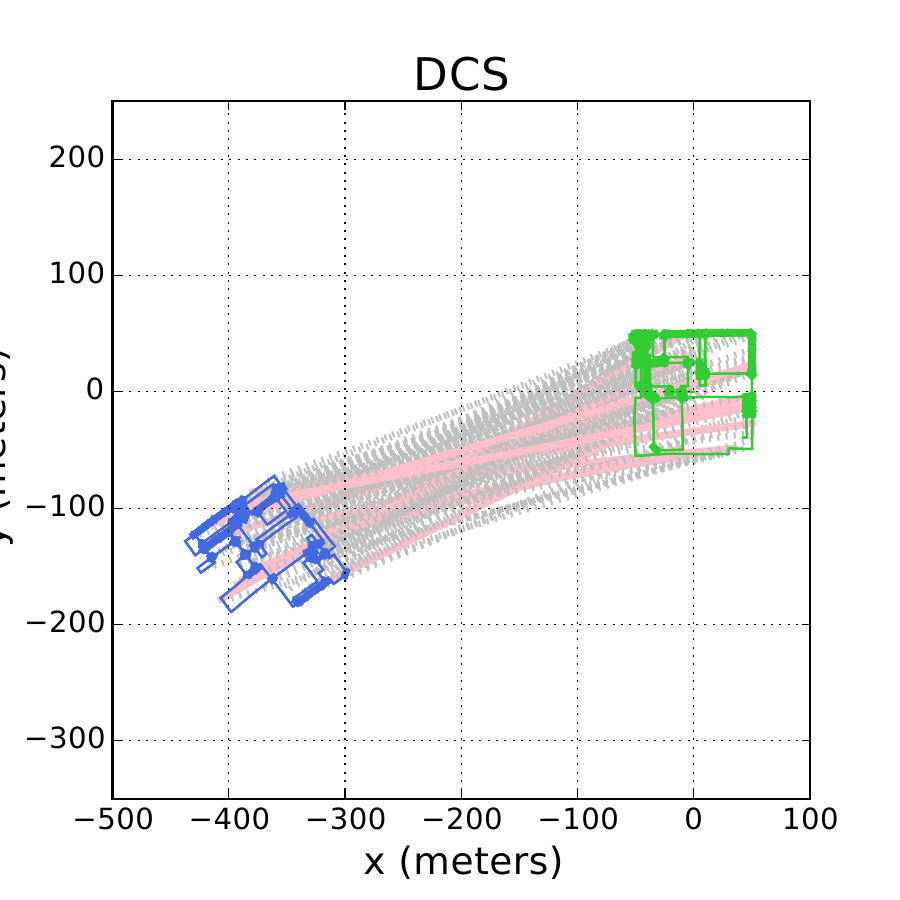}%
    \label{sfig:dcs}%
  } \\%
  \vspace*{-0.5cm}%
  \subfigure{%
    \includegraphics[width=0.40\columnwidth,trim={0.05cm 0.4cm 0.05cm 0.5cm},clip]{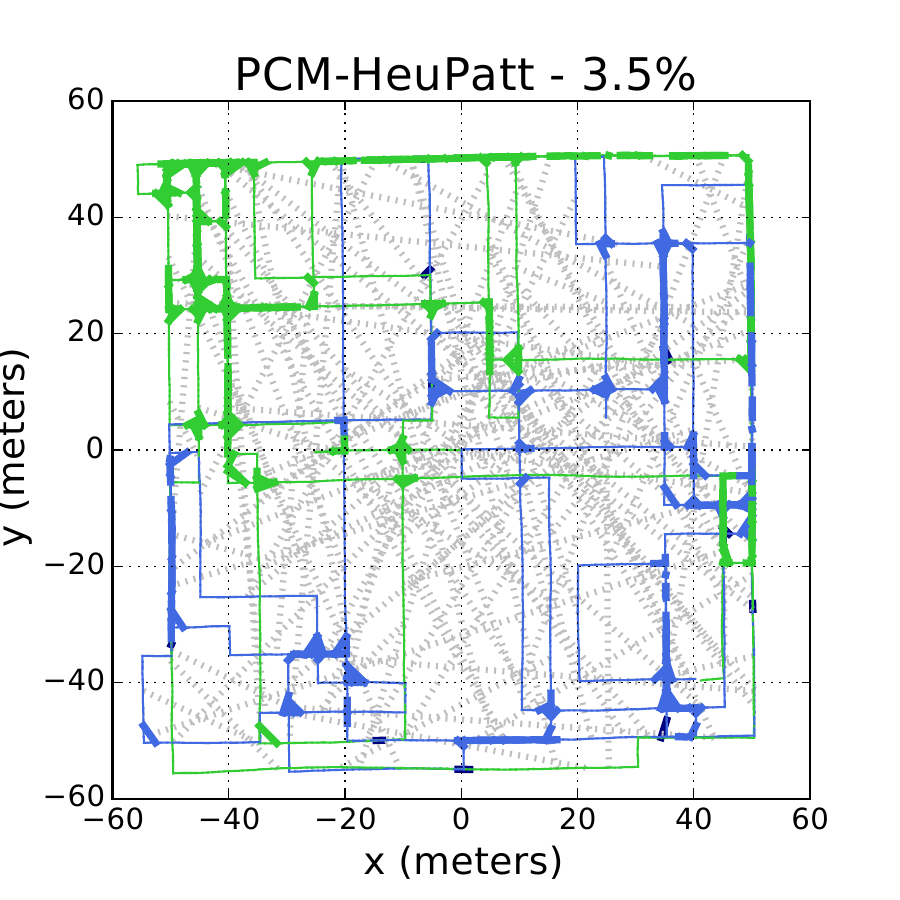}%
    \label{sfig:heu3p5}%
  }%
  \subfigure{%
    \includegraphics[width=0.40\columnwidth,trim={0.05cm 0.4cm 0.05cm 0.5cm},clip]{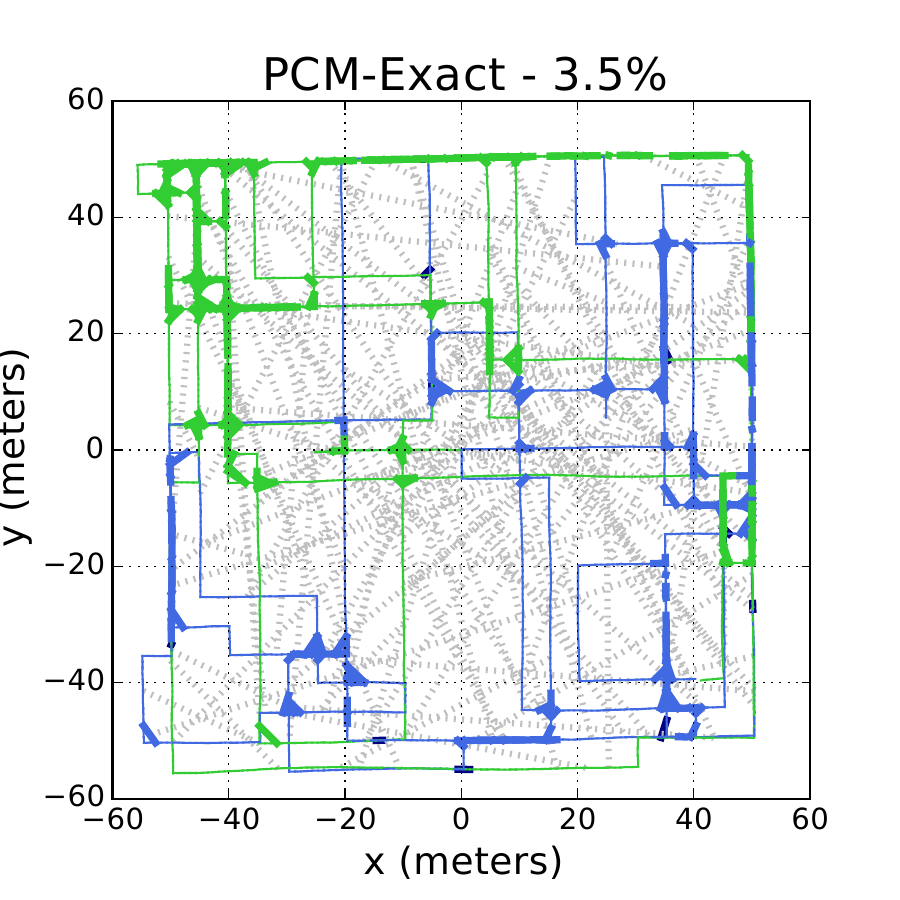}%
    \label{sfig:exact3p5}%
  }%
  \subfigure{%
    \includegraphics[width=0.40\columnwidth,trim={0.05cm 0.4cm 0.05cm 0.5cm},clip]{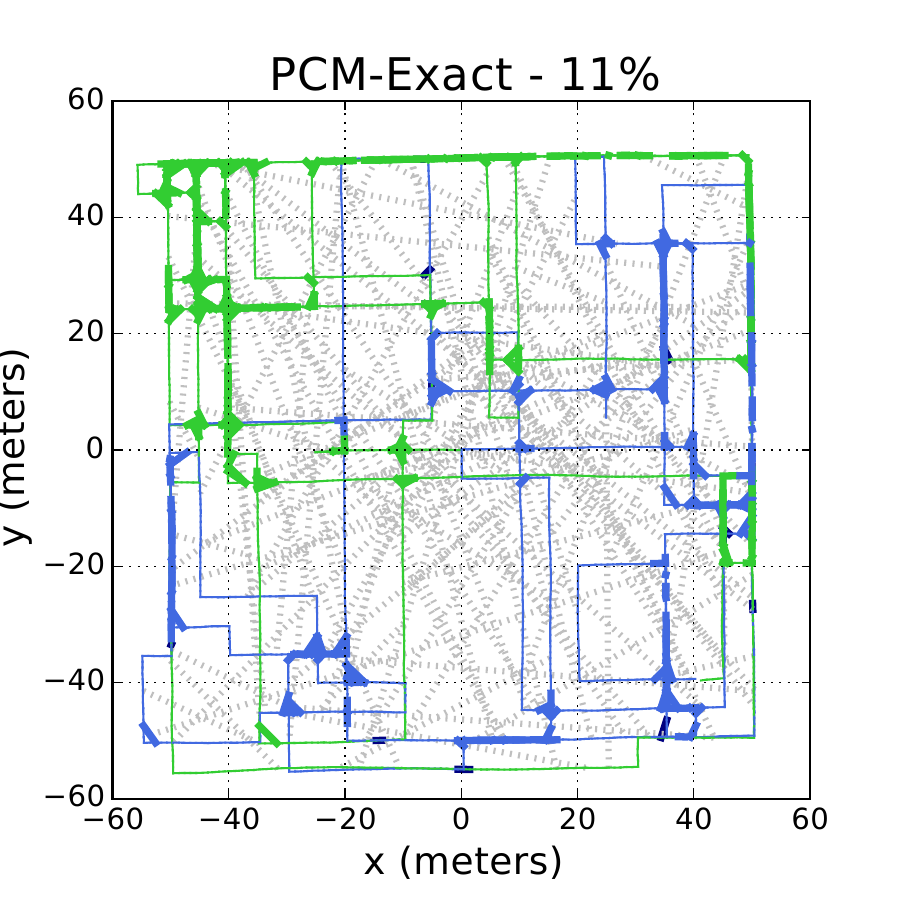}%
    \label{sfig:exact11}%
  }%
  \subfigure{%
    \includegraphics[width=0.40\columnwidth,trim={0.05cm 0.4cm 0.05cm 0.5cm},clip]{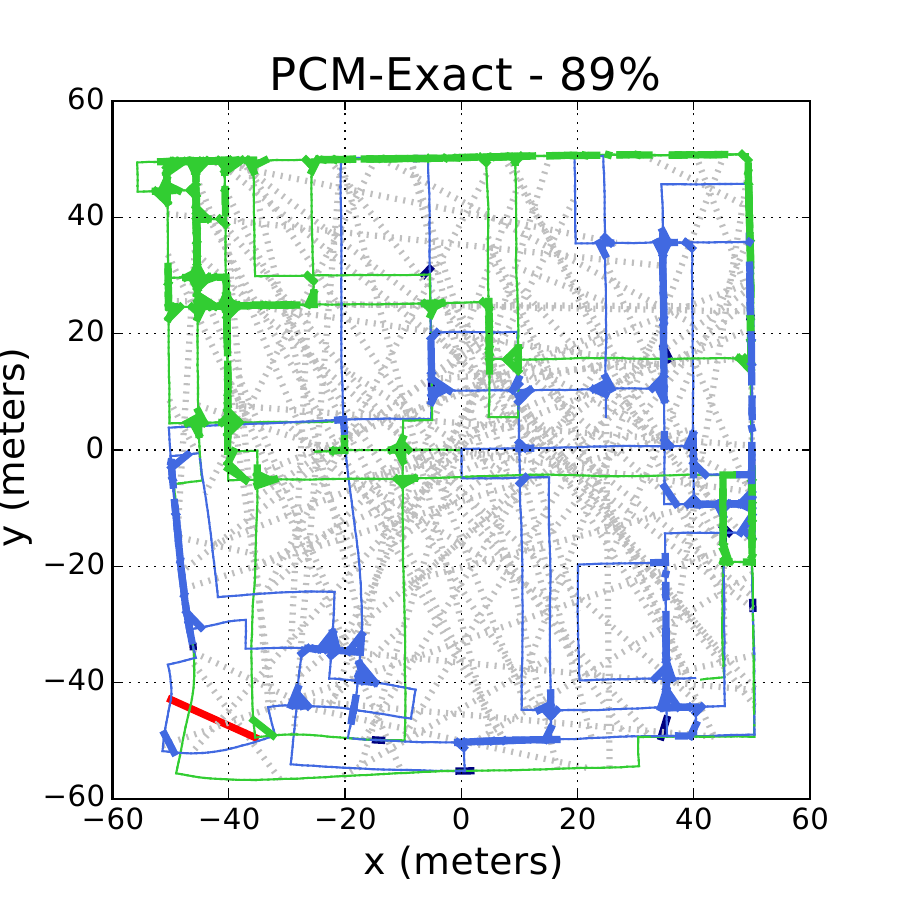}%
    \label{sfig:exact91}%
  }%
  \subfigure{%
    \includegraphics[width=0.40\columnwidth,trim={0.05cm 0.3cm 0.05cm 0cm},clip]{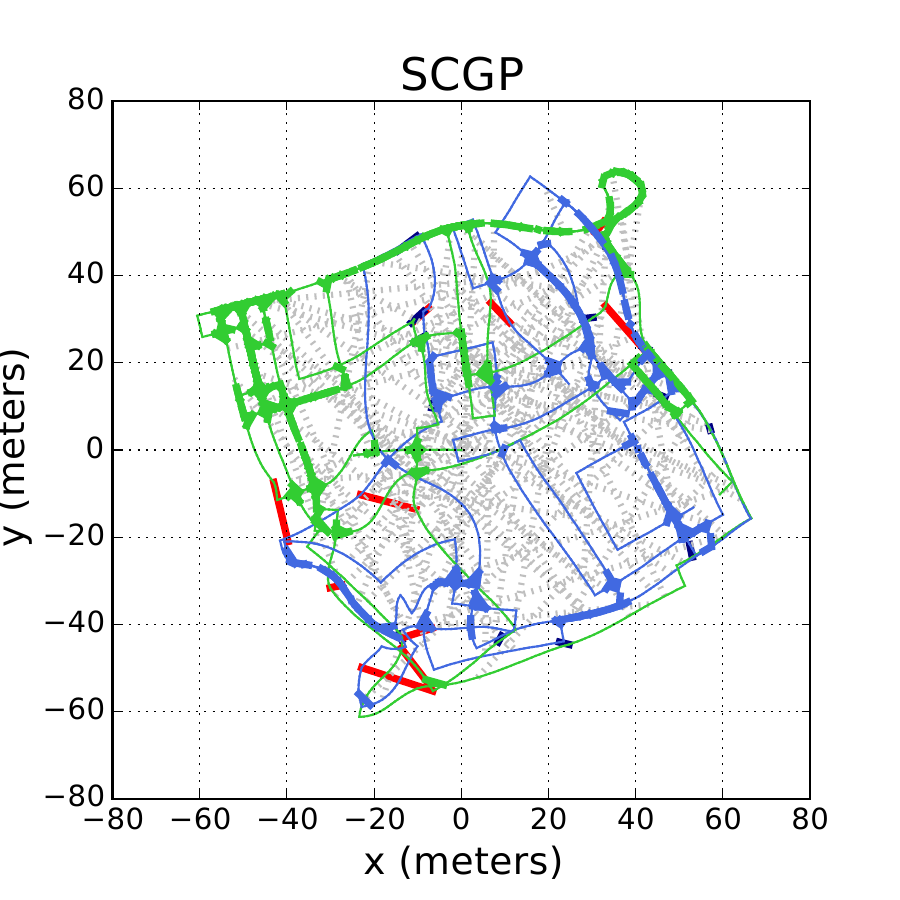}%
    \label{sfig:scgp}%
  }%
  \caption{Example plots of the maps estimated by PCM-Exact, PCM-HeuPatt, RANSAC, DCS, and SCGP for one of the generated city datasets.  Correctly labeled inlier factors are shown in bold dark blue with correctly disabled outliers shown as dotted gray. Accepted outliers are shown in bold red with disabled inliers shown in pink. }
  \vspace*{4mm}
  \label{fig:2D_examples}
\end{figure*}

To test our method's accuracy and consistency on a full SLAM dataset, we took a portion of the City10000 dataset released with iSAM (\cite{kaess2008isam}) and split it to form two separate robot trajectories. After removing all factors connecting the two graphs, we generated 81 different versions of this dataset by randomly selecting a subset of the true loop closures between the two graphs to be used as inliers, as well as randomly adding outlier loop closures to the graph. As before, some of the outliers are internally consistent to simulate perceptual aliasing and some are generated randomly with random mean and covariance. In this experiment, the number of inlier loop closures was 15, there were two groups of 5 perceptual aliased outliers, and the number of random outliers was 90.

\subsubsection {Parameter Sweep}
\label{sec:sweep}

Because RANSAC is significantly dependent on the threshold value set, we ran a parameter sweep for the likelihood threshold over all 81 datasets. \cref{fig:2d_tpr_fpr} summarizes this experiment. The true positive rate (TPR) and false positive rate (FPR) of PCM is relatively unaffected by the choice of the threshold parameter as long as it is less than about 85 percent. RANSAC, on the other hand, has a different FPR for each threshold selected and never has an FPR of zero. This is because PCM conservatively evaluates the consistency of each measurement and determines consistency of a group of measurements as a whole, while RANSAC selects the largest set of measurements that are likely given a single randomly selected measurement. The last plot shows the average normalized chi-squared value of the residual for the entire graph after solving with the selected factors. This value should be close to zero if the graph is consistent.

The results show that PCM does significantly better at restricting the set of measurements to those that are consistent with one another, decreasing the likelihood of getting a false measurement. This is essential because of the extreme susceptibility of SLAM to false loop closures. PCM-HeuPatt is also almost indistinguishable from PCM-Exact.

\subsubsection {Accuracy Analysis}

To evaluate the accuracy of PCM, we compared its performance on all 81 datasets to SCGP, RANSAC (using the two second to lowest thresholds from \cref{fig:2d_tpr_fpr}), and dynamic covariance scaling (\cite{agarwal2013robust}) or DCS. $\gamma$ for both PCM-Exact and PCM-HeuPatt was set so that it corresponded to the equivalent of 11\% likelihood.

\cref{table:2D_comp} gives an overall summary of the results. We used the Mean Squared Error (MSE) of the trajectory of the two graphs (with respect to the no-outlier case), the residual, and the normalized chi squared value of the nonlinear least squares solver as metrics to evaluate the solution accuracy. The rotation MSE was calculated via $\epsilon_{\text{rot}} = \frac{1}{n} \sum_i \left| \left|\log(R_{i\_\text{true}}^\top R_{i\_\text{est}}) \right| \right|_F$ over each pose $i$ and the translation MSE was calculated in the normal manner. For this experiment all MSE were calculated with respect to the absolute trajectory value.

PCM has the lowest trajectory MSE, and DCS has the lowest residual. Note that DCS also has the highest trajectory MSE, which is as expected. DCS seeks to minimize the least-squares residual error and depends on a good initialization to determine what measurements are consistent enough to not be turned off. Without this initialization, DCS has no reason to believe that the inter-map factors are not outliers and thus turns off all the inter-map factors in the graph.

Once given the matrix $\bvec{Q}$, RANSAC and both PCM methods take about the same amount of time to find the consistency set. The average time to estimate the Mahalanobis distances without the use of analytical Jacobians, parallelization, and incremental updates was 70.8s.

\cref{fig:2D_examples} shows example plots of the estimated maps. Both SCGP and RANSAC have trouble disabling all inconsistent measurements. PCM-HeuPatt accurately approximates PCM-Exact and both do well at disabling inconsistent measurements. When PCM does accept measurements not generated from the true distibution, they are still consistent with the uncertainty of the local graphs.

\begin{table*}[t!]
   \caption{Results from using DCS\iffalse(\cite{agarwal2013robust})\fi, SCGP\iffalse(\cite{olson2005single})\fi, RANSAC\iffalse(\cite{hartley2003multiple})\fi (with two different thresholds), and PCM to robustly merge segments extracted from two sessions of the NCLT dataset\iffalse(\cite{ncarlevaris-2015a})\fi. NO-OUT corresponds to a version with none of the measurements labeled as outliers.  We evaluated the mean squared error (MSE) of the two graphs with respect to the ground truth. The worst results for each metric are shown in {\color{red} red} and the best are shown in {\textbf{\color{blue}blue}}. \label{table:nclt_comp}}
   \centering
   \scalebox{0.9}{
   \begin{tabular}{|l|cc|cc|c|cc|c|c|}

     \hline
 & \multicolumn{2}{c|}{Rel. Pose MSE} & \multicolumn{2}{c|}{Traj. MSE} &  \multicolumn{1}{c|}{Residual} & \multicolumn{2}{c|}{Inliers} & \multicolumn{1}{c|}{Chi2} & \multicolumn{1}{c|}{Evaluation} \\ \cline{2-5} \cline{7-8}
 & \multicolumn{1}{c|}{Trans. ($m^2$)} & \multicolumn{1}{c|}{Rot.} & \multicolumn{1}{c|}{Trans. ($m^2$)} & \multicolumn{1}{c|}{Rot.} & Error & \multicolumn{1}{c|}{TP} & \multicolumn{1}{c|}{FP} & Value & Time (sec) \\ \hline
     \hline
     NO-OUT & 455.4763 & 0.0308 & 0.0501 & 0.0005 & 765.072 & 10 & 0 & 0.3428 & N/A \\ \hline
     \hline
     DCS & {\color{red}206782.2303} & 0.7154 & 0.0502 & {\bf\color{blue}0.0005} & {\bf\color{blue}724.061} & {\color{red}0} & {\bf\color{blue}0} & {\bf\color{blue}0.2568} & N/A \\ \hline
     SCGP & 522.2352 & {\bf\color{blue}0.0162} & 0.0502 & {\bf\color{blue}0.0005} & 748.351 & 3 & {\bf\color{blue}0} & 0.3417 & {\color{red}0.0021} \\ \hline
     RANSAC - 1\% & 1244.3818 & 0.0697 & 0.1036 & 0.0015 & 4228.21 & {\bf\color{blue}10} & 6 & 1.8643 & {\bf\color{blue}$<$ 0.0001} \\ \hline
     RANSAC - 3.5\% & 13507.7032 & {\color{red}17.4156} & {\color{red}0.1146} & {\color{red}0.0040} & {\color{red}7457.54} & {\bf\color{blue}10} & {\color{red}7} & {\color{red}3.2795} & 0.0001 \\ \hline
     PCM-HeuPatt & {\bf\color{blue}386.6876} & 0.0245 & {\bf\color{blue}0.0501} & {\bf\color{blue}0.0005} & 817.803 & {\bf\color{blue}10} & 3 & 0.3635 & 0.0001 \\ \hline
   \end{tabular}
   }
     \vspace{-3mm}
   \end{table*}

\begin{figure*}[tb!]%
  \centering%
  \subfigure{%
    \includegraphics[width=0.66\columnwidth,trim={0cm 0.3cm 0.2cm 0cm},clip]{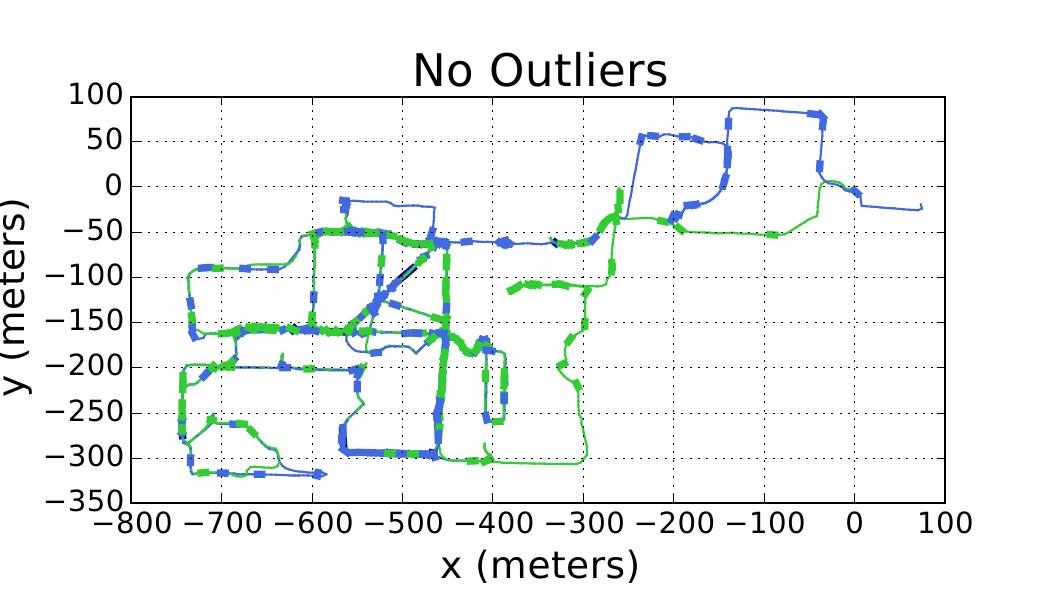}%
    \label{sfig:nclt_gt}%
  }%
  \subfigure{%
    \includegraphics[width=0.66\columnwidth,trim={0cm 0.3cm 0.2cm 0cm},clip]{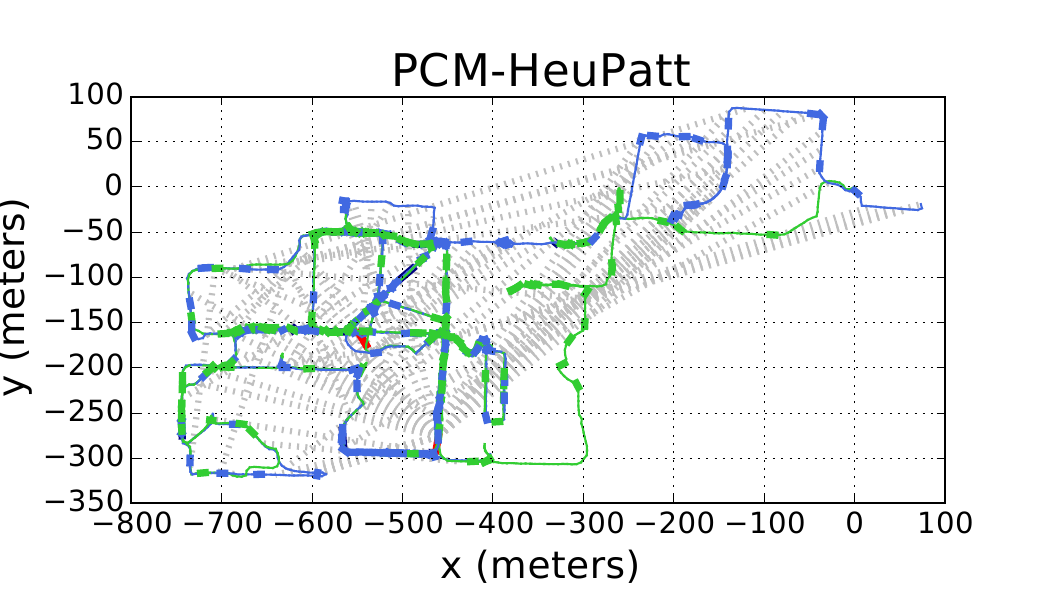}%
    \label{sfig:nclt_pcm}%
  }%
  \subfigure{%
    \includegraphics[width=0.66\columnwidth,trim={0cm 0.3cm 0.2cm 0cm},clip]{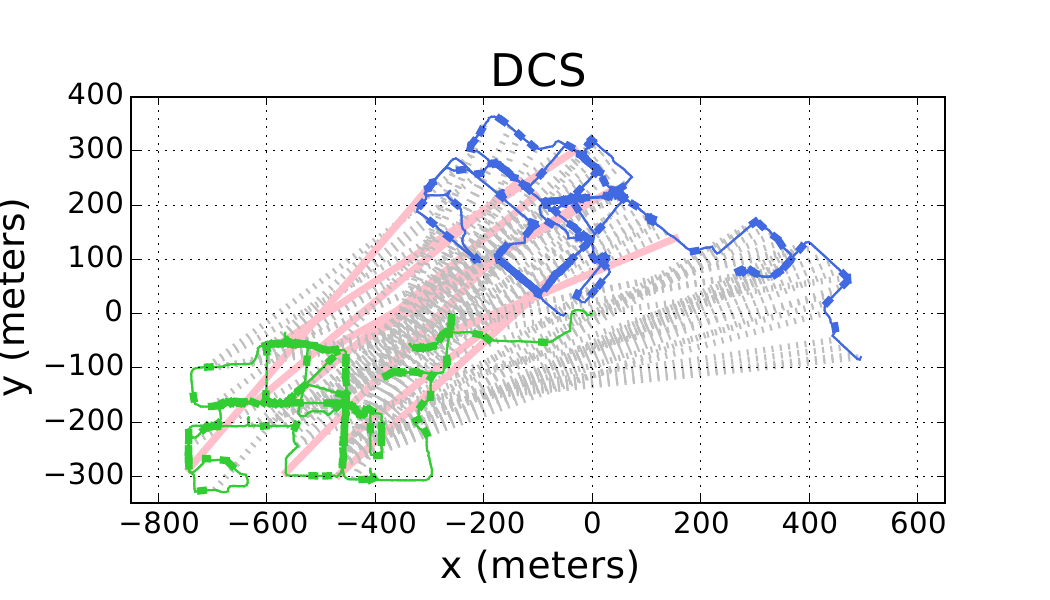}%
    \label{sfig:nclt_dcs}%
  } \\%
  \vspace*{-0.3cm}%
  \subfigure{%
    \includegraphics[width=0.66\columnwidth,trim={0cm 0.3cm 0.2cm 0.5cm},clip]{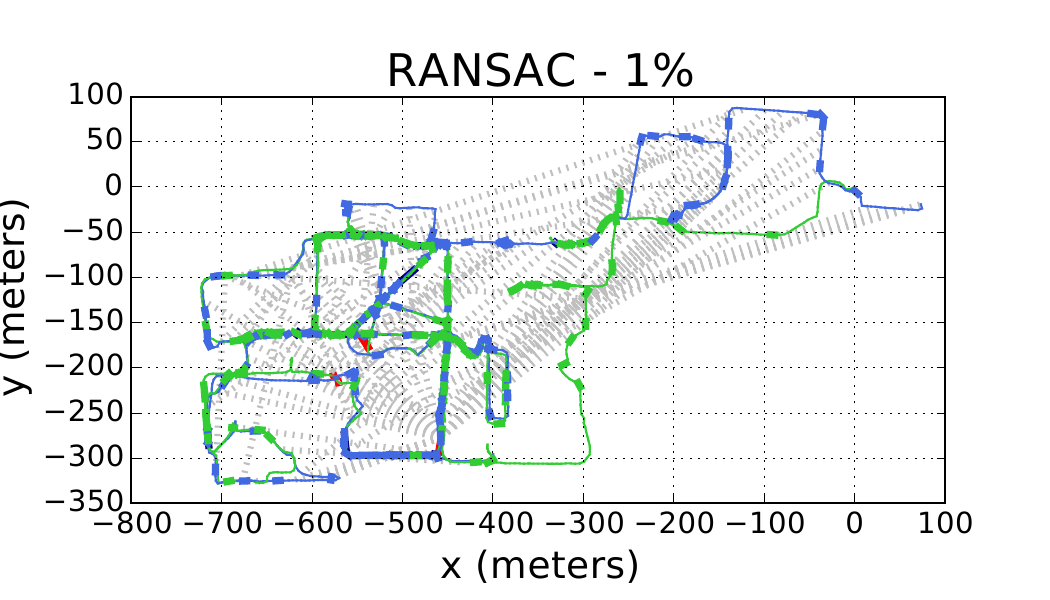}%
    \label{sfig:nclt_ransac}%
  }%
  \subfigure{%
    \includegraphics[width=0.66\columnwidth,trim={0cm 0.3cm 0.2cm 0.5cm},clip]{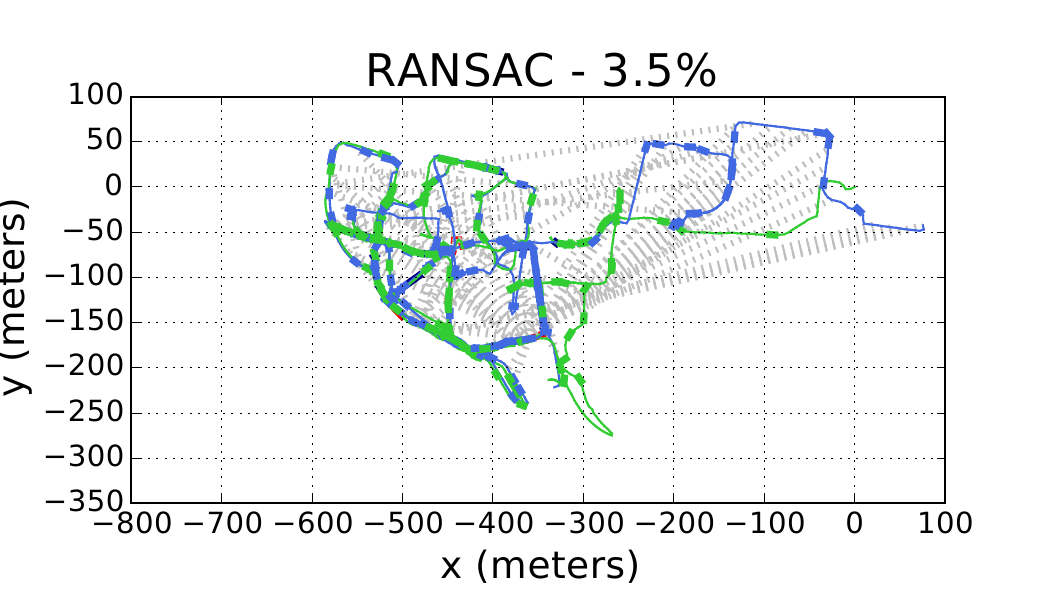}%
    \label{sfig:nclt_ransac}%
  }%
  \subfigure{%
    \includegraphics[width=0.66\columnwidth,trim={0cm 0.3cm 0.2cm 0.5cm},clip]{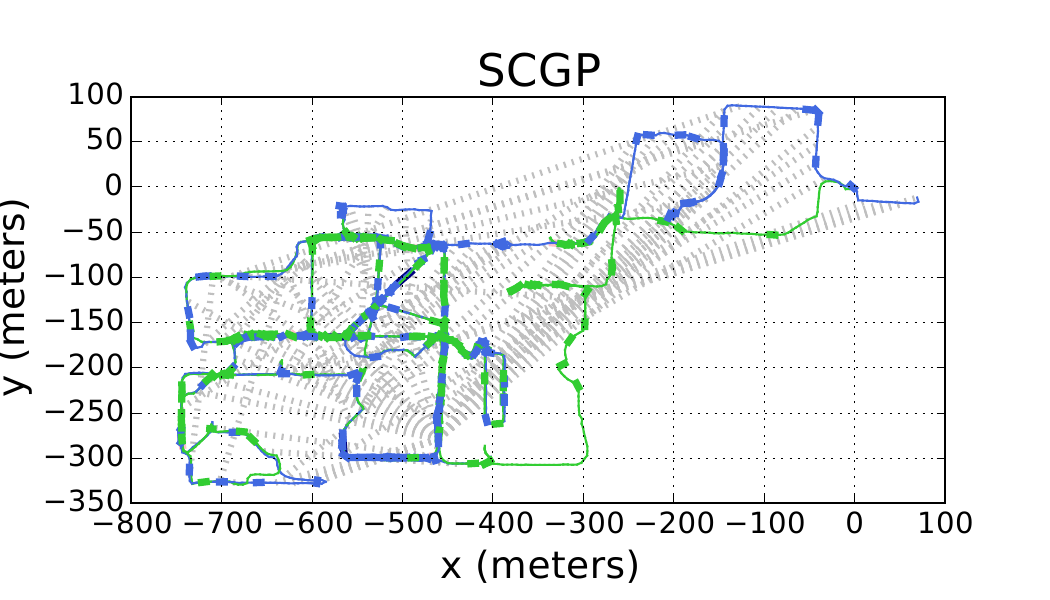}%
    \label{sfig:nclt_scgp}%
  }%
  \caption{Plots of the trajectories of two partial sessions of the NCLT
    dataset as estimated by PCM-HeuPatt, RANSAC, DCS, and SCGP.  \label{fig:nclt}}
\end{figure*}
\begin{figure}[t]%
  \centering%
  \includegraphics[width=0.98\columnwidth,trim={0.1cm 0.6cm 0.1cm 0cm},clip]{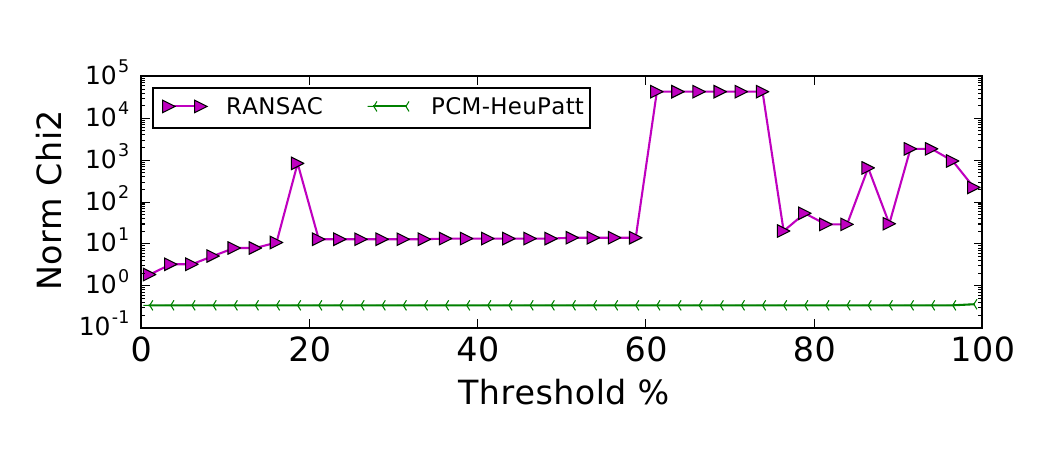}%
  \caption{The normalized chi-squared value (Chi2) of the resulting NCLT graph versus the
    threshold value $\gamma$ for PCM-HeuPatt and RANSAC. The Chi2 value for RANSAC is never below
  one signifying that the selected factors are probabilistically inconsistent, while the Chi2 value for PCM is relatively constant and below 1.0 regardless of threshold. \label{fig:ncltchi2} }
\end{figure}

\vspace{-0.15cm}
\subsection{Real-World Pose-Graph SLAM}
\label{sec:real-world}

We evaluate PCM on the 3D University of Michigan North Campus Long-Term Vision and LiDAR Dataset (NCLT) released by \cite{ncarlevaris-2015a}. The NCLT dataset was collected using a Segway robot equipped with a LiDAR and Microstrain IMU, along with a variety of other sensors. There are 27 sessions in all with an average length of 5.5 km per session.

For our experiment, we took two sessions collected about two weeks apart, removed the first third of one and the last third of the other, and then generated potential loop-closure measurements between the two graphs by aligning every fourth scan on each graph using GICP (\cite{segal2009generalized}) and selecting the match with the lowest cost function. We then labeled these registrations as inliers and outliers by thresholding the translation and rotation mean-squared error of the estimated pose transformations with respect to the ground-truth poses for the dataset derived by performing pose-graph optimization on all 27 sessions. Finally, to increase the difficulty of the dataset, we removed all but one sixteenth of the measurements labeled as inliers from the graph, resulting in a graph with ten inliers and 98 outliers.

In this experiment, we compare PCM-HeuPatt with DCS, SCGP, and RANSAC. \cref{fig:ncltchi2} shows the normalized chi-squared value of the resulting graphs for RANSAC and PCM versus threshold.  \cref{table:nclt_comp} provides a comparison of results and \cref{fig:nclt} shows the estimated maps. The MSE was calculated using the same method as in the prior section, however in this test we calculated trajectory and map relative pose error separately. The trajectory MSE calculates the error in the estimated relative pose between consecutive nodes allowing us to evaluate graph correctness, while the relative map pose MSE evaluates the offset between the maps.

PCM results in the graph with the best trajectory MSE and the best translational MSE for the relative pose of the two graphs and results in a consistent graph regardless of threshold. It also detects all the inlier measurements as well as three of the measurements labeled as outliers. DCS once again disables all measurements. SCGP results in a good graph but only enables three of the inlier measurements, and finally RANSAC (for both of the lowest thresholds tried) enables all inliers and several outliers and results in an inconsistent graph regardless of the threshold selected.

Note that while in this experiment PCM admits more false positives than in the last experiment, the measurements it accepts are consistent with the inlier measurements and local trajectories even though they were labeled as outliers (\cref{fig:ncltchi2}). In fact, notice that PCM has a better MSE for the relative map pose then the no outlier (NO-OUT) version of the graph. This suggests that by maximizing the consistent set, PCM is selecting measurements that are actually inliers but were mis-labeled as outliers when compared to the ground-truth. After verification this turned out to be the case.

It is also important to note that although SCGP results in a good graph for this dataset, as shown in the earlier experiment, this does not occur in all cases. In addition, if it fails to select the maximum consistent set of measurements, this can be catastrophic in the case of perceptual aliasing. 

\section{Group-$k$ Consistency Maximization}
\label{sec:gkcm}

In this section, we generalize the notion of consistency to sets of $k>2$ measurements and use this generalized definition to formulate a combinatorial optimization problem.

While maximizing pairwise consistency in \cite{mangelson2018pairwise} outperformed other existing robust SLAM methods, pairwise consistency is not always a sufficient constraint to remove outlier measurements. For example, a set of three range measurements may all intersect in a pairwise manner even if the set of measurements do not intersect at a common point, indicating that they are pairwise consistent but not group-3 consistent.

As currently framed, the consistency check described in \cite{mangelson2018pairwise} is only dependent on two measurements. In some scenarios, such as with the range measurements described above, we may want to define a consistency function that depends on more than two measurements.

\subsection{Group-$k$ Consistency}

To handle the situation where consistency should be enforced in groups of greater than two measurements we now define a novel notion of \textit{group-$k$ internally consistent sets}.

\begin{definition}
  A set of measurements $\bvec{\widetilde{Z}}$ is \textbf{group-$\mathbf{k}$ internally consistent} with respect to a consistency metric $C$ and the threshold $\gamma$ if
\begin{align}
  C(\{\bvec{z}_o, \cdots, \bvec{z}_k\}) \leq \gamma, \quad \forall \quad \{\bvec{z}_0, \cdots, \bvec{z}_k\} \in \mathcal{P}_k(\bvec{\widetilde{Z}})
  \label{eq:gk_consistency}
\end{align}
where $C$ is a function measuring the consistency of the set of measurements $\{\bvec{z}_o, \cdots, \bvec{z}_k\}$, $\mathcal{P}_k(\mathbf{\bvec{\widetilde{Z}}})$ is the the set of all permutations of $\bvec{\widetilde{Z}}$ with cardinality $k$, and $\gamma$ is chosen a priori.
\end{definition}
This definition of consistency requires that every combination of measurements of size $k$ be consistent with $C$ and $\gamma$. We note that the $n$-invariant introduced by \cite{shi2021robin} is a specific case of our generalized consistency function that does not depend on the relative transformation between poses where measurements were taken. The appropriate choice of consistency function is problem dependent and therefore left to the user to determine. However, we define consistency functions for our two target applications in \cref{sec:range_slam} and \cref{sec:ma_visual_pgo}.

As with pairwise consistency, establishing group-$k$ consistency does not guarantee full joint consistency. We settle for checking group-$k$ consistency and use it as an approximation for joint consistency to keep the problem tractable.

\subsection{Group-$k$ Consistency Maximization}

Analogous to pairwise consistency defined by \cite{mangelson2018pairwise}, we now want to find the largest subset of measurements that is internally \emph{group-$k$ consistent}. We use the same assumptions described in \cref{sec:comb_formulation}.

As in PCM, our goal is to find the largest consistent subset of $\bvec{Z}$. We accomplish this by introducing a binary switch variable $s_u$ for each measurement in $\bvec{Z}$ and let $s_u$ be 1 if the measurement is contained in the chosen subset and 0 otherwise. Letting $\bvec{S}$ be the vector containing all $s_u$, our goal is to find $\bvec{S}^*$ to the following optimization problem

\begin{equation}
  \begin{gathered}
    \bvec{S}^{*} = \underset{\bvec{S}\in\{0,1\}^m}{\operatorname{argmax}} \: \norm{\bvec{S}}_0 \\
    \text{s.t.}~~
 C(\{\bvec{z}_0, \cdots, \bvec{z}_k\}) ~ s_0 \cdots s_k \leq \gamma \\~~~~~~~~~~~~~~~~~~~~~~~ \forall \{\bvec{z}_0, \cdots, \bvec{z}_k\} \in \mathcal{P}_k(\bvec{Z})
\end{gathered}
\label{eq:comb_formulation_gkcm}
\end{equation}
where $m$ is the number of measurements in $\bvec{Z}$ and $\bvec{z}_u$ is the measurement corresponding to $s_u$. We refer to this problem as the Group-$k$ Consistency Maximization, or G$k$CM, problem. This problem is a generalization of PCM, and for $k=2$ they become identical.

\begin{figure}[!tb]%
  \centering%
  \subfigure[Generalized Graph]{%
    \includegraphics[width=0.4\columnwidth]{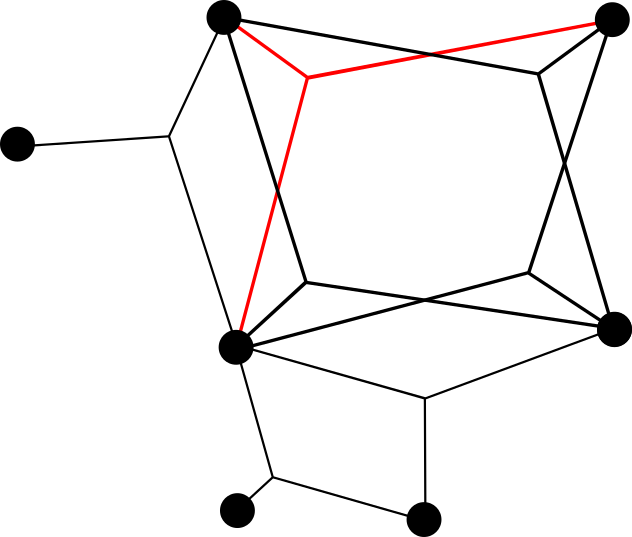}%
    \label{sfig:gen_graph_edge}%
  }%
  \hfil%
  \subfigure[Maximum Clique]{%
    \includegraphics[width=0.4\columnwidth]{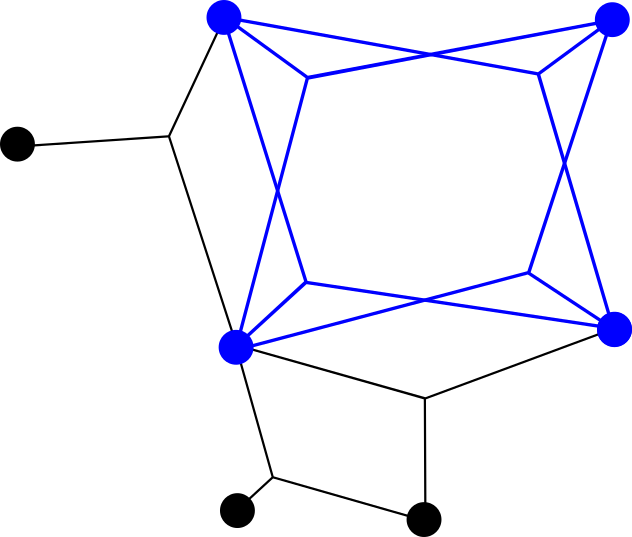}%
    \label{sfig:gen_graph_clique}%
  }%
  \caption{An example of a generalized consistency graph with edges made of $3$-tuples. \subref{sfig:gen_graph_edge} highlights that each edge denotes consistency of $3$ measurements. \subref{sfig:gen_graph_clique} highlights the maximum clique of the generalized graph in blue.}
  \label{fig:gen_graph}
\end{figure}

\subsection{Solving Group-$k$ Consistency Maximization}

As with PCM, we can solve the G$k$CM problem by finding the maximum clique of a consistency graph. However, because we want to find the largest subset that is group-$k$ internally consistent, we need to operate over generalized graphs. In graph theory, a \emph{k-uniform hypergraph} (or \emph{generalized graph}), $G$, is defined as a set of vertices $V$ and a set of $k$-tuples of those vertices $\mathcal{E}$ (\cite{bollobas1965generalized}). Each $k$-tuple is referred to as an edge and a \emph{clique} within this context is a subgraph of $G$ where every possible edge exists in $\mathcal{E}$. We now introduce the concept of a \emph{generalized consistency graph}:
\begin{definition}
  A \textbf{generalized consistency graph} is a generalized graph $G=\{V, \mathcal{E}\}$ with $k$-tuple edges, where each vertex $v \in V$ represents a measurement and each edge $e \in \mathcal{E}$ denotes consistency of the vertices it connects.
\end{definition}
Solving \cref{eq:comb_formulation_gkcm} is equivalent to finding the maximum clique of a generalized consistency graph and consists of the following two steps: Building the generalized consistency graph; and finding the maximum clique. The next two sections explain these processes in more detail.


\section{Building the Generalized Consistency Graph}
\label{sec:build_graph}

The graph is built by creating a vertex for each measurement and performing the relevant consistency checks to determine what edges should be added. If the graph is created all at once, there are $\binom{m}{k}$ checks to perform. If the graph is being built incrementally by checking the consistency of a newly added measurement with those already in the graph then the number of checks is $\binom{m-1}{k-1}$. This means that as $k$ increases the number of checks that need to be performed increases factorially with $k$. Thus, it is important that the consistency function in \cref{eq:gk_consistency} be computationally efficient. Note that all the checks are independent, allowing for the computation to be parallelized on a CPU or GPU to decrease the time to perform the necessary checks.

Due to the explosive growth in the number of checks, we develop a method that can substantially reduce the number of checks to be computed. We utilize the fact that a set of measurements that are group-$k$ consistent will almost always be group-$(k-1)$ consistent, assuming that the group-$k$ and group-$(k-1)$ consistency functions are similar. Using this observation, we can first perform $\binom{m}{k-1}$ checks ($\binom{m-1}{k-2}$ checks in incremental scenarios) and record which combinations of measurements failed the consistency check. Then when performing the group-$k$ check we only need to check combinations for which all groups of $(k-1)$ measurements passed their respective check. The process can be done starting with group-2 checks and building up to the final group-$k$ check. We show in \cref{subsec:generalized_graph_eval} that we can significantly decrease time spent performing consistency checks.

For example, three range measurements are consistent if they all intersect at the same point. However, if any pair of the three measurements do not intersect then any group-3 consistency check containing that pair of measurements will also fail. Thus, when processing measurements, we can check the $\binom{m}{2}$ pairwise intersections of measurements and do only the group-3 checks on combinations where consistency is possible instead of the total $\binom{m}{3}$ checks.

\begin{algorithm*}[!tb]
    \small
  \caption{Exact Algorithm for Finding the Maximum Clique of a k-Uniform Hypergraph. \newline \textbf{Input:} Graph $G=(V,\mathcal{E})$,  \textbf{Output:} Maximum Clique $S_{max}$}
  \vspace{-5mm}
  \label{alg:gmc_exact}
  \begin{multicols}{2}
  \begin{algorithmic}[1]
    \Function{MaxClique}{$G=(V,\mathcal{E})$}
    \State $S_{max} \gets \emptyset$
    \For{$i=1$ to $n$} \label{alg:gmc_exact:node_loop}
    \If{$d(v_i)+1 \geq |S_{max}|$}
    \For{\textbf{each} $e \in E(v_i)$} \label{alg:gmc_exact:edge_for}
    \State $S \gets e \cup v_i; $ $U \gets \emptyset$
    \State $R \gets $ \Call{CombinationsOfSize}{$S, k-1$} \label{alg:gmc_exact:R}
    \For{\textbf{each} $v_j \in N(v_i)$}
    \If{$j > i$}
    \If{$d(v_j)+1 \geq |S_{max}|$}
    \If{$R \subset E(v_j)$} \label{alg:gmc_exact:limit_R}
    \State $U \gets U \cup \{v_j\}$    \label{alg:gmc_exact:U_init}
    \EndIf
    \EndIf
    \EndIf
    \EndFor
    \State \Call{Clique}{$G, R, S, U$}
    \EndFor
    \EndIf
    \EndFor
    \EndFunction
  \item[]
  \item[]
  \item[]
  \item[]
  \item[]
  \end{algorithmic}
  \begin{algorithmic}[1]
    \Function{Clique}{$G=(V,\mathcal{E})$, $R$, $S$, $U$}
    \If{$U = \emptyset$}
    \If{$|S| > |S_{max}|$}
    \State $S_{max} \gets S$ \label{alg:gmc_exact:update}
    \EndIf
    \EndIf
    \While{$|U| > 0$}
    \If{$|S| + |U| \leq |S_{max}|$}
    \State \textbf{return}
    \EndIf
    \State Select any vertex $u$ from $U$
    \State $U \gets U \setminus \{u\}; ~S_{rec} \gets S \cup \{u\}$ \label{alg:gmc_exact:update_S}
    \State $N'(u) := \{w | w \in N(u) $ \textbf{and} $d(w) \geq |S_{max}|\}$
    \State $U_{rec} \gets \emptyset;~R_{rec} \gets R$
    \For{\textbf{each} $ p\in $ \Call{CombinationsOfSize}{$S, k-2$}}
    \State $R_{rec} \gets R_{rec} \cup \{p \cup \{u\}\}$ \label{alg:gmc_exact:update_R}
    \EndFor
    \For{\textbf{each} $ q \in U \cap N'(u)$}
    \If{$R_{rec} \subset E(q)$}
    \State $U_{rec} \gets U_{rec} \cup \{q\}$ \label{alg:gmc_exact:update_U}
    \EndIf
    \EndFor
    \State \Call{Clique}{$G, R_{rec}, S_{rec}, U_{rec}$}
    \EndWhile
    \EndFunction
  \end{algorithmic}
  \end{multicols}
  \vspace{-5mm}
\end{algorithm*}

\section{Finding the Maximum Clique of a Generalized Graph}
\label{sec:max_cliqu_of_generalized_graph}

Once the graph has been built, we can find the largest consistent set by finding the maximum clique of the graph. The PCM algorithm used the exact and heuristic methods presented by \cite{pattabiraman2015fast} but these algorithms were not designed for generalized graphs and used only a single thread. Here, we generalize these algorithms to function over $k$-uniform hypergraphs and provide a parallelized implementation of their algorithms.

\begin{figure}[!tb]%
  \centering%
  \includegraphics[width=.93\columnwidth]{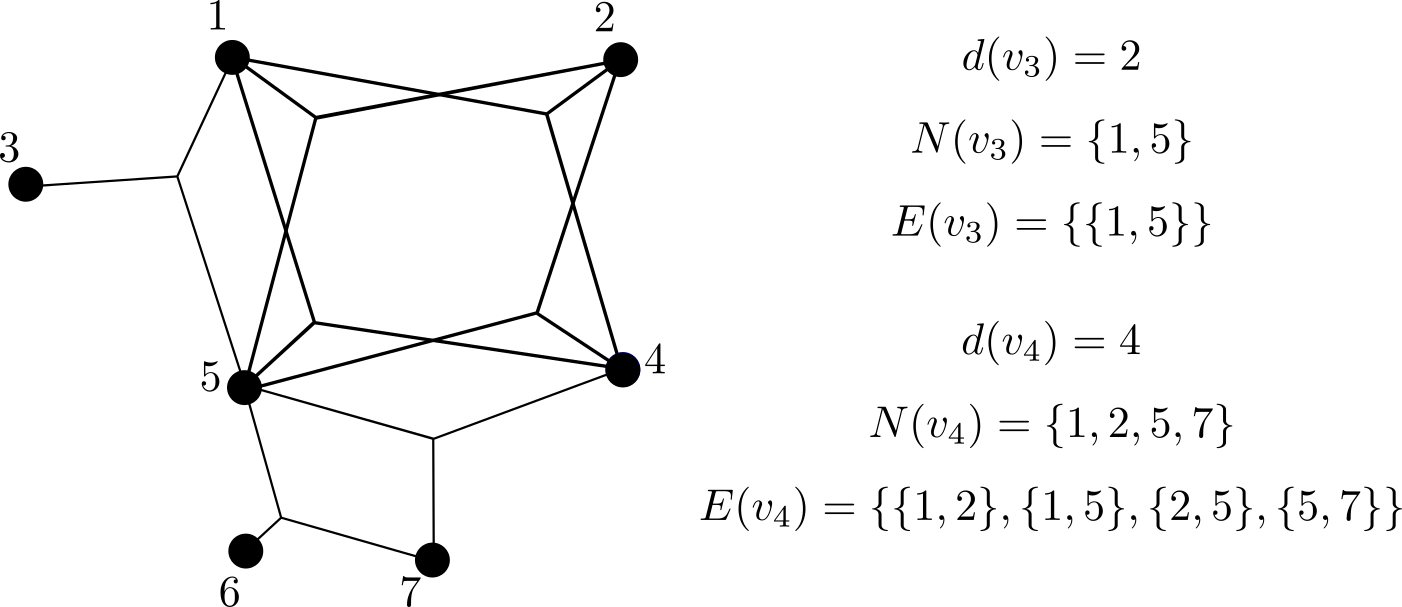}%
  \caption{Examples of the degree, neighborhood, and edge set definitions for generalized graphs.}%
  \label{fig:gen_graph_notation}%
\end{figure}

\begin{algorithm*}
    \small
  \caption{Heuristic Algorithm for Finding the Maximum Clique of a k-Uniform Hypergraph. \newline \textbf{Input:} Graph $G=(V,\mathcal{E})$,  \textbf{Output:} Potential Maximum Clique $S_{max}$}
  \vspace{-5mm}
  \label{alg:gmc_heuristic}
  \begin{multicols}{2}
  \begin{algorithmic}[1]
    \Function{MaxCliqueHeu}{$G=(V,\mathcal{E})$}
    \State $S_{max} \gets \emptyset$
    \For{$i=1$ to $n$}
    \If{$d(v_i)+1 \geq |S_{max}|$} \label{alg:gmc_heuristic:degree_check}
    \State Select $e \in E(v_i)$ with max connect. in $E(v_i)$ \label{alg:gmc_heuristic:select_edge}
    \State $S \gets e \cup v_i; $ $U \gets \emptyset$
    \State $R \gets $ \Call{CombinationsOfSize}{$S, k-1$}
    \For{\textbf{each} $v_j \in N(v_i)$}
    \If{$d(v_j)+1 \geq |S_{max}|$}
    \If{$R \subset E(v_j)$}
    \State $U \gets U \cup \{v_j\}$
    \EndIf
    \EndIf
    \EndFor
    \If{$|S| + |U| > |S_{max}|$}
    \State \Call{CliqueHeu}{$G, R, S, U$}
    \EndIf
    \EndIf
    \EndFor
    \EndFunction
    \item[]
  \end{algorithmic}
  \begin{algorithmic}[1]
    \Function{CliqueHeu}{$G=(V,\mathcal{E})$, $R$, $S$, $U$}
    \If{$U = \emptyset$}
    \If{$|S| > |S_{max}|$}
    \State $S_{max} \gets S$
    \EndIf
    \EndIf
    \State Select a vertex $u \in U$ with max connect. in $E(v_i)$ \label{alg:gmc_heuristic:select_vertex}
    \State $U \gets U \setminus \{u\}; ~S_{rec} \gets S \cup \{u\}$
    \State $N'(u) := \{w | w \in N(u) $ \textbf{and} $d(w) \geq |S_{max}|\}$
    \State $U_{rec} \gets \emptyset;~R_{rec} \gets R$
    \For{\textbf{each} $ p\in $ \Call{CombinationsOfSize}{$S, k-2$}}
    \State $R_{rec} \gets R_{rec} \cup \{p \cup \{u\}\}$
    \EndFor
    \For{\textbf{each} $ q \in U \cap N'(u)$}
    \If{$R_{rec} \subset E(q)$}
    \State $U_{rec} \gets U_{rec} \cup \{q\}$
    \EndIf
    \EndFor
    \State \Call{CliqueHeu}{$G, R_{rec}, S_{rec}, U_{rec}$}
    \EndFunction
  \end{algorithmic}
  \end{multicols}
  \vspace{-4mm}
\end{algorithm*}

We start by defining relevant notation. We denote the $n$ vertices of the graph $G=(V, \mathcal{E})$ as $\{v_1, \cdots, v_n\}$. Each vertex has a neighborhood $N(v_i)$, that is the set of vertices connected to that vertex by at least one edge. The degree of $v_i$, $d(v_i)$, is the number of vertices in its neighborhood. We also define an edge set, $E(v_i)$, for each vertex consisting of a set of $(k-1)$-tuples of vertices. The edge set is derived from the set of $k$-tuples in $\mathcal{E}$ containing the given vertex by removing the given vertex from each edge. \Cref{fig:gen_graph_notation} shows an example of these values for a given graph.

\subsection{Algorithm Overview}

The generalized exact and heuristic algorithms presented in \cref{alg:gmc_exact} and \cref{alg:gmc_heuristic} respectively are similar in structure to the algorithms by \cite{pattabiraman2015fast} but require additional checks to guarantee a valid clique is found since the algorithms now operate over generalized graphs.

The exact algorithm, \cref{alg:gmc_exact}, begins with a vertex $v$ and finds cliques of size $k$ that contain $v$ (MaxClique line \ref{alg:gmc_exact:edge_for}). A set of vertices, $U$, that would increase the clique size by one is found (MaxClique line \ref{alg:gmc_exact:limit_R}) from the set of edges $R$ that a valid candidate vertex must have (MaxClique line \ref{alg:gmc_exact:R}). The Clique function then recursively iterates through potential cliques and updates $R$ and $U$ (Clique lines \ref{alg:gmc_exact:update_R}, \ref{alg:gmc_exact:update_U}). The clique is tracked with $S$ and a check is performed to see if $S>S_{max}$ where $S_{max}$ is replaced with $S$ if the check passes. The process is repeated for each vertex in the graph (MaxClique line \ref{alg:gmc_exact:node_loop}). The exact algorithm evaluates all possible cliques, and as such, the time complexity of the exact algorithm is exponential in the worst case.

The heuristic algorithm, \cref{alg:gmc_heuristic}, has a similar structure to the exact algorithm but uses a greedy search to find a potential maximum clique more quickly. For each node with a degree greater than the size of the current maximum clique (MaxCliqueHeu line \ref{alg:gmc_heuristic:degree_check}), the algorithm selects a clique of size $k$ who has the greatest number of connections in $E(v_i)$ (MaxCliqueHeu line \ref{alg:gmc_heuristic:select_edge}). This is done by summing the number of connections each node in $N(v_i)$ has in $E(v_i)$ and selecting the edge $e \in E(v_i)$ with the sum total of connections. If the selected clique can potentially be made larger than $S_{\text{max}}$, then a greedy search selects nodes based on the largest number of connections in $E(v_i)$ (CliqueHeu line \ref{alg:gmc_heuristic:select_vertex}). The generalized heuristic algorithm presented in \cref{alg:gmc_heuristic} has the same complexity of $O(n\Delta^2)$ as the original algorithm presented by \cite{pattabiraman2015fast} despite the modifications made to operate on generalized graphs.

Both algorithms are guaranteed to find a valid clique and can be easily parallelized by using multiple threads to simultaneously evaluate each iteration of the for loop on line \ref{alg:gmc_exact:node_loop} of MaxClique and MaxCliqueHeu. This significantly decreases the run-time of the algorithm. Our released C++ implementation allows the user to specify the number of threads to be used.

A heuristic was introduced by \cite{chang2021kimera} to avoid computing the maximum clique from scratch in incremental scenarios. When a new measurement is received the maximum clique will either remain unchanged or a larger clique will exist. A more efficient search can be performed by only searching for cliques that contain the new measurement and comparing the largest clique with that measurement to the current maximum clique. We implement this heuristic in our maximum clique algorithms over generalized graphs so that G$k$CM can be performed in both batch and incremental scenarios.

\subsection{Evaluation of generalized graph operations}
\label{subsec:generalized_graph_eval}

We carried out several experiments to evaluate the effectiveness of \cref{alg:gmc_exact}, \cref{alg:gmc_heuristic}, and our hierarchy based approach to evaluating consistency. We also provide a comparison of \cref{alg:gmc_exact,alg:gmc_heuristic} with the MILP algorithm in \cite{shi2022optimal} and the maximum $k$-core algorithm in \cite{shi2021robin} which, to our knowledge, are the only other algorithms that find or approximate the maximum clique of a generalized graph.

\subsubsection{Hierarchy of Consistency Evaluation}

In this first experiment, we first evaluate the effectiveness of using a hierarchy or consistency checks to decrease the time required to perform all the required checks in a group-4 scenario. For a given number of measurements $m$, we randomly picked $\frac{m}{10}$ measurements that would be in the maximum clique meaning that these measurements are always consistent with each other. We also randomly identified other groups of measurements that are consistent with each other so that the consistency graph that would be formed would have 20\% of the total edges present in the graph. Once all the measurements were generated we evaluated the runtime to check the consistency of all measurements in both a batch and incremental manner. We performed three different evaluations. The first was using just a group-4 check, the second was a group-3 check followed by a group-4 check, and the last was a group-2 check followed by a group-3 check followed by a group-4 check. We performed each test 100 times and took an average of the run time.

\begin{figure}[!tbh]
    \centering
    \includegraphics[width=0.98\columnwidth]{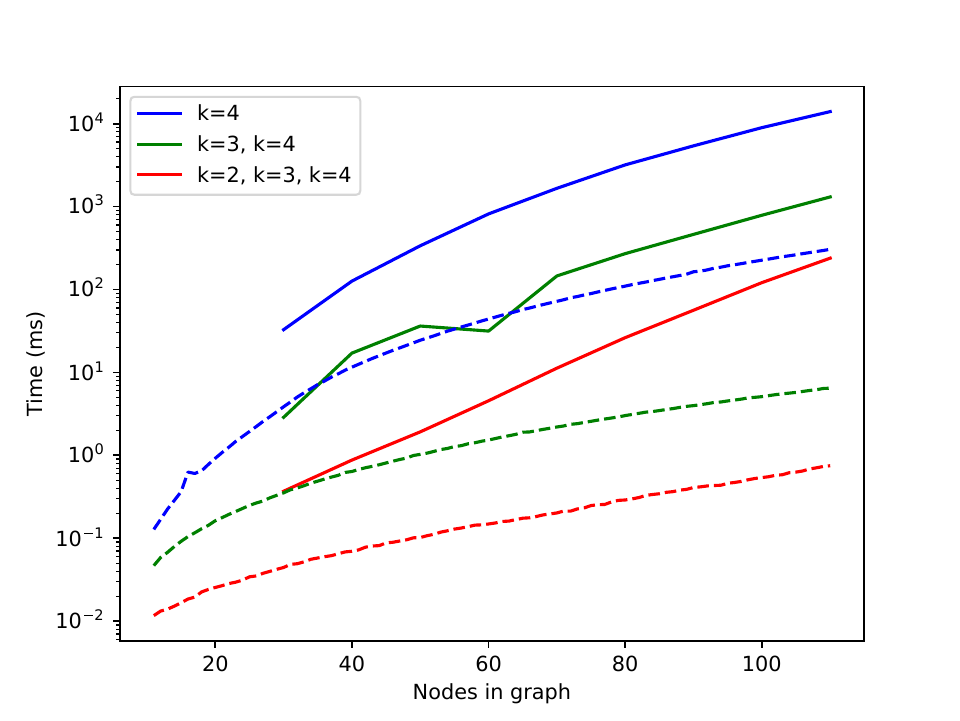}
    \caption{Time to complete all consistency checks in both batch and incremental scenarios using different hierarchies of checks. Batch results are represented by solid lines while incremental results are dashed. \textit{Note the log scale on the time axis.}}
    \label{fig:hierarchy_timing}
\end{figure}

We first tested the efficacy of this heuristic in a batch scenario where all measurements are processed after the data has been collected. For a given 4-uniform hypergraph size we generated a graph with $n$ nodes as described previously. We varied the number of measurements between 30 and 110 and tested in increments of 10 measurements recording the time required to process all checks. We repeated the test 100 times and averaged the time to perform all checks. As can be seen in \cref{fig:hierarchy_timing} for each hierarchy used we achieved about an order of magnitude speed up for this setup. Using just a group-4 check required an average of 13.9 s to perform the checks for all 110 measurements and we reduced that time to an average of 0.244 s when group-2 and group-3 checks were used to filter out inconsistent combinations.

We also tested the hierarchy of checks in incremental scenarios. The setup for the incremental scenario was the same as the batch scenario except that we started with 10 measurements and measured the time required to process all the required checks as each new measurement came in. We saw similar speed gains as those found in the batch scenario decreasing the required time from 305 ms to 0.73 ms. The entirety of the results for this experiment can be seen in \cref{fig:hierarchy_timing}.

We note that this approach is most effective in high outlier regimes where the vast majority of the measurements will not be consistent with one another. In scenarios where the outlier ratio is low, using the proposed hierarchy approach provides no benefit and can even result in increased run times because few combinations will be filtered out. As such we note that the efficacy of this technique will be situation dependent. This technique also shows one advantage of our consistency function formulation over the use of invariants as used in several other works (\cite{shi2021robin,lusk2021clipper,gentner2023gmcr}). This technique may not be usable with invariant functions since a lower-order invariant may not exist for a given application.

\subsubsection{Timing Comparison of Maximum Clique Algorithms}
In this second experiment, we evaluate the runtime characteristics of the proposed maximum clique algorithms over generalized graphs in \cref{alg:gmc_exact,alg:gmc_heuristic} as well as the MILP algorithm (\cite{shi2022optimal}) and the maximum k-core method (\cite{shi2021robin}).

We begin by outlining how to find a maximum $k$-core as well as the MILP problem. A $k$-core of a graph $G$ is the largest subgraph of $G$ such that every vertex in $G$ has a degree of at least $k$. The maximum $k$-core is the $k$-core with maximum $k$ for $G$ or where the ($k+1$)-core is the empty set. We base our maximum $k$-core algorithm based on the work by \cite{matula1983smallest} which presents a linear-time algorithm. The $k$-core algorithm used by \cite{shi2021robin} finds the $k$-core of a 2-uniform hypergraph by embedding a k-uniform hypergraph into a 2-uniform hypergraph. For example, an edge in a 3-uniform hypergraph that connects nodes $a$, $b$, and $c$, would form three edges in a 2-uniform graph. These edges would connect nodes $a$ and $b$, $a$ and $c$, and $b$ and $c$. While \cite{shi2021robin} note that the maximum $k$-core often provides a good approximation of the maximum clique we note that this is for the maximum clique of a 2-uniform hypergraph. We will show later that this does not always hold for $k$-uniform hypergraphs.

The MILP algorithm developed in \cite{shi2022optimal} is defined as follows. Given a $k$-uniform hypergraph $G(V, \mathcal{E})$ where $\left| V \right|=N$ the maximum clique can be found by solving the following MILP:
\begin{equation}
  \begin{aligned}
    \max_{\mathbf{b} \in \{0,1\}^N} \; & \sum_{i=1}^{N}b_i \\
    \text{s.t.} \; &\sum_{i \in \mathcal{M}} b_i \leq k-1, \; \forall \mathcal{M} \subset V, \left|\mathcal{M}\right|=k, \mathcal{M} \notin \mathcal{E}
  \end{aligned}
  \label{eq:robin_milp}
\end{equation}
The algorithm seeks to maximize the number of vertices in the group subject to the constraints that for every potential edge $\mathcal{M}$ that does not appear in $\mathcal{E}$, the sum of the variables $b_i$ must be less than $k$. We implement this MILP algorithm in C++ using the Gurobi solver (\cite{gurobi2017gurobi}).

\begin{figure*}[!tb]
  \centering%
  \includegraphics[width=1.98\columnwidth]{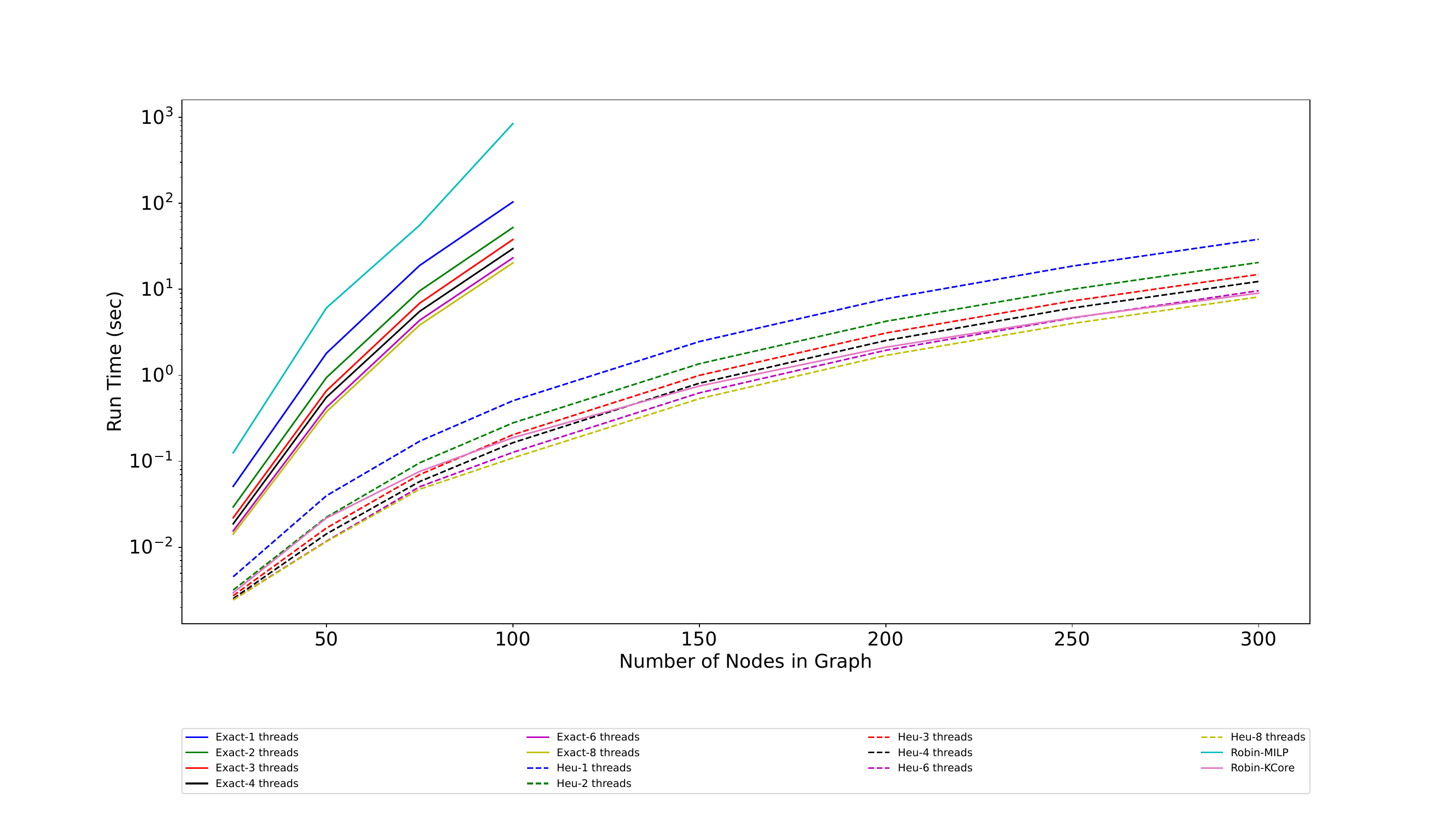}%
  \caption{Average run-time for generalized maximum clique algorithms proposed in this paper. We also compare against other generalized maximum clique algorithms in \cite{shi2022optimal,shi2021robin}. This includes both the time to evaluate the necessary data structures such as neighborhoods/edge sets and the time to estimate the maximum clique. Using eight threads, the heuristic algorithm was able to find the maximum clique of a graph with 250 nodes in a few seconds. }%
  \label{fig:gmc_timing}%
  \vspace{3mm}
\end{figure*}

In this experiment, we randomly generated 3-uniform hypergraphs with various node counts ranging from 25 vertices to 300 vertices. Each graph contained all the edges necessary to contain a maximum clique of cardinality ten and additional randomly selected edges to meet a specified graph density. While the run times of the algorithms are dependent on the density of the graph, for this experiment, we chose to hold the density of the graph constant at 0.1 such that approximately 10 percent of all potential edges were contained in the graph. We generated 100 sample graphs for each number of nodes and used all four of the algorithms listed above to estimate the maximum clique of each graph and measured the average run-time for each.  \Cref{fig:gmc_timing} shows the results of this experiment using various numbers of threads ranging from one to eight. The exact algorithm and the MILP algorithm were only used for graphs with a total number of nodes of 100 or less because of the exponential nature of the algorithms.

As can be seen in \cref{fig:gmc_timing} the MILP algorithm in \cite{shi2022optimal} is the slowest algorithm followed by the exact algorithm in \cref{alg:gmc_exact}. This is to be expected since both algorithms are guaranteed to find the maximum clique and do not use any heuristics or approximations that would cause them to find a suboptimal solution. Both the $k$-core and heuristic algorithm (\cref{alg:gmc_heuristic}) run significantly faster with the heuristic maximum clique algorithm using eight threads being the fastest in this test case. This is primarily due to the graph density. As noted above, the performance of both \cref{alg:gmc_exact} and \cref{alg:gmc_heuristic} are largely dependent on the graph's density and that run time will increase as the density increases. We suspect that as the graph density grows the runtime for \cref{alg:gmc_heuristic} would exceed that of the maximum $k$-core algorithm. Additionally, if we look at the rate of increase for the $k$-core and heuristic algorithms, the $k$-core algorithm seems to scale slightly better with the number of vertices in the graph. For larger graphs than those used in this experiment, we expect that the $k$-core algorithm will outperform \cref{alg:gmc_heuristic} in terms of speed.

\subsubsection{Heuristic Evaluation}

\begin{figure}[!tb]%
  \centering%
  \includegraphics[width=.93\columnwidth]{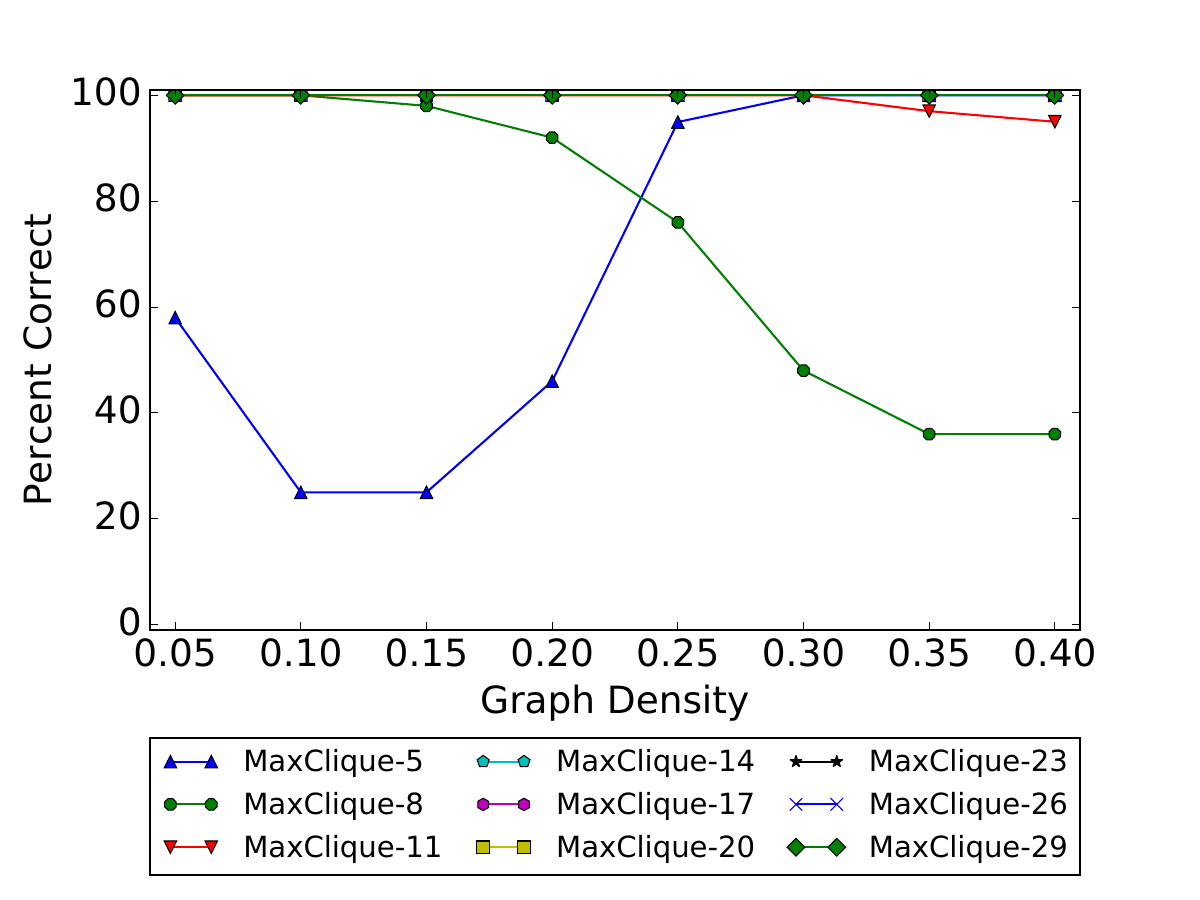}%
  \caption{Evaluation of the heuristic algorithm proposed in Algorithm \ref{alg:gmc_heuristic}. Individual lines denote the cardinality of the maximum clique inserted into the graph. The horizontal axis denotes the density of edges in the graph and the vertical axis denotes the percentage of test cases where the algorithm returned a clique of the correct cardinality. The heuristic algorithm returned cliques of the correct size 100 percent of the time for the graphs with max clique size of 14, 17, 20, 23, 26, and 29. }%
  \label{fig:gmc_heuristic_evaluation}%
\end{figure}

In the third experiment, we again randomly generated 3-uniform hypergraphs, however, in this case, we varied the density of the graph and the size of the inserted clique, while holding the total number of nodes at 100. For each graph, we used the MaxCliqueHeu algorithm to estimate the maximum clique and then evaluated whether or not the algorithm was successful in finding a clique of the same size as the clique we inserted. We again generated 100 sample graphs for each combination of inserted-clique size and graph density. \Cref{fig:gmc_heuristic_evaluation} plots the summarized results. If the algorithm happened to return a maximum clique larger than the inserted clique, then the associated sample was dropped.

This experiment shows that the size of the maximum clique and the success rate of the proposed heuristic algorithm are correlated. In addition, it shows that, with the exception of the case when the inserted clique was very small (cardinality 5), the density of the graph and the success rate are inversely correlated. As such, the heuristic seems to perform best when the size of the maximum clique is large and/or when the connectivity of the graph is relatively sparse.

We evaluated the k-core method similarly but don't show the results in \cref{fig:gmc_heuristic_evaluation} . In this experiment we found that the $k$-core never found the exact maximum clique and further examination of the $k$-core showed that it was not a good approximation of the maximum clique. Our application experiments later show that the $k$-core can, at times, provide a good approximation of the maximum clique of a generalized graph which leads us to conclude that the quality of the approximation is largely dependent on the structure of the generalized graph and if that structure is maintained when embedding the graph.

\section{Range-based SLAM}
\label{sec:range_slam}

This section of the paper will consider G$k$CM in the context of a single-agent range-based SLAM scenario. Given that range-based SLAM often has multiple landmarks from which measurements are being generated, we make the following additional assumption:
\begin{assumption}
    Measurements to different beacons are known to be inconsistent (i.e. data association is known). We will relax this assumption later in one of our experiments.
\end{assumption}

For this application, we will use the following $k=4$ consistency check,
\begin{equation}
    \centering
    \begin{gathered}
    C(\bvec{z}_{ai}, \bvec{z}_{bi}, \bvec{z}_{ci}, \bvec{z}_{di}) =
    \norm{h(\bvec{X}_{abcd}, \bvec{Z}_{abc}^i) - \bvec{z}_{di}}_{\Sigma} \leq \gamma
    \end{gathered}
    \label{eq:range_consistency_check}
\end{equation}
where $\bvec{z}_{di}$ is a range measurement from pose $d$ to beacon $i$, $\bvec{X}_{abcd}$ is a tuple of poses $\bvec{x}_a$, $\bvec{x}_b$, $\bvec{x}_c$, and $\bvec{x}_d$, and $\bvec{Z}_{abc}^i$ is a tuple of range measurements from poses $\bvec{x}_a$, $\bvec{x}_b$, and $\bvec{x}_c$ to beacon $i$. The value $\gamma$ is a threshold value and the function $h(\bvec{X}_{abcd}, \bvec{R}_{abc}^i)$ is a measurement model defined as
\begin{equation}
    \centering
    h(\bvec{X}_{abcd}, \bvec{R}_{abc}^i) = \norm{\bvec{l}(\bvec{X}_{abc}, \bvec{R}_{abc}^i) - \bvec{p}_{d}}_2
\end{equation}
where $\bvec{X}_{abc}$ is a tuple of poses $\bvec{x}_a$, $\bvec{x}_b$, and $\bvec{x}_c$, and $\bvec{p}_i$ is the position of pose $i$. The function $ \bvec{l}(\bvec{X}_{abc}, \bvec{R}_{abc}^i)$ is a trilateration function that depends on the input poses and the associated range measurements and returns an estimate of the beacon's location. The covariance, $\Sigma$, is a function of the covariances on the measurements $\bvec{z}$ and the poses $\bvec{x}$. The joint covariance, $\Sigma_j$, of the poses and beacon location is calculated by forming the measurement Jacobian of a factor graph and using methods described by \cite{kaess2009covariance}. Once the joint covariance has been obtained the covariance, $\Sigma$, is calculated as $\Sigma = H \Sigma_T H^T + R_{z_d}$ where $H = \frac{\partial h} {\partial \bvec{x}, \bvec{l}}$ and $\Sigma_T = \text{blockdiag}(\Sigma_j, \Sigma_{z_d})$.

The metric checks that the range to the intersection point of three range measurements matches the range of the fourth measurement at some confidence level. The check is done four times for a given set of four measurements, where each permutation of three measurements is used to localize the beacon. As with the consistency function in PCM, this consistency function follows a $\chi^2$ distribution, meaning that the threshold value, $\gamma$, can be easily chosen without knowledge of the data. Given the combinatorial nature of group consistency, the trilateration algorithm needs to be fast and accurate. The algorithm described by \cite{zhou2011closed} fits these criteria and presents a closed-form algorithm that performs comparably to an iterative nonlinear optimization approach, but without the need for an initial guess or an iterative solver.

\subsection{Degenerate Configurations}
\begin{figure}[!tb]
    \centering
    \includegraphics[width=0.93\columnwidth]{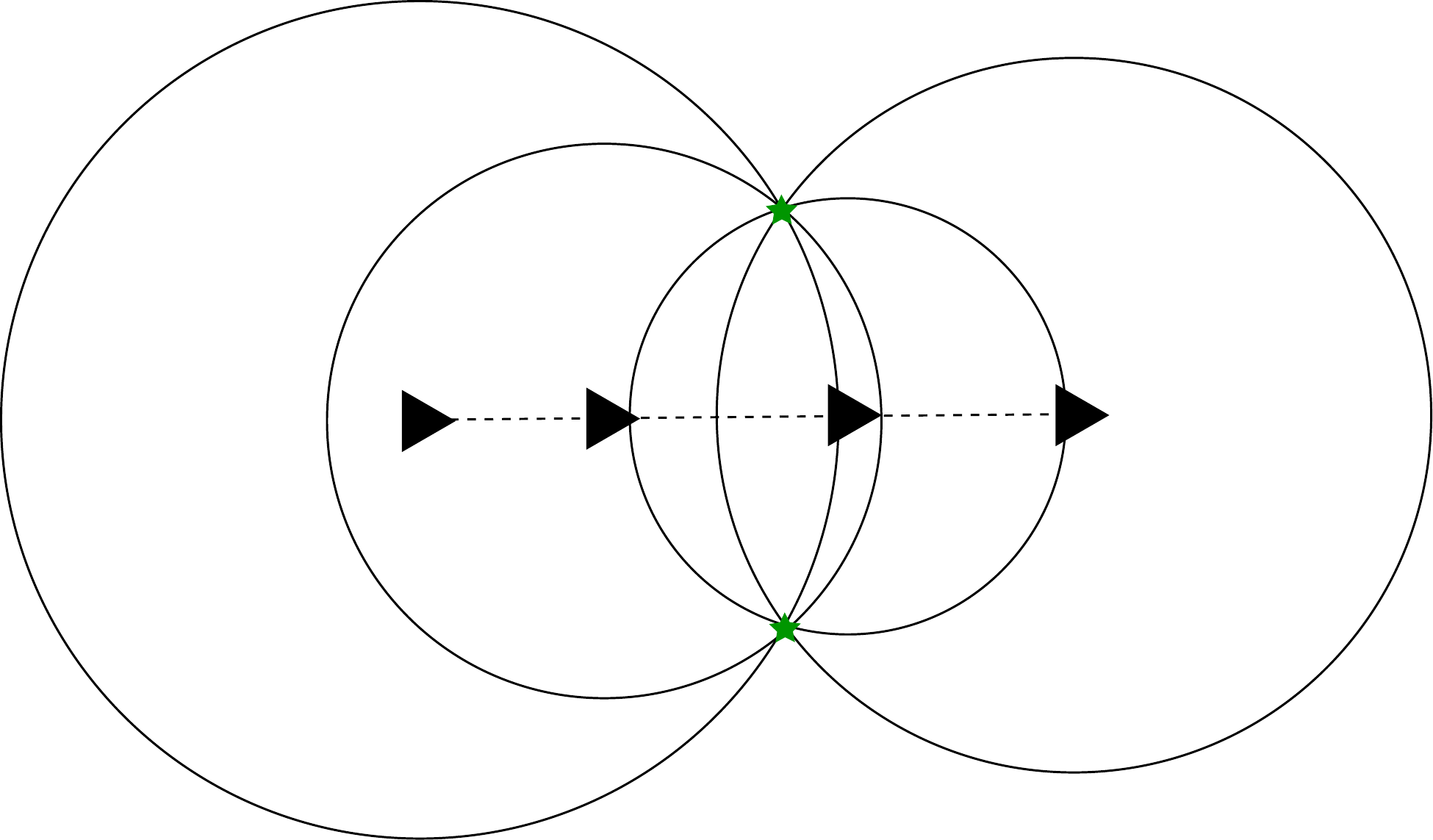}
    \caption{Degenerate pose configuration where range measurements do not result in a unique landmark location.}
    \label{fig:degenerate_config}
\end{figure}

Since our consistency check defined in \cref{eq:range_consistency_check} uses a trilateration algorithm, we need to discuss the scenarios where trilateration fails to provide a unique solution. The first case is where the poses are collinear as shown in \cref{fig:degenerate_config}, and the second is when two of the three poses occupy the same position. The trilateration algorithm by \cite{zhou2011closed} is robust to such configurations and can return two estimates for the beacon's location. The consistency check in \cref{eq:range_consistency_check} can pass if either estimate is deemed consistent.

If the trilateration algorithm is not robust to such scenarios, then a test to detect a degeneracy can be designed. If the test indicates the poses are in a degenerate configuration, one or more of the measurements can be stored in a buffer whose consistency with the maximum clique can be tested later. If a degeneration is still present, then the consistency of the measurement must be tested another way or the measurement be labeled inconsistent. In practice, we found that degenerate configurations did not present an issue because the odometry noise caused pose estimates used in the consistency check to not be degenerate even when the true configuration of poses was degenerate.

\section{Range-based SLAM Evaluation}
\label{sec:range_slam_results}

In this section, we evaluate the performance of G$k$CM on several synthetic datasets where a robot is exploring and taking range measurements to static beacons. Due to the run-time constraints, G$k$CM was only evaluated using the heuristic algorithm in \cref{alg:gmc_heuristic}, denoted as G$k$CM-HeuPatt going forward. We compare the results of G$k$CM-HeuPatt to the results of PCM, where the consistency check for PCM is the check used by \cite{olson2005single}, and both MILP and $k$-core variants of ROBIN. Since ROBIN-MILP uses the same consistency graph as G$k$CM, it will produce identical results to G$k$CM when the exact algorithm (\cref{alg:gmc_exact}) is used because both algorithms are guaranteed to find the maximum clique.

\subsection{Simulated 2D World}
First, we simulate a two-dimensional world where a robot navigates in the plane. We simulate three different trajectories, (Manhattan world, circular, and a straight line) along with range measurements to static beacons placed randomly in the world. Gaussian noise was added to all range measurements and a portion of the measurements were corrupted to simulate outlier measurements. Half of the corrupted measurements were generated in clusters of size 5, and the other half as single random measurements using a Gaussian distribution with a random mean and a known variance. We assume that the variances of the range measurements are known and that these variances are used when performing the consistency check. The simulation was run multiple times varying values such as the trajectory and beacon locations, and statistics were recorded for comparison.

\subsubsection{Monte Carlo Experiment}
This first example was done to show how well G$k$CM performs in situations with large percentages of outliers. In this experiment, a trajectory of 75 poses was simulated with measurements being taken at each pose and 60 of the measurements were corrupted to be outliers. G$k$CM was used to identify consistent measurements which were used to solve the range-based SLAM problem in \cref{eq:mle_problem} using GTSAM (\cite{kaess2011isam2}). The experiment averaged statistics over 100 runs and results are shown in \cref{tab:large_mc_stats} where we report the mean and standard deviation for each metric.

\begin{figure*}[tbh!]%
  \centering%
  \subfigure{%
    \includegraphics[width=0.7\columnwidth,trim={0.05cm 0.3cm 0.05cm 0cm},clip]{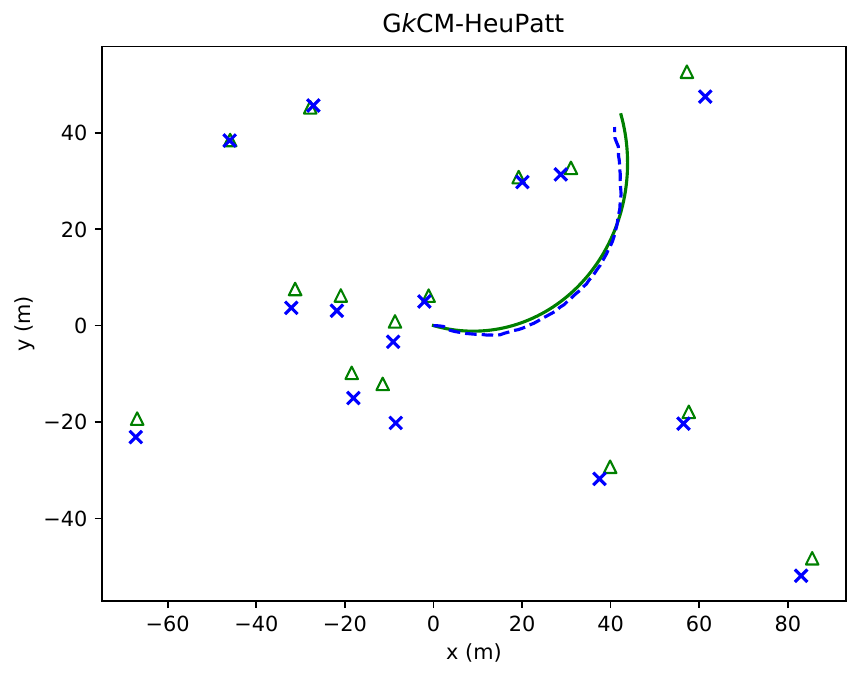}%
    \label{sfig:range_mc_gkcm}%
  }%
  \subfigure{%
    \includegraphics[width=0.7\columnwidth,trim={0.05cm 0.3cm 0.05cm 0cm},clip]{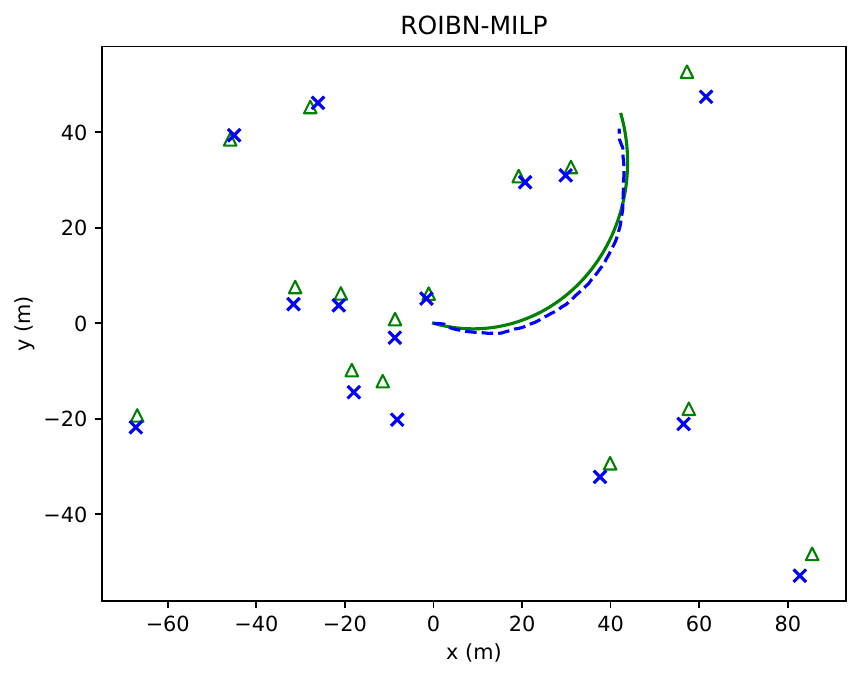}%
    \label{sfig:range_mc_milp}%
  }%
  \\
  \subfigure{%
    \includegraphics[width=0.7\columnwidth,trim={0.05cm 0.3cm 0.05cm 0cm},clip]{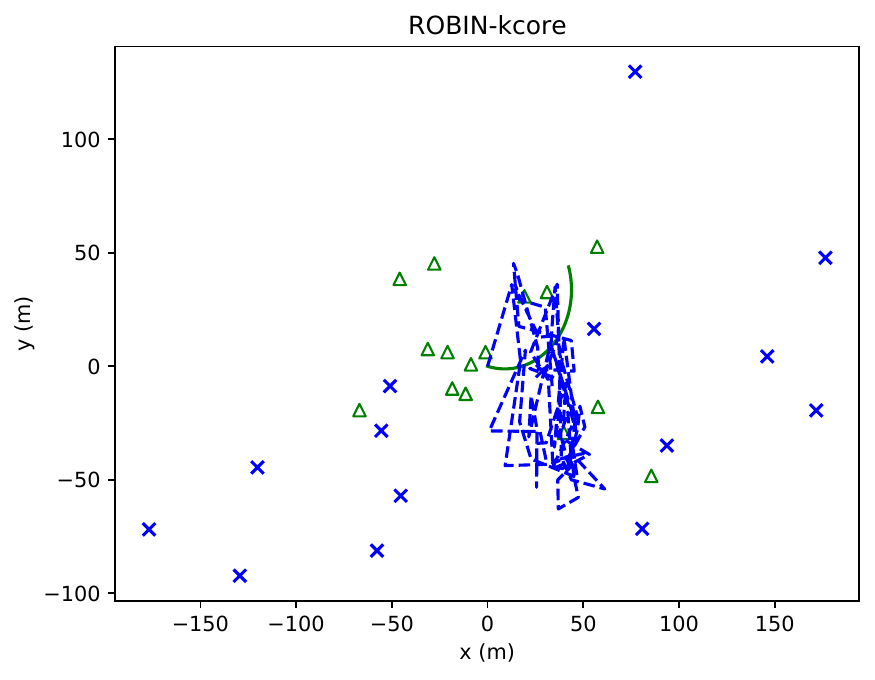}%
    \label{sfig:range_mc_kcore}%
  }%
  \subfigure{%
    \includegraphics[width=0.7\columnwidth,trim={0.05cm 0.3cm 0.05cm 0cm},clip]{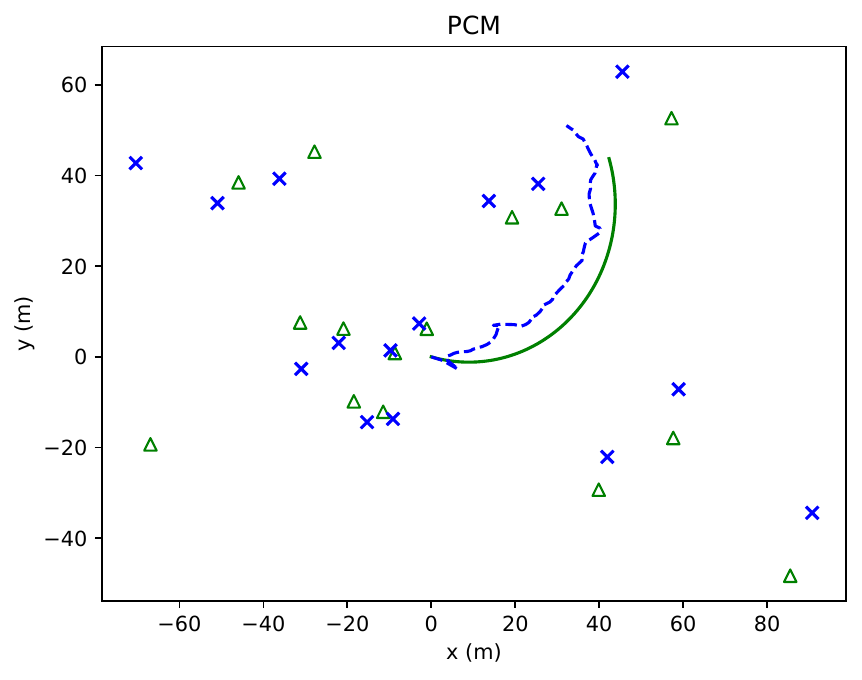}%
    \label{sfig:range_mc_pcm}%
  }%
  \caption{Results of a single run of the Monte Carlo experiment where 80\% of the 75 measurements considered are outliers. The blue dashed lines and x denote the estimated trajectory and beacon locations. The green line and triangles denote the true trajectory and beacon locations.}
  \label{fig::large_mc}
\end{figure*}

As can be seen, ROBIN-MILP performs the best across most metrics followed closely by G$k$CM-HeuPatt. The only difference between the two methods is that ROBIN-MILP uses an algorithm that is guaranteed to find the maximum clique while G$k$CM-HeuPatt is not. If \cref{alg:gmc_exact} had been used instead of \cref{alg:gmc_heuristic} the results between G$k$CM and ROBIN-MILP would have been identical. We note that the only statistic where either ROBIN-MILP or G$k$CM-HeuPatt was not the best method was the true positive rate (TPR). In this case, both PCM and ROBIN-$k$-core did better at including more inliers in their set of consistent measurements. However, we argue that since the presence of even a single outlier can have serious negative effects on the map quality the false positive rate (FPR) is a better metric to evaluate each algorithm. The normalized $\chi^2$ entry in the table tells us how well the state estimates fit the data that has been selected and ideally $\chi^2 < 3.84$ indicating that the data fits within a 95\% confidence interval. As can be seen, the data estimates fit the data reasonably well for both G$k$CM-HeuPatt and ROBIN-MILP and poorly for PCM and ROBIN-$k$-core. This is because PCM assumes that group-2 consistency is a suitable replacement for group-4 consistency. In this application, the embedding of the 4-uniform hypergraph into a 2-uniform hypergraph does not seem to maintain a similar enough structure for the maximum $k$-core to provide a good approximation of the maximum clique of the original graph. This is indicated by the high FPR and $\chi^2$ values produced when using the maximum k-core. \Cref{fig::large_mc} shows visual examples of the maps produced using the set of selected measurements by each method. The fact that the k-core algorithm performs so poorly arises from the fact that embedding of the 4-uniform hypergraph did not maintain the structure of the graph meaning that the $k$-core did not approximate the maximum clique well despite using group-4 consistency.

\begin{table*}[!tbh]
    \centering
    \vspace{5mm}
    \caption{Statistics comparing G$k$CM-HeuPatt to other methods in the Monte Carlo Experiment. Best results are in \textbf{BOLD}}
  \resizebox*{\linewidth}{!}{
\begin{tabular}{|c|cc|cc|cc|cc|cc|ccc|}
\hline
\multirow{2}{*}{} & \multicolumn{2}{c|}{Trans. RMSE (m)}     & \multicolumn{2}{c|}{Rot.  RMSE (rad)}    & \multicolumn{2}{c|}{Beacon Error (m)}     & \multicolumn{2}{c|}{Residual}           & \multicolumn{2}{c|}{Inliers}               & \multicolumn{3}{c|}{$\chi^2$}                                       \\ \cline{2-14}
                  & \multicolumn{1}{c|}{Avg} & Std           & \multicolumn{1}{c|}{Avg} & Std           & \multicolumn{1}{c|}{Avg} & Std            & \multicolumn{1}{c|}{Avg} & Std          & \multicolumn{1}{c|}{TPR} & FPR             & \multicolumn{1}{c|}{Avg} & \multicolumn{1}{c|}{Std} & Median        \\ \hline
G$k$CM-HeuPatt    & 1.15                     & 1.23          & 0.11                     & 0.07          & \textbf{20.92}           & \textbf{37.17} & 289.82                   & 502          & 0.79                     & \textbf{0.0037} & 1.42                     & 2.45                     & \textbf{0.44} \\ \hline
PCM               & 2.75                     & 4.63          & 0.22                     & 0.20          & 24.00                    & 39.00          & 10707                    & 16866        & \textbf{0.98}            & 0.015           & 46.25                    & 72.78                    & 6.24          \\ \hline
ROBIN-MILP        & \textbf{1.11}            & \textbf{0.95} & \textbf{0.10}            & \textbf{0.06} & 21.07                    & 37.54          & \textbf{279.36}          & \textbf{461} & 0.79                     & \textbf{0.0037} & \textbf{1.36}            & \textbf{2.25}            & \textbf{0.44} \\ \hline
ROBIN-$k$-core       & 9.62                     & 12.59         & 0.78                     & 0.83          & 45.43                    & 59.41          & 3.81e7                   & 7.19e6       & 0.90                     & 0.102           & 11488                    & 20462                    & 30.83         \\ \hline
\end{tabular}
}
    \label{tab:large_mc_stats}
\end{table*}

\begin{figure}[!tbh]
\vspace{-3mm} 
    \centering
    \includegraphics[width=\columnwidth, trim=0 0 0 25, clip]{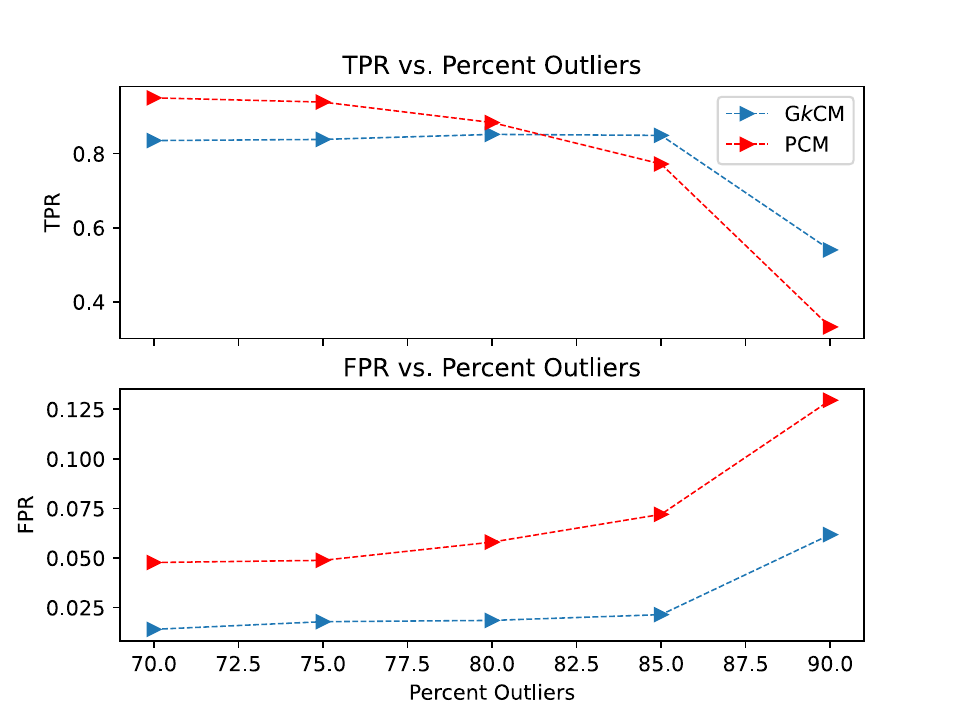}
    \caption{Results showing the normalized TPR ($\mathit{TP}/(\mathit{TP} + \mathit{FN}))$ and FPR $\mathit{FP}/(\mathit{FP} + \mathit{TN})$ by varying the number of outliers for a fixed trajectory.}
    \label{fig:varied_outlier_ratio}
\end{figure}

Additionally, we wished to know at what ratio of outliers to inliers did the performance of G$k$CM-HeuPatt begin to drop off. To measure this we simulated robot odometry for 100 poses and corrupted the measurements taken to a beacon with enough outliers to achieve a certain percentage of outliers. We ran the set of measurements through G$k$CM-HeuPatt and observed if the selected set of consistent measurements matched the set of inlier measurements. Using the same robot odometry, this was done with several different outlier percentages ranging from 70\% to 90\% outliers. The process was repeated for multiple trajectories and the true/false positive rates for each outlier percentage were recorded. Results can be seen in \cref{fig:varied_outlier_ratio}.

The figure shows that the true and false positive rates for G$k$CM-HeuPatt are fairly constant until about 85 percent of the measurements are outliers while the true positive rate decreases with the number of outliers for PCM and the false positive rate increases. These results are expected because as more outliers are present, it is more likely that either an outlier clique will form or that an outlier measurement will intersect with the inlier set with a pairwise basis showing the need for group-k consistency.

\subsection{Hardware Experiment}
This experiment evaluates the ability of G$k$CM-HeuPatt to reject outliers in range-based SLAM using an underwater vehicle and acoustic ranging to beacons of an unknown location. We use the dataset collected by \cite{olson2006robust}, where they use SCGP to detect and remove outlier range measurements. The data collected uses four beacons placed around the area of exploration. The underwater vehicle collects between 400 and 600 measurements to each beacon over the course of the experiment. Due to the exponential runtime of the maximum clique algorithms and the factorial increase in the number of consistency checks, we down-sample the number of measurements to 100 measurements to each beacon. Given that outliers are present in the data, we randomly selected 80 outlier measurements and 20 inlier measurements as classified by the SCGP algorithm. This was done to ensure that there is an inlier clique in the data while showing that we can reject outliers in high-outlier regimes.

\begin{figure*}[tbh!]%
  \centering%
  \subfigure{%
    \includegraphics[width=0.65\columnwidth,trim={0.05cm 0.3cm 0.05cm 0cm},clip]{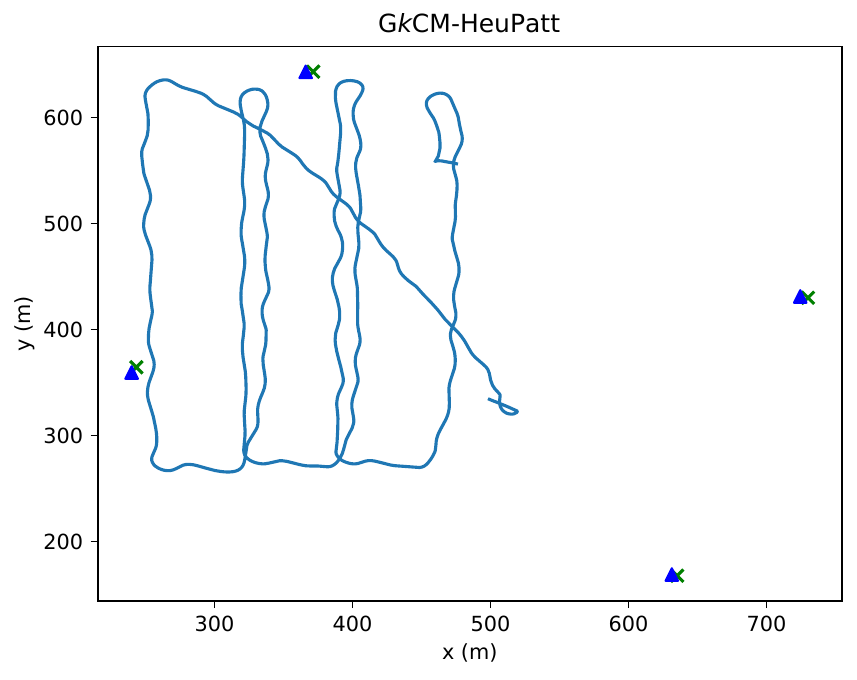}%
    \label{sfig:goats_gkcm}%
  }%
  \subfigure{%
    \includegraphics[width=0.65\columnwidth,trim={0.05cm 0.3cm 0.05cm 0cm},clip]{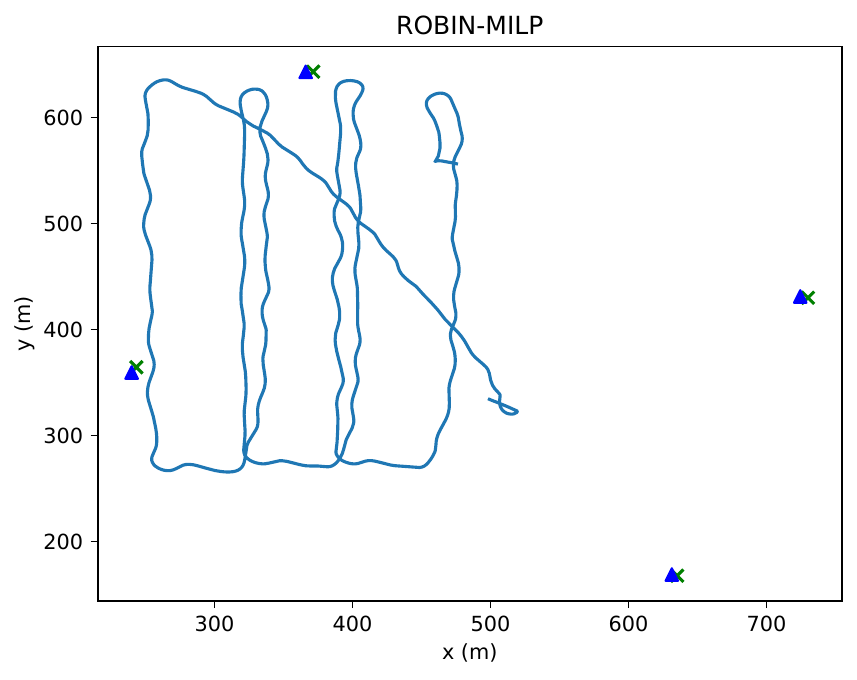}%
    \label{sfig:goats_milp}%
  }%
  \\
  \subfigure{%
    \includegraphics[width=0.65\columnwidth,trim={0.05cm 0.3cm 0.05cm 0cm},clip]{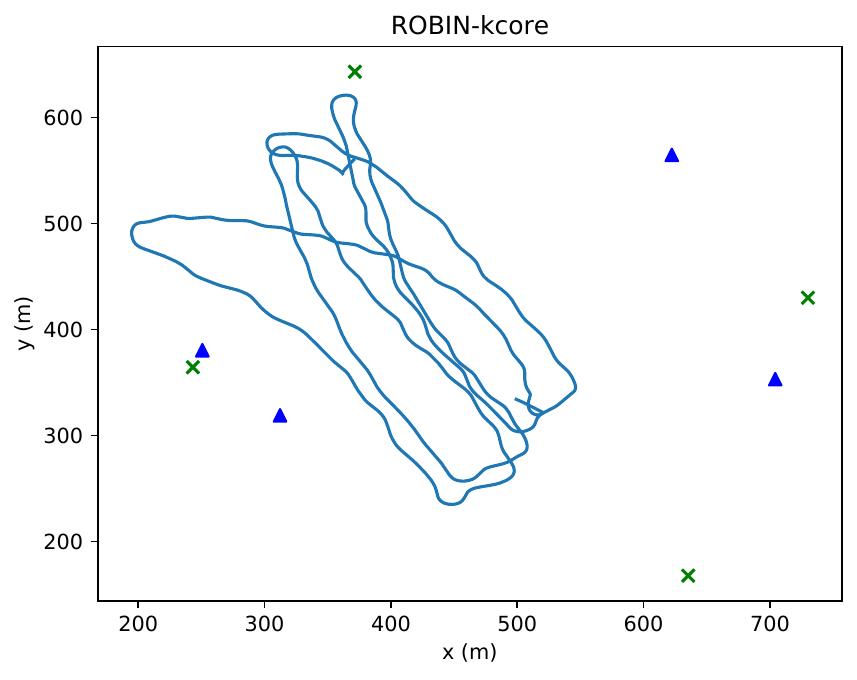}%
    \label{sfig:goats_kcore}%
  }%
  \subfigure{%
    \includegraphics[width=0.65\columnwidth,trim={0.05cm 0.3cm 0.05cm 0cm},clip]{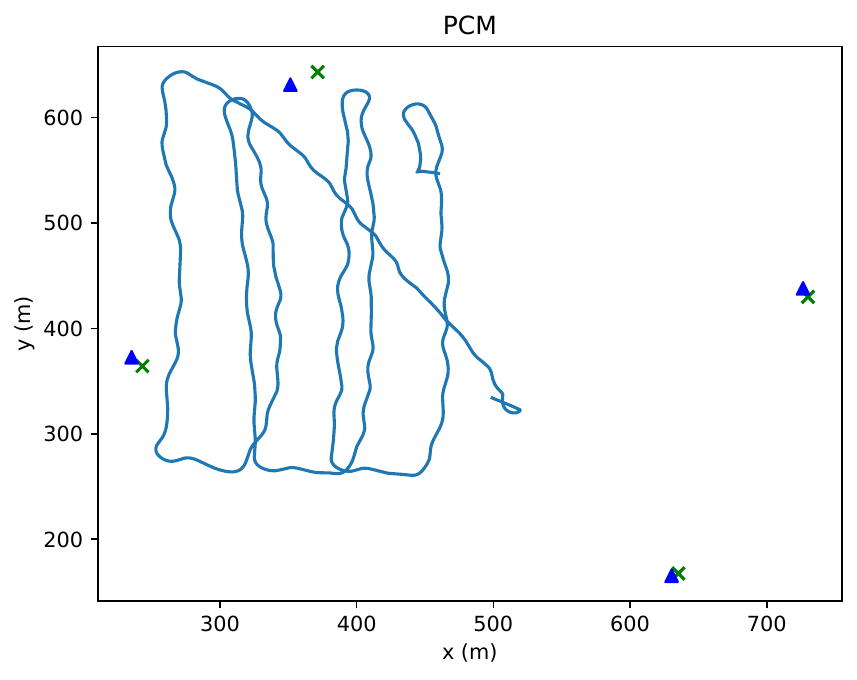}%
    \label{sfig:goats_pcm}%
  }%
  \subfigure{%
    \includegraphics[width=0.65\columnwidth,trim={0.05cm 0.3cm 0.05cm 0cm},clip]{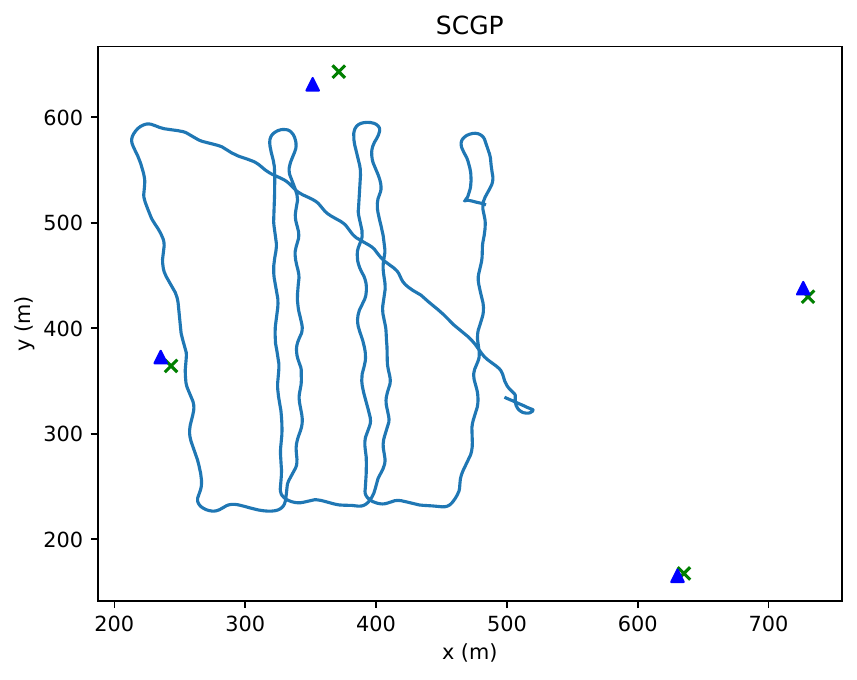}%
    \label{sfig:goats_scgp}%
  } \\
  \caption{Example plots of the maps estimated by G$k$CM-HeuPatt, PCM, MILP and $k$-core variants of ROBIN, and SCGP on the data collected in \cite{olson2006robust}. No truth data for the vehicle is available but the true beacon locations are denoted by a green x and the estimated locations by a blue triangle.}
  \label{fig:goats_hardware_exp}
\end{figure*}

We compare the results of G$k$CM with the results produced by SCGP, PCM, and ROBIN using both the MILP and $k$-core algorithms. A summary of our results can be found in \cref{tab:goats_results} and visual results in \cref{fig:goats_hardware_exp}. Both G$k$CM and the MILP variant of ROBIN perform best across all metrics. This was expected since both algorithms utilize group-$k$ consistency to evaluate the consistency of the range measurements and utilize maximum clique algorithms over generalized graphs. The primary difference between the two is that we tested G$k$CM using the heuristic maximum clique algorithm, \cref{alg:gmc_heuristic}, which can find a suboptimal clique, whereas the MILP (and the exact algorithm in \cref {alg:gmc_exact}) are guaranteed to find the maximum clique. The next best performers are PCM and SCGP where PCM had better beacon estimates while SCGP had a lower normalized $\chi^2$ score. We note that the performance of these algorithms varied greatly with the subset of measurements chosen to be used in the test. However, in all the tests we ran they never performed better than either G$k$CM or ROBIN-MILP although they occasionally achieved similar performance. The worst performing algorithm ended up being the ROBIN-k-core variant which is surprising considering that the algorithm uses a group-$k$ consistency metric. As noted before the shortcoming of using the maximum k-core is that the embedding of the 4-uniform hypergraph into a 2-uniform hypergraph does not preserve the maximum clique structure.


\begin{table}[!tbh]
    \centering
    \vspace{5mm}
    \caption{Results for the $\chi ^2$ value, residual and landmark error for the hardware experiment. Best results are in \textbf{BOLD}}
\begin{tabular}{|c|c|c|c|}
\hline
            & Chi2            & Residual       & LM Error (m)  \\ \hline
G$k$CM-HeuPatt        & \textbf{0.0066} & \textbf{49.83} & \textbf{5.32} \\ \hline
ROBIN-MILP  & \textbf{0.0066} & \textbf{49.83} & \textbf{5.32} \\ \hline
ROBIN-k-core & 53.71           & 412338         & 185           \\ \hline
PCM         & 1.02            & 7717           & 12.25         \\ \hline
SCGP        & 0.39            & 2952           & 35.19         \\ \hline
\end{tabular}
\label{tab:goats_results}
\end{table}

\subsection{Data Association}
In this experiment, we remove the assumption that the correspondence between a range measurement and its beacon is known. To accomplish this, we modified both the exact and heuristic algorithms to track the $n$ largest cliques where $n$ is the number of beacons in the environment assuming the number of beacons is known. Since each clique corresponds to consistent measurements that belong to a unique beacon, we enforce the constraint that a measurement cannot appear in more than one clique.

\begin{figure}[!tbh]
\vspace{-2mm} 
    \centering
    \includegraphics[width=0.93\columnwidth, trim=0 0 0 25]{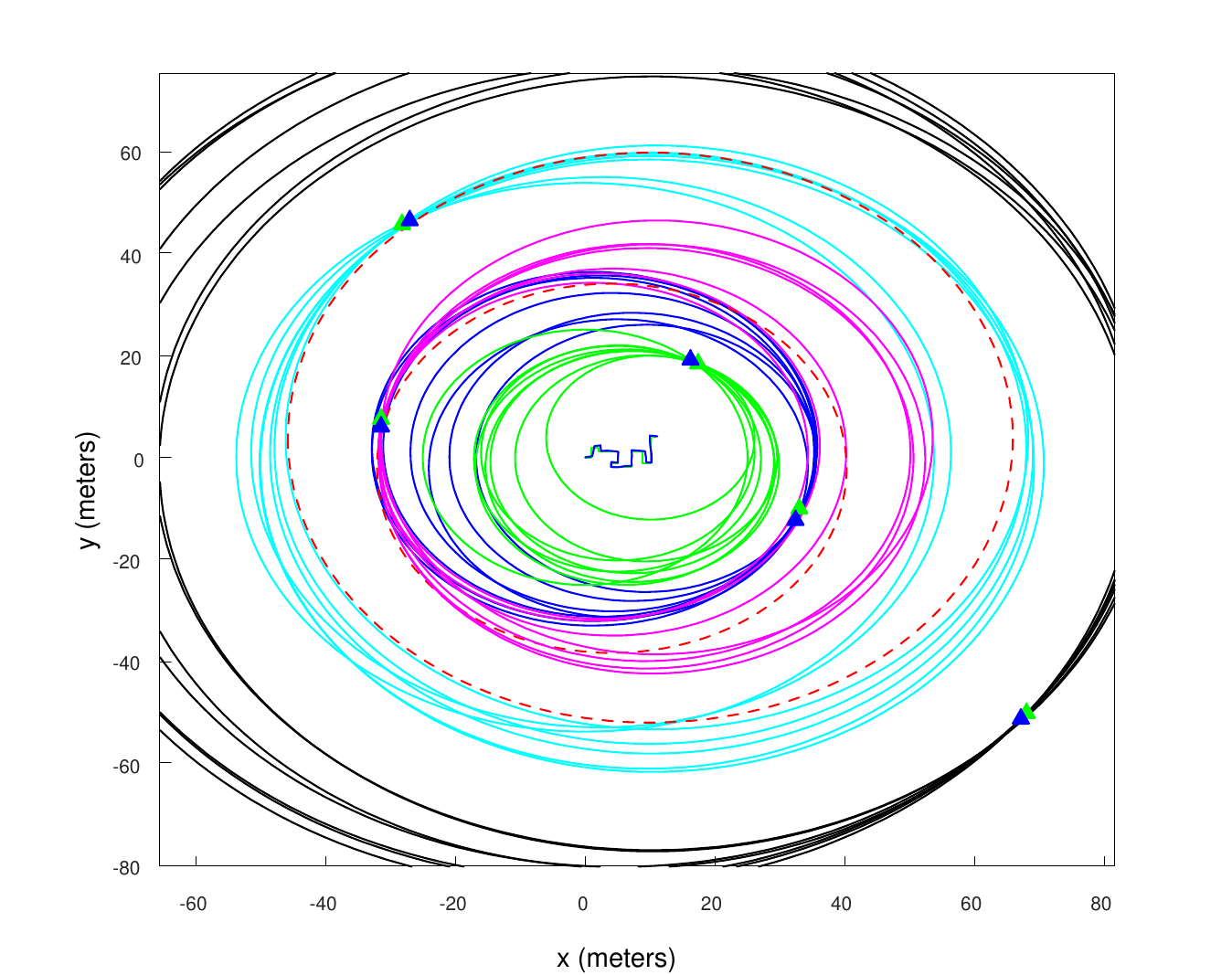}
    \caption{Results of G$k$CM-HeuPatt for performing data association and outlier rejection. Each clique found is shown in a different color. Measurements labeled as outliers included in the maximum clique are red dashed lines.}
    \label{fig:data_association}
\end{figure}

This experiment was run on a short trajectory of 30 poses where five measurements were received at each pose (one to each beacon). As such 150 measurements are being considered by the G$k$CM algorithm. Results were averaged over 81 different trials. Visual results can be seen in \cref{fig:data_association} while statistics are in \cref{tab:data_assoc_stats}. G$k$CM correctly identifies the five cliques corresponding to the different beacons and outperforms PCM in all the metrics. We do not provide a comparison with the other algorithms since they are unable to track multiple cliques simultaneously.


\begin{table*}[tbh]
    \centering
    \vspace{5mm}
    \caption{Statistics for G$k$CM and PCM in Data Association experiment. Best results are in \textbf{BOLD}}

\resizebox*{\linewidth}{!}{
\begin{tabular}{|c|cc|cc|cc|cc|cc|ccc|}
\hline
\multirow{2}{*}{} & \multicolumn{2}{c|}{Translational RMSE (m)} & \multicolumn{2}{c|}{Rotational  RMSE (rad)} & \multicolumn{2}{c|}{Beacon Error (m)} & \multicolumn{2}{c|}{Residual}    & \multicolumn{2}{c|}{Inliers}   & \multicolumn{3}{c|}{$\chi^2$}                                                         \\ \cline{2-14}
                  & \multicolumn{1}{c|}{Avg}      & Std         & \multicolumn{1}{c|}{Avg}      & Std         & \multicolumn{1}{c|}{Avg}    & Std     & \multicolumn{1}{c|}{Avg} & Std   & \multicolumn{1}{c|}{TPR} & FPR   & \multicolumn{1}{c|}{Avg} & \multicolumn{1}{c|}{Std} & \multicolumn{1}{l|}{Median} \\ \hline
G$k$CM-HeuPatt            & \textbf{0.6259}                        & \textbf{0.5646}      & \textbf{0.2469}                        & \textbf{0.0629}      & \textbf{34.14}                       & 52.06   & \textbf{72.12}                    & \textbf{88.67} & 0.84                    & \textbf{0.008} & \textbf{1.05}                     & \textbf{1.26}                     & \textbf{0.306}                       \\ \hline
PCM               & 3.153                         & 3.611       & 0.4521                        & 0.2218      & 40.28                       & \textbf{51.03}   & 1010                     & 1037  & \textbf{0.95}                    & 0.017 & 13.87                    & 14.16                    & 29.28                       \\ \hline
\end{tabular}
}
\label{tab:data_assoc_stats}
\end{table*}

\subsection{Tuning Experiment}
\begin{figure}[!tbh]
    \centering
    \includegraphics[width=\columnwidth, trim=0 0 0 25, clip]{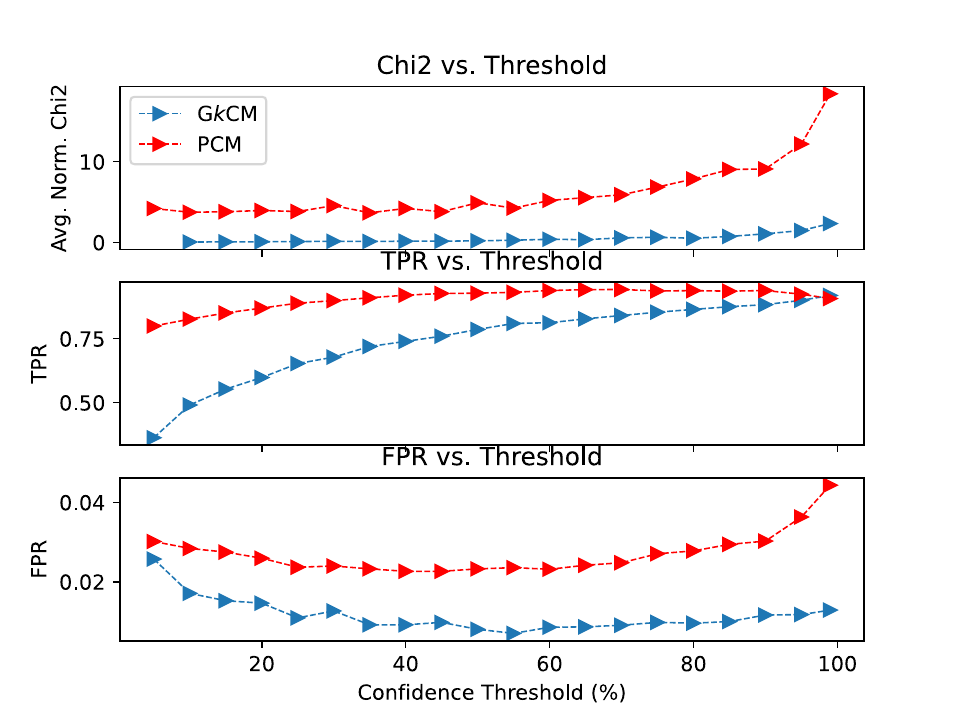}
    \caption{Results showing the normalized chi2 value, TPR, and FPR by varying the consistency threshold value, $\gamma$, for a fixed trajectory.}
    \label{fig:varied_chi2}
\end{figure}

PCM has the nice property that changing the threshold value, $\gamma$, did not significantly impact the results of the algorithm. Due to enforced group consistency, as opposed to pairwise, we designed an experiment to test if G$k$CM has a similar property. We accomplished this by fixing a robot trajectory of 50 poses and the associated measurements and running G$k$CM multiple times with a different value for $\gamma$ each time. The measurements contained 40 outliers that were generated as described previously. We averaged the $\chi^2$ value and the true and false positive rates over multiple runs. \Cref{fig:varied_chi2} shows how the above values vary with the consistency threshold for both G$k$CM and PCM.

As can be seen, G$k$CM-HeuPatt performs better than PCM in both the normalized $\chi^2$ and false positive rate, which is more important in our application than the true positive rate. The results indicate that the performance of G$k$CM-HeuPatt varies more with the threshold $\gamma$ than PCM, especially at very low and high confidence thresholds. As such, we recommend that confidence values be used from the $50-90\%$ confidence range where performance was less variable with the confidence threshold.



\subsection{Incremental Update}
\begin{figure}[!tbh]
    \centering
    \includegraphics[width=\columnwidth, trim=0 0 0 25, clip]{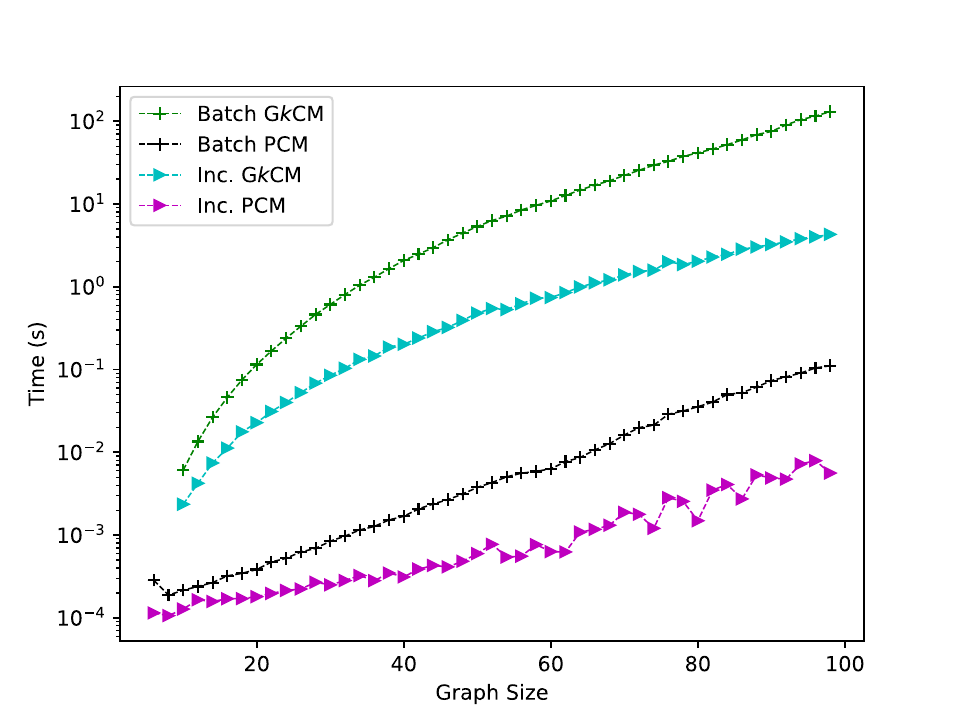}
    \caption{Timing data for both batch and incremental updates for G$k$CM and PCM. This includes the time to perform the relevant consistency checks and the new maximum clique. \emph{Note the log-scale on the vertical axis}.}
    \label{fig:timing_plot}
\end{figure}

In this last experiment we evaluate the incremental heuristic described by \cite{chang2021kimera} since their experiments only evaluated the heuristic for a $k$-uniform hypergraph where $k=2$. For this experiment, we generate a trajectory of 100 poses and measurements, and at each step, we evaluate how long both an incremental and batch update take. Updates include performing consistency checks and finding the maximum clique. We record the runtime for the graph size and average statistics over multiple runs. We plot the runtime against the size of the graph in \cref{fig:timing_plot}.

As can be seen, the incremental update with this heuristic provides similar benefits for G$k$CMas it does for PCM. On average, for a graph of 100 nodes with 80 outliers, it takes a batch solution over 40 seconds to solve for the maximum clique while it takes only 3 seconds for the incremental update. These findings validate the results presented by \cite{chang2021kimera} and also allow for G$k$CMto be run closer to real-time.

\section{Multi-agent Vision-based pose graph SLAM}
\label{sec:ma_visual_pgo}

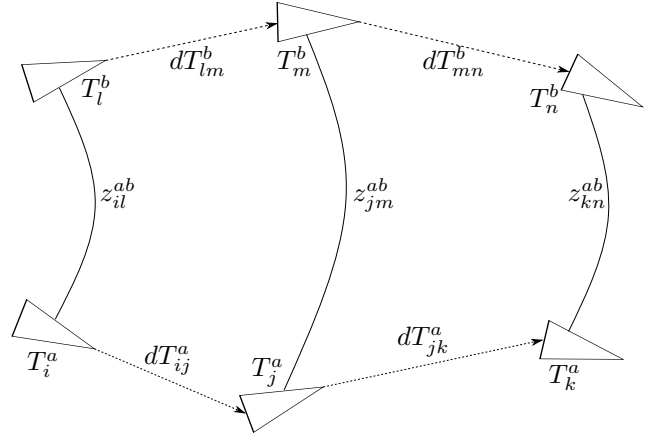
\begin{figure}[!tbh]
  \centering
 \begin{overpic}[width=0.98\columnwidth]{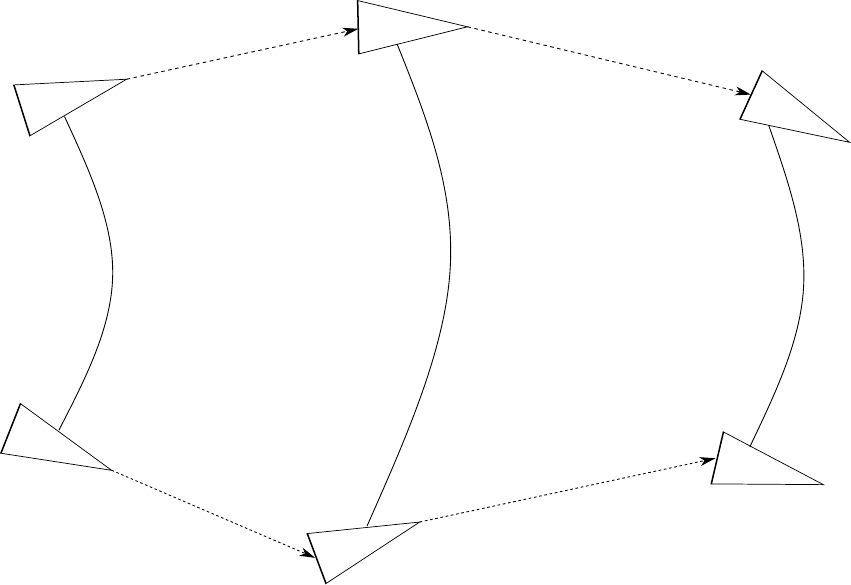}
  \put(2.5, 10){$T_i^a$}
  \put(38, 9){$T_j^a$}
  \put(85, 8){$T_k^a$}

  \put(21, 11){$dT_{ij}^a$}
  \put(61, 14){$dT_{jk}^a$}

  \put(11, 53.5){$T_l^b$}
  \put(42, 58){$T_m^b$}
  \put(82, 52){$T_n^b$}

  \put(25, 58){$dT_{lm}^b$}
  \put(65, 58){$dT_{mn}^b$}

  \put(14, 37){$z_{il}^{ab}$}
  \put(54, 37){$z_{jm}^{ab}$}
  \put(88, 37){$z_{kn}^{ab}$}
 \end{overpic}
  \caption{A visual of the setup used in the consistency check for a multi-agent visual PGO scenario. There are two agents each with odometry and several inter-vehicle measurements as outlined in \cref{eq:ma_visual_measurement}. All measurements are expressed in frame of robot $a$.} 
  \label{fig:ma_pgo_visual}
\end{figure}

In this section, we consider G$k$CM in the context of a multi-agent application where each agent is equipped with a monocular camera. We assume that the inter-vehicle measurements are generated by using a place recognition algorithm to identify if two images are of the same location followed by the use of bundle adjustment to estimate the relative pose between the two images up to a scale factor. The measurement takes the following form
\begin{equation}
  z_{ij}^{ab} = (\alpha, \epsilon, R_{ij}^{ab})
  \label{eq:ma_visual_measurement}
\end{equation}
where $R_{ij}^{ab}$ expresses the rotation of agent $b$ at pose $j$ in the frame of agent $a$ at pose $i$, and $\alpha$ and $\epsilon$ are the azimuth and elevation angles describing the direction of pose $j$ with respect to pose $i$. Lacking the scale factor causes the measurement to lose a degree of freedom over the full relative pose transformation and means that a pairwise check is no longer sufficient to check the consistency of the inter-vehicle measurements. We provide a visual reference of the setup in \cref{fig:ma_pgo_visual} showing the relationship of the poses and measurements between the two agents. The check, $C(z_{il}^{ab}, z_{jm}^{ab}, z_{kn}^{ab})$ has two parts that we outline below. The first part is very similar to the consistency check in \cref{eq:pairwise_consistency} except that we trace a loop using only the rotations as shown in \cref{eq:ma_visual_rotation_consistency_check}. This portion is repeated for all combinations of two measurements in the group of three.

\begin{equation}
  \begin{aligned}
  C(z_{il}^{ab}, z_{jm}^{ab})_R &= \left |\left| \hat{R}_{ij}^a \oplus R_{jm}^{ab} \oplus (\ominus \hat{R}_{lm}^b) \oplus (\ominus R_{il}^{ab}) \right|\right|_{\Sigma} \\ 
  C(z_{il}^{ab}, z_{jm}^{ab})_R &\leq \gamma_R
  \end{aligned}
  \label{eq:ma_visual_rotation_consistency_check}
\end{equation}

If this check passes, then we proceed with the second part of the consistency check, which verifies that the azimuth and elevation angles are consistent as outlined below
\begin{equation}
\begin{aligned}
  C(z_{il}^{ab}, z_{jm}^{ab}, z_{kn}^{ab})_d &= \left |\left| h(\bvec{X}_{ijk}^a, \bvec{X}_{lmn}^b, \bvec{Z}^{ab}_{il, jm, kn}) - \begin{pmatrix} \alpha \\ \epsilon \end{pmatrix} \right|\right|_{\Sigma} \\ 
  C(z_{il}^{ab}, z_{jm}^{ab}, z_{kn}^{ab})_d &\leq \gamma_d
\end{aligned}
  \label{eq:ma_visual_direction_consistency_check}
\end{equation}
where the function $h(\bvec{X}_{ijk}^a, \bvec{X}_{lmn}^b, \bvec{Z}^{ab}_{il, jm, kn})$ calculates the expected azimuth and elevation angle using the poses on agents $a$ and $b$ associated with the measurements in $\bvec{Z}$. The function $h$ first uses two of the measurements in $\bvec{Z}$ to estimate the scale in the direction of translation by solving a linear least-squares problem as follows
\begin{equation}
  \begin{aligned}
    \bvec{s} &= A^{-1} \bvec{b} \\
    A &= \begin{pmatrix} \bar{t}_{il} & R^a_{ij}\bar{t}_{jm} \end{pmatrix}\\
    \bvec{b} &= - (t^a_{ij} - R^a_{ij} R^{ab}_{jm} R^b_{ml} t^b_{lm}) \\
    \bar{t} &= \begin{pmatrix}
      \cos(\alpha)\cos(\epsilon) & \sin(\alpha)\cos(\epsilon) & \cos(\epsilon)
    \end{pmatrix}^T
  \end{aligned}
  \label{eq:ma_ls}
\end{equation}
where $\bar{t}$ is the unit vector denoting the direction indicated by the azimuth and elevation angles of a particular measurement, $t_{ij}^a$ denotes the position of $j$ with respect to $i$ for agent $a$, and $\bvec{s}$ is a two vector denoting the scale on the two measurements used. While testing this function, we found that it was sensitive to the errors in the rotations in the poses and measurements that would often result in a negative scale. To alleviate this, we first perform an initialization step using the single-loop technique described by \cite{carlone2015initialization}. We elected to use the single-loop technique because it has an algebraic solution and is less computationally intense than methods such as chordal initialization that perform very well in more complex pose graphs.

Once the scale has been recovered, the next step is to estimate the relative pose and uncertainty between the agents by applying the scale as follows
\begin{equation*}
  \begin{aligned}
    T_{il}^{ab} &= \begin{pmatrix} R_{il}^{ab} & s \bar{t}_{il}^{ab} \\
                                    0_{1x3} & 1 \end{pmatrix} \\
    \Sigma_T &= H \Sigma_z H^T
  \end{aligned}
\end{equation*}
where $\Sigma_T$ is the covariance matrix on the full relative pose, $\Sigma_z$ is the covariance on the measurement and $H$ is the jacobian of the function that applies the scale.
The next step is to express the poses involved in the third measurement in a common reference frame. These poses, which we will call $T_k^a$ and $T_n^b$, can be found as shown below.
\begin{equation*}
  \begin{aligned}
    T_k^a &= T_k^a \\
    T_n^b &= T_{il}^{ab} \oplus T_{ln}^b
  \end{aligned}
\end{equation*}
With this information, we can calculate the expected azimuth and elevation angles using \cref{eq:az_elev}.

\begin{equation}
  \begin{aligned}
    dT &= (\ominus T_k^a) \oplus T_n^b \\
    \alpha &= \text{atan2}(dT.y, dT.x) \\
    \epsilon &= \text{atan2}(dT.z, \sqrt{dT.x^2 + dT.y^2})
  \end{aligned}
  \label{eq:az_elev}
\end{equation}
The covariance on $dT$ is calculated using methods described by \cite{mangelson2020characterizing}, and the covariance on the expected azimuth and elevation angles are found by further pre and post-multiplying by the Jacobian matrix of their respective functions. This process is repeated three times so that each measurement can be validated.


\subsection{Degenerate Configurations}

This application is also susceptible to degenerate configurations. In scenarios such as when two poses lie exactly on top of each other, the relative pose cannot be recovered because the matrix in \cref{eq:ma_ls} becomes singular. Should such a degeneracy occur, solutions such as those discussed in \cref{sec:range_slam} may be used. In practice, however, such degeneracies never arose due to the noise inherent in the problem.

\section{Multi-agent Visual PGO Evaluation}
\label{sec:ma_visual_eval}  

This section describes the experimental setup and the results obtained for each of the experiments evaluating the efficacy of G$k$CM for multi-agent visual PGO applications. In each experiment we compare our results with the pairwise consistency approach in PCM, and with both the MILP and $k$-core variants of ROBIN. Due to run-time constraints, G$k$CM and PCM were only evaluated using \cref{alg:gmc_heuristic}.

\subsection{Simulation Results}
We first evaluated our algorithm in simulation. We randomly initialized two agents in a predetermined area and let the agents move about in a Manhattan world scenario. After generating the path, we searched for poses on each robot within half a meter of each other to generate an inter-robot measurement. For computational purposes, we selected 100 measurements to be used, and if 100 measurements were not generated we reran the simulation until 100 measurements were present. We then ran the G$k$CM, PCM, and ROBIN (using both the MILP and $k$-core clique solvers) to find the largest inlier set. Statistics such as the TPR, FPR, and RMSE of the resulting PGO routine were recorded and averaged over 100 runs. To initialize the relative pose, we use chordal initialization (\cite{carlone2015initialization}) to generate rotation estimates. To initialize the translation, we picked two measurements from the set of inlier measurements and estimated the scale by solving \cref{eq:ma_ls} and applying the scale to the unit vector created by the azimuth and elevation angles.

\begin{figure*}[tbh!]%
  \centering%
  \subfigure{%
    \includegraphics[width=0.70\columnwidth,trim={0.05cm 0.3cm 0.05cm 0cm},clip]{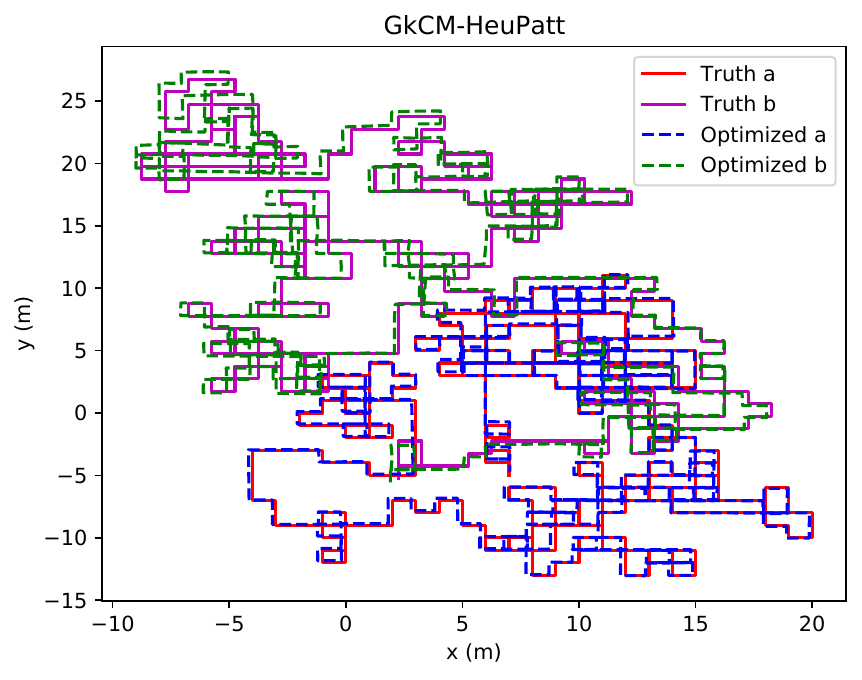}%
    \label{sfig:ma_mc_gkcm}%
  }%
  \subfigure{%
    \includegraphics[width=0.70\columnwidth,trim={0.05cm 0.3cm 0.05cm 0cm},clip]{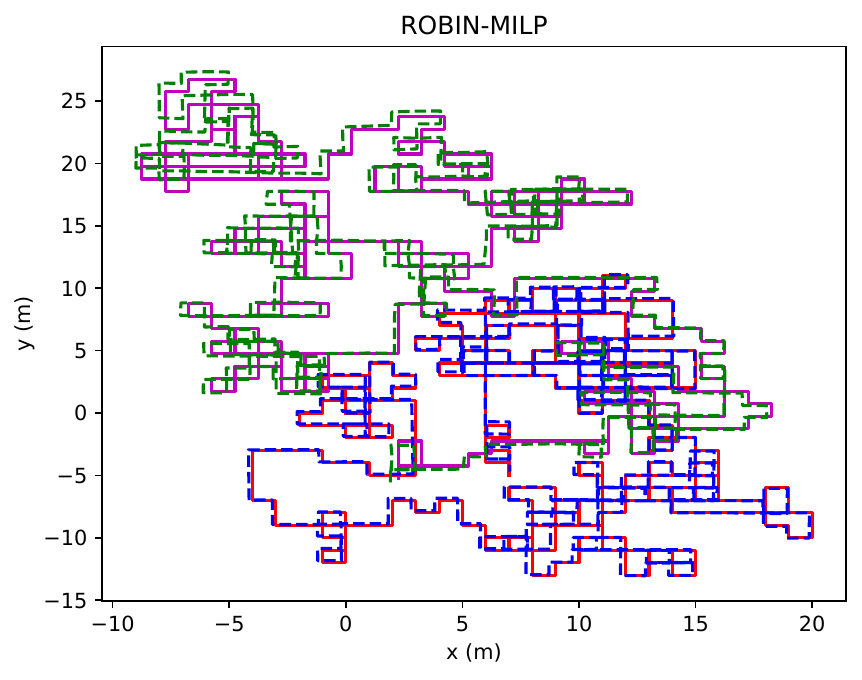}%
    \label{sfig:ma_mc_milp}%
  }%
  \\
  \subfigure{%
    \includegraphics[width=0.70\columnwidth,trim={0.05cm 0.3cm 0.05cm 0cm},clip]{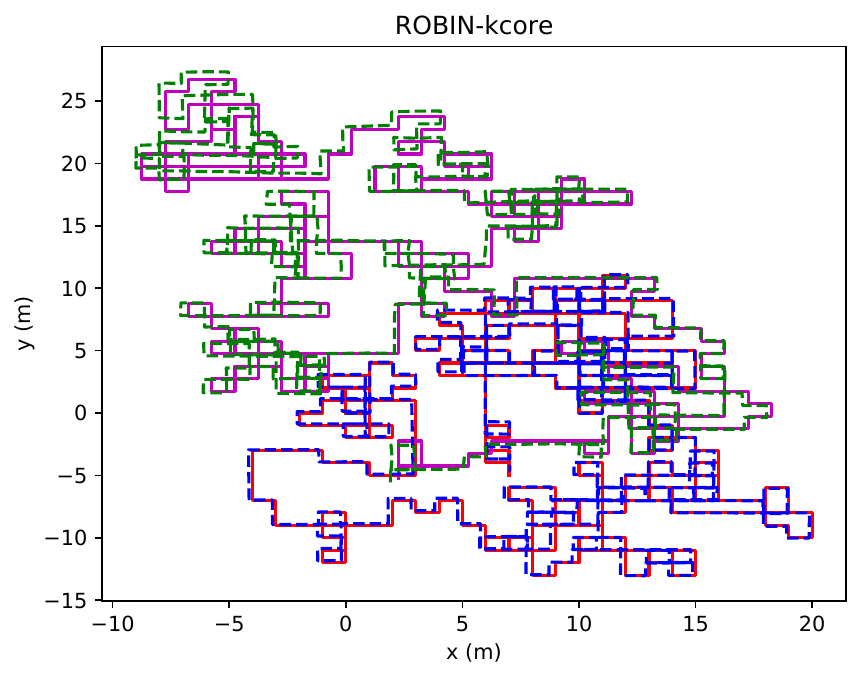}%
    \label{sfig:ma_mc_kcore}%
  }%
  \subfigure{%
    \includegraphics[width=0.70\columnwidth,trim={0.05cm 0.3cm 0.05cm 0cm},clip]{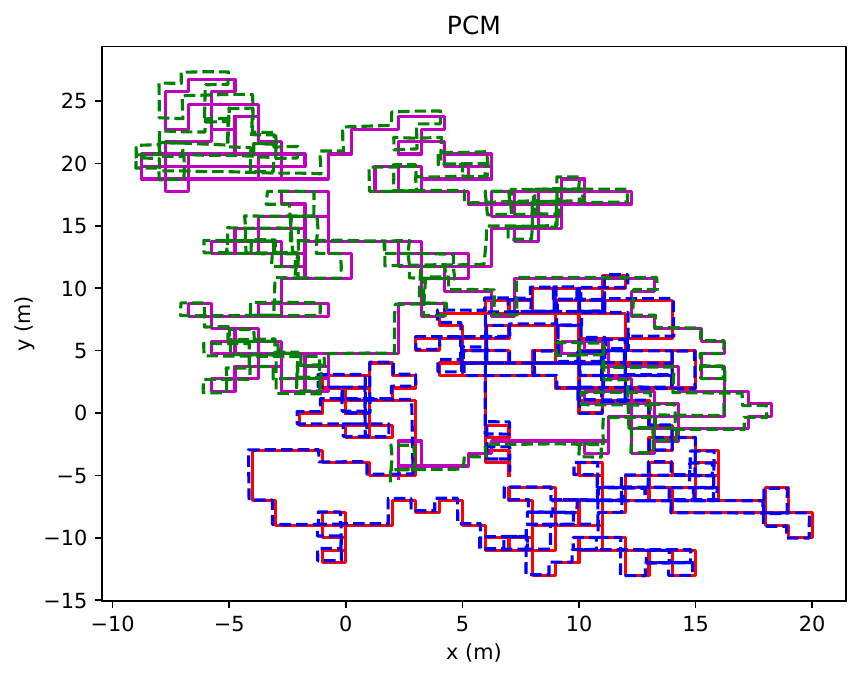}%
    \label{sfig:ma_mc_pcm}%
  }%
  \caption{Results for each of the algorithms in the multi-agent visual PGO simulation. The red and blue solid lines denote the true trajectories of each vehicle while the dashed lines denote the estimated trajectories.}
  \label{fig:ma_visual_pgo_sim}
\end{figure*}

As can be seen from the results, all of the methods performed fairly equally across all the metrics.  G$k$CM had the best performance in the most tracked statistics. However, we note that the results of all methods are comparable. The biggest discrepancies lie in both the residual and normalized $\chi^2$ statistics for the resulting map. We believe that these discrepancies are a result of differences in the initialization. We could not guarantee that the same measurements were used to initialize the scale on the translation vector between the two agents since the inlier cliques were not guaranteed to be the same. Often the cliques had five or six measurements in common with one or two that varied between them. Due to the nature of least-squares methods, this caused the scale to vary, and occasionally not be good enough for the map to converge to the exact same solution. We note that this is a problem with the initialization of the graph, which is not the focus of this research. Each of these methods was able to reject nearly all of the outlier measurements. Visual results for a single trial can be seen in \cref{fig:ma_visual_pgo_sim}.

\begin{table*}[!tbh]
    \centering
    \vspace{3mm}
    \caption{Statistics comparing G$k$CM to other methods in the multi-agent simulation experiment. Best results are in \textbf{BOLD}}
\begin{tabular}{|c|cc|cc|cc|cc|ccc|}
\hline
\multirow{2}{*}{} & \multicolumn{2}{c|}{Trans. Error (m)}    & \multicolumn{2}{c|}{Rot.  Error (rad)}    & \multicolumn{2}{c|}{Residual}            & \multicolumn{2}{c|}{Inliers}          & \multicolumn{3}{c|}{$\chi^2$}                                        \\ \cline{2-12}
                  & \multicolumn{1}{c|}{Avg} & Std          & \multicolumn{1}{c|}{Avg} & Std           & \multicolumn{1}{c|}{Avg} & Std           & \multicolumn{1}{c|}{TPR} & FPR        & \multicolumn{1}{c|}{Avg} & \multicolumn{1}{c|}{Std} & Median         \\ \hline
G$k$CM-HeuPatt    & \textbf{1.26}& \textbf{1.87} & 0.09 & \textbf{0.06} & \textbf{8918} & \textbf{40183} & 0.74 & \textbf{2.2e-4} & \textbf{2.95} & \textbf{13.27} & 0.011 \\ \hline
PCM & 1.36 & 2.00 & 0.086 & 0.082 & 11314 & 40705 & \textbf{0.99} & \textbf{2.2e-4} & 3.73 & 13.43 & 0.014 \\ \hline
ROBIN-MILP        & 1.34 & 2.18 & 0.094 & 0.082 & 11399 & 48786 & 0.74  & \textbf{2.2e-4} & 3.77 & 16.13 & \textbf{0.010} \\ \hline
ROBIN-$k$-core       & 1.41 & 2.05 & \textbf{0.085} & 0.082 & 11862 & 41290 & \textbf{0.99} & \textbf{2.2e-4} & 3.91 & 13.62 & 0.015          \\ \hline
\end{tabular}
\label{tab:ma_visual_pgo_sim}
\end{table*}

\subsection{Hardware Results}

Having shown that the G$k$CM algorithm is effective at rejecting outliers in a simulation environment, we now seek to validate our algorithm using hardware data. In this experiment, we used the NCLT dataset presented by \cite{ncarlevaris-2015a} because it provides data of a robot exploring the same area over multiple sessions, and has images from several cameras placed on the robot. We selected two sessions that occurred in the same season and around the same time of day to facilitate place recognition, which was done using the OpenFABMap library developed by \cite{glover2012openfabmap}. We trained OpenFABMap using data collected in one session, and created matches to places in the second session. Due to the size of the dataset, we elected to use only one of the cameras onboard the robot and used every 50th image.

With this setup, we generated 267 matches of which 25 were identified as true matches while the others were labeled as outliers due to perceptual aliasing. We estimated the relative pose between the two images by decomposing the essential matrix using OpenCV, which was then refined through bundle adjustment using GTSAM \cite{dellaert2012factor}. From this relative pose, we generated the measurement in \cref{eq:ma_visual_measurement} using \cref{eq:az_elev} and transformed the covariance of the relative pose to the measurement covariance using the Jacobian of \cref{eq:az_elev}. Once the measurements were generated, we then checked the consistency of all the measurements and found the maximum clique using each of the methods discussed previously.

\begin{figure*}[tbh!]%
  \centering%
  \subfigure{%
    \includegraphics[width=0.70\columnwidth,trim={0.05cm 0.3cm 0.05cm 0cm},clip]{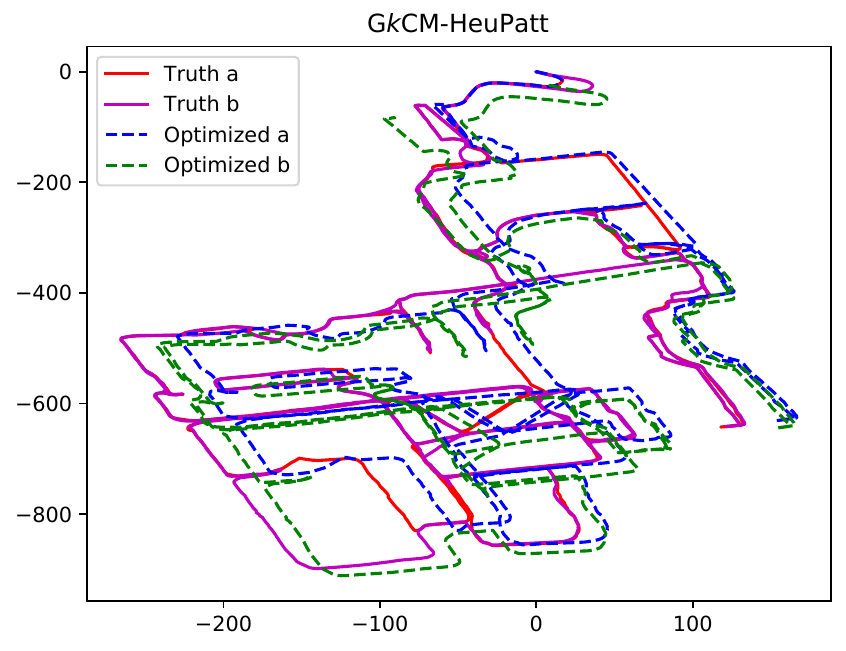}%
    \label{sfig:ma_hardware_gkcm}%
  }%
  \subfigure{%
    \includegraphics[width=0.70\columnwidth,trim={0.05cm 0.3cm 0.05cm 0cm},clip]{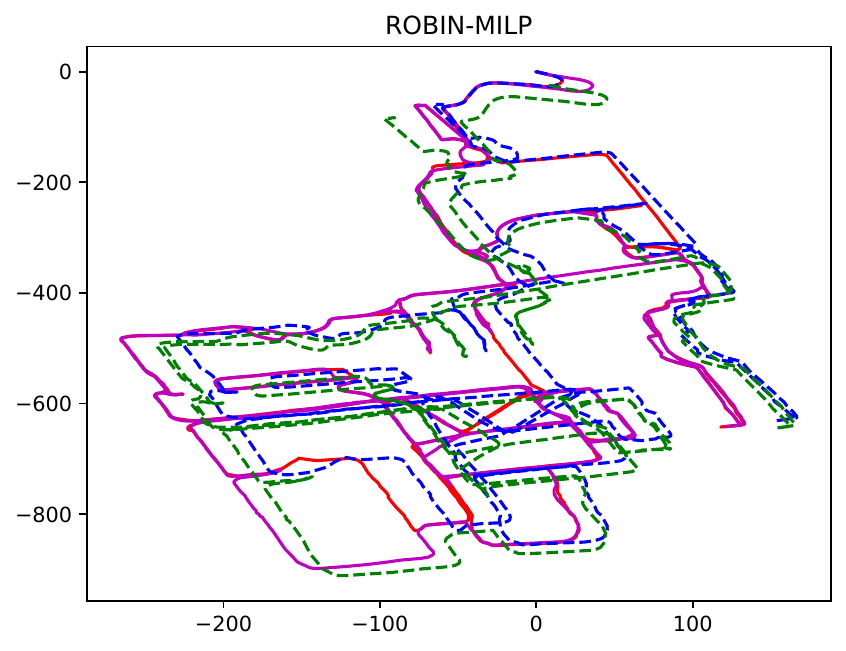}%
    \label{sfig:ma_hardware_milp}%
  }%
  \\
  \subfigure{%
    \includegraphics[width=0.70\columnwidth,trim={0.05cm 0.3cm 0.05cm 0cm},clip]{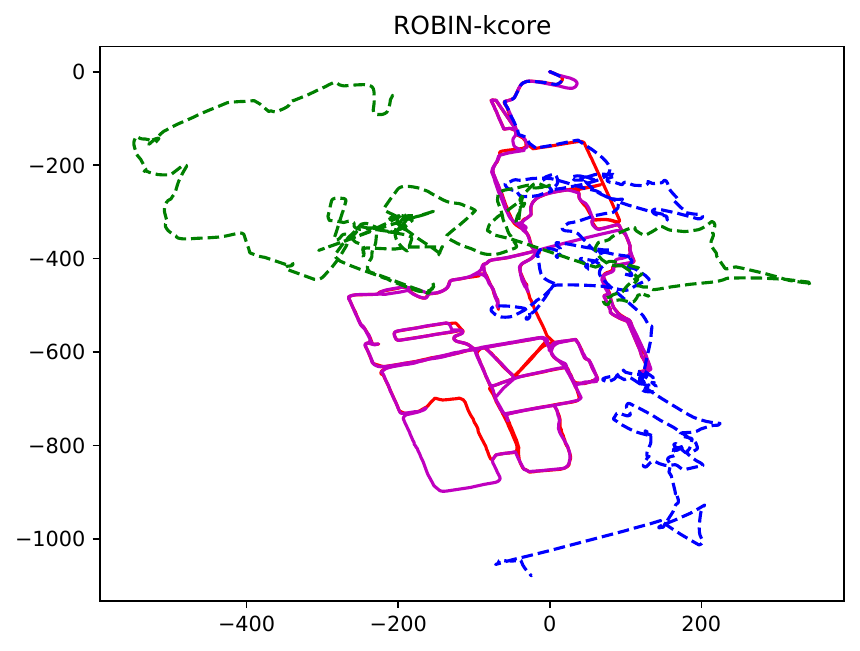}%
    \label{sfig:ma_hardware_kcore}%
  }%
  \subfigure{%
    \includegraphics[width=0.70\columnwidth,trim={0.05cm 0.3cm 0.05cm 0cm},clip]{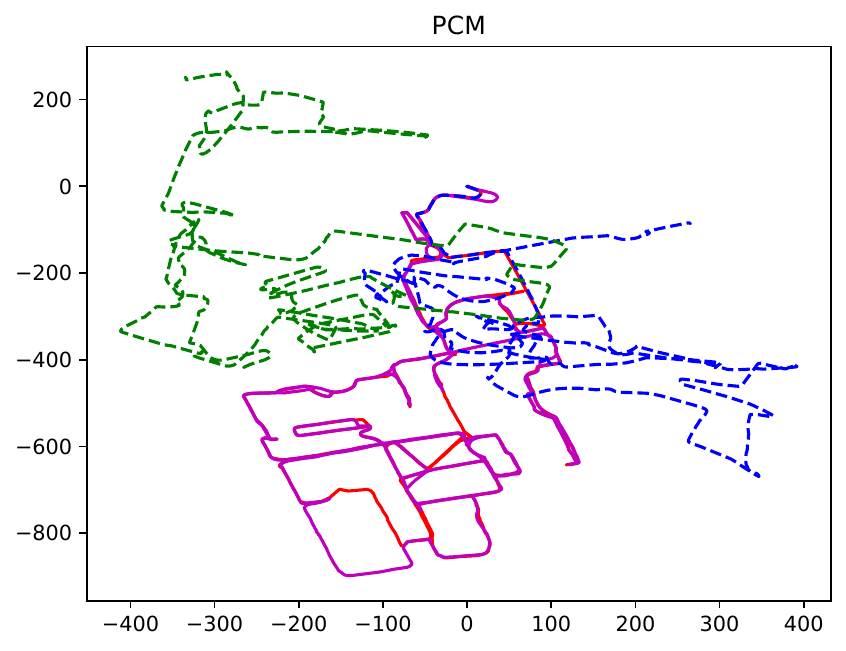}%
    \label{sfig:ma_hardwae_pcm}%
  }%
  \caption{Results for the hardware experiment. The solid red and purple lines are the true robot trajectories generated using GPS, Lidar, and odometry. The dotted lines are the estimated trajectories using odometry and inter-vehicle measurements from a camera.}
  \label{fig:ma_visual_hardware}
  \vspace{5mm}
\end{figure*}

\begin{table*}[!tbh]
  \centering
  \caption{Numerical results for the multi-agent hardware experiment. Best results are in \textbf{BOLD}.}
  \label{tab:ma_visual_hardware}
\begin{tabular}{|c|c|c|c|c|}
\hline
Metric                                                                     & G$k$CM-HeuPatt  & ROBIN-MILP & ROBIN-$k$-core & PCM     \\ \hline
\begin{tabular}[c]{@{}c@{}}Relative Translation\\ Error (m)\end{tabular}   & 2.70e-3 & 2.70e-3    & \textbf{2.64e-3}     & 2.70e-3 \\ \hline
\begin{tabular}[c]{@{}c@{}}Relative Rotational \\ Error (rad)\end{tabular} & \textbf{0.012}   & \textbf{0.012}      & 0.025       & 0.018   \\ \hline
Residual                                                                   & \textbf{19899}   & \textbf{19899}      & 5.58e6      & 2.24e5  \\ \hline
Normalized $\chi^2$                                                        & \textbf{0.12}    & \textbf{0.12}       & 34.99       & 1.40    \\ \hline
\end{tabular}
\end{table*}

A visual representation of our results can be found in \cref{fig:ma_visual_hardware}, while the statistics are in \cref{tab:ma_visual_hardware}. As can be seen, both G$k$CM and the MILP are able to successfully filter out the outlier measurements and fuse the maps from the two robots. Both the k-core and PCM methods found cliques that contained outlier measurements that ruined the shape of the resulting map. In \cref{tab:ma_visual_hardware}, we present several statistics including the relative translation and rotational error defined as $\xi = \text{Log}((T_{k+1}^k)^{-1}(\hat{T}_k^{-1} \hat{T}_{k+1}))$.

Both the G$k$CM-HeuPatt and the MILP techniques have the best results across all the metrics identified, except for a small difference in the average relative translational error, while the PCM and k-core methods produced unusable results. The $\chi^2$ value for PCM post-optimization was surprisingly low considering the poor quality of the map produced. This is attributed to the small number of inter-vehicle constraints when compared with the number of poses in the graph. Further investigation of the relative error statistics shows that the majority of the error is contained between a rather small number of poses in the graph.


The results produced by the G$k$CM-HeuPatt and MILP methods successfully merge the maps with some error in the offset in the origins of the two maps. The map of the second agent (shown by the greed dotted line in \cref{fig:ma_visual_hardware}) should lie nearly on top of the blue dotted line but often is shifted down and to the left. We believe this level of error to be acceptable for several reasons. The first is that much less information was used when fusing the maps in our estimated results than was used in generating the ground truth. The ground truth data fused RTK-GPS, lidar scans, and the odometry data from all 22 sessions. Lidar scans were aligned to generate constraints both in a single session and between sessions. This allowed many full-degree-of-freedom constraints to be generated within each session and between the different sessions. Comparatively, our solution uses only a subset of the visual information provided, and only uses this information to generate constraints between the two sessions. No loop closure constraints were generated when the robot visited a location it had visited before in the same session. Additionally, using only a single camera increases the difficulty of the problem because the scale is not observable in the inter-session constraints. This problem does not arise when comparing lidar scans.

Lastly, it seems a little surprising that the simulation experiments and hardware experiments produced such different results. In simulation all methods had a fairly similar level of performance. We attribute this to the length of the trajectories in each case. The simulated experiments had a trajectory consisting of about 500 nodes per robot while the two sessions in the NCLT dataset had over 23000 and 29000 nodes. This means that the loops traversed in the NCLT dataset are longer and that the associated covariances used when performing the consistency checks were larger resulting in a denser consistency graph since more combinations passed the consistency check. The result of this is that more measurements were labeled as consistent when evaluating the NCLT dataset. PCM did not have the benefit of the direction information to filter out more outlier measurements like G$k$CM and ROBIN-MILP. ROBIN-$k$core had the benefit of the direction information but the denser graph meant that the structure of the maximum clique was not preserved when the 3-hypergraph was embedded in a 2-hypergraph resulting in a poor set of measurements being chosen.

\section{Conclusion}
\label{sec:conclusion}

In this paper, we presented a unification of the theory of consistency, starting with pairwise consistency and generalizing to group-$k$ consistency. We present the group-$k$ consistency maximization algorithm and the associated maximum clique algorithms that function over generalized graphs and show that we can effectively choose a consistent set of measurements in high-outlier regimes in both range-based SLAM and multi-agent visual SLAM problems. Techniques to alleviate the exponential nature of evaluating consistency and finding the maximum clique were presented. Lastly, we released an open-source implementation of our library for future use by the research community.



\begin{acks}
We would like to thank Allan Papalia and the Marine Robotics Group for providing the range-based SLAM dataset for us to use in evaluating our algorithms.
\end{acks}

\begin{funding}
    The authors disclosed receipt of the following financial support for the research, authorship, and/or publication of this article: This work has been funded by the Office of Naval Research [award number N00014-21-1-2435], \BF{Also get my stuff}.
\end{funding}

\begin{dci}
    The Authors declare that there is no conflict of interest.
\end{dci}


\bibliographystyle{style/SageH}
\bibliography{references/strings-short, references/library}

\end{document}